\RequirePackage[
  colorlinks=true, % enable colored links
  allcolors=blue, % set all colors to blue
  hypertexnames=false, % use non-conflicting hypertext names
  pdfpagelabels=false, % disable the generation of PDF page labels
]{hyperref}

\documentclass[draft]{agujournal2019}
\usepackage{url} %this package should fix any errors with URLs in refs.
\usepackage{lineno}
\usepackage{soul}
\usepackage{algorithmic}
\usepackage{algorithm}
\usepackage{bbm}
\usepackage{amsmath}
\usepackage{amssymb}
\usepackage{mathtools}
\usepackage{xcolor}
\usepackage{array}

\justifying

\draftfalse

\journalname{}
\begin{document}

%%%%%%%%%%%%%%%%%%% VERSION WITH COLOR ON MODIFICATIONS
% \DeclareRobustCommand{\rev}[1]{\textcolor{purple}{#1}}
%%%%%%%%%%%%%%%%%%%
%%%%%%%%%%%%%%%%%%% VERSION WITHOUT COLOR 
\DeclareRobustCommand{\rev}{}
%%%%%%%%%%%%%%%%%%% 

%%%%%%%%%%%%%%%%%%%%%%%%%%%%%%%%%%%%%%%%%%%%%%%
%  TITLE
%
% (A title should be specific, informative, and brief. Use
% abbreviations only if they are defined in the abstract. Titles that
% start with general keywords then specific terms are optimized in
% searches)
%
%%%%%%%%%%%%%%%%%%%%%%%%%%%%%%%%%%%%%%%%%%%%%%%

% Example: \title{This is a test title}

\title{\rev{Learning Regionalization using Accurate Spatial Cost Gradients within a Differentiable High-Resolution Hydrological Model: Application to the French Mediterranean Region}}

%%%%%%%%%%%%%%%%%%%%%%%%%%%%%%%%%%%%%%%%%%%%%%%
%
%  AUTHORS AND AFFILIATIONS
%
%%%%%%%%%%%%%%%%%%%%%%%%%%%%%%%%%%%%%%%%%%%%%%%

% Authors are individuals who have significantly contributed to the
% research and preparation of the article. Group authors are allowed, if
% each author in the group is separately identified in an appendix.)

% List authors by first name or initial followed by last name and
% separated by commas. Use \affil{} to number affiliations, and
% \thanks{} for author notes.
% Additional author notes should be indicated with \thanks{} (for
% example, for current addresses).

% Example: \authors{A. B. Author\affil{1}\thanks{Current address, Antartica}, B. C. Author\affil{2,3}, and D. E.
% Author\affil{3,4}\thanks{Also funded by Monsanto.}}

\authors{
Ngo Nghi Truyen Huynh\affil{1},  
Pierre-André Garambois\affil{1},
François Colleoni\affil{1}, 
Benjamin Renard\affil{1}, 
Hélène Roux\affil{2}, 
Julie Demargne\affil{3}, 
Maxime Jay-Allemand\affil{3},
Pierre Javelle\affil{1}
}

\affiliation{1}{INRAE, Aix-Marseille Université, RECOVER, 3275 Route Cézanne, 13182 Aix-en-Provence, France}
\affiliation{2}{Institut de Mécanique des Fluides de Toulouse (IMFT), Université de Toulouse, CNRS, 31400 Toulouse, France}
\affiliation{3}{HYDRIS Hydrologie, Parc Scientifique Agropolis II, 2196 Boulevard de la Lironde, 34980 Montferrier-sur-Lez, France}
% \affiliation{4}{Fourth Affiliation}

%\affiliation{=number=}{=Affiliation Address=}
%(repeat as many times as is necessary)

% Corresponding author mailing address and e-mail address:

% (include name and email addresses of the corresponding author.  More
% than one corresponding author is allowed in this LaTeX file and for
% publication; but only one corresponding author is allowed in our
% editorial system.)

% Example: \correspondingauthor{First and Last Name}{email@address.edu}

\correspondingauthor{Ngo Nghi Truyen Huynh}{ngo-nghi-truyen.huynh@inrae.fr}

%%%%%%%%%%%%%%%%%%%%%%%%%%%%%%%%%%%%%%%%%%%%%%%
% KEY POINTS
%%%%%%%%%%%%%%%%%%%%%%%%%%%%%%%%%%%%%%%%%%%%%%%
%  List up to three key points (at least one is required)
%  Key Points summarize the main points and conclusions of the article
%  Each must be 140 characters or fewer with no special characters or punctuation and must be complete sentences

% Example:
% \begin{keypoints}
% \item	List up to three key points (at least one is required)
% \item	Key Points summarize the main points and conclusions of the article
% \item	Each must be 140 characters or fewer with no special characters or punctuation and must be complete sentences
% \end{keypoints}

\begin{keypoints}
\item \rev{Novel approach for regional calibration of a distributed hydrologic model using learnable and non-linear descriptors-to-parameters mappings} %New Hybrid Data Assimilation and Parameter Regionalization (HDA-PR) approach 
\item Original combination of numerical adjoint model and neural network Jacobian\rev{: accurate gradients enable high-dimensional optimization}
%\item \rev{Neural networks enable learning non-linear and multivariate descriptors-to-parameters mappings} 
\item{\rev{Extensive case study in flash-flood-prone Mediterranean region shows effective regionalization of high-resolution model with neural network}} 
%Effective regionalization of high-resolution hydrological model for flash floods in the Mediterranean region
\end{keypoints}

%two differentiable codes can be linked and their gradient chained
%%%%%%%%%%%%%%%%%%%%%%%%%%%%%%%%%%%%%%%%%%%%%%%
%
%  ABSTRACT and PLAIN LANGUAGE SUMMARY
%
% A good Abstract will begin with a short description of the problem
% being addressed, briefly describe the new data or analyses, then
% briefly states the main conclusion(s) and how they are supported and
% uncertainties.

% The Plain Language Summary should be written for a broad audience,
% including journalists and the science-interested public, that will not have 
% a background in your field.
%
% A Plain Language Summary is required in GRL, JGR: Planets, JGR: Biogeosciences,
% JGR: Oceans, G-Cubed, Reviews of Geophysics, and JAMES.
% see http://sharingscience.agu.org/creating-plain-language-summary/)
%
%%%%%%%%%%%%%%%%%%%%%%%%%%%%%%%%%%%%%%%%%%%%%%%

\begin{abstract}
Estimating spatially distributed hydrological parameters in ungauged catchments poses a challenging regionalization problem and requires imposing spatial constraints given the sparsity of discharge data. A possible approach is to search for a transfer function that quantitatively relates physical descriptors to conceptual model parameters. This paper introduces a Hybrid Data Assimilation and Parameter Regionalization (HDA-PR) approach incorporating learnable regionalization mappings, based on either multi-linear regressions or \rev{artificial} neural networks \rev{(ANNs)}, into a differentiable hydrological model. \rev{This approach demonstrates how two differentiable codes can be linked and their gradients chained, enabling} the exploitation of heterogeneous datasets across extensive spatio-temporal computational domains within a high-dimensional regionalization context, using accurate adjoint-based gradients. The inverse problem is tackled with a multi-gauge calibration cost function accounting for information from multiple observation sites. HDA-PR was tested on high-resolution, hourly and kilometric regional modeling of \rev{126 flash-flood-prone catchments in the French Mediterranean region. The results highlight a strong regionalization performance of HDA-PR especially in the most challenging upstream-to-downstream extrapolation scenario with ANN, achieving median Nash-Sutcliffe efficiency (NSE) scores from 0.6 to 0.71 for spatial, temporal, spatio-temporal validations, and improving NSE by up to 30\% on average compared to the baseline model calibrated with lumped parameters. Multiple evaluation metrics based on flood-oriented hydrological signatures also indicate that the use of an ANN leads to better performances than a multi-linear regression in a validation context. ANN enables to learn a non-linear descriptors-to-parameters mapping which provides better model controllability than a linear mapping for complex calibration cases.}
%The regionalization method is amenable to state-parameter correction from multi-source data over a range of time scales needed for operational data assimilation, and it is adaptable to other differentiable geophysical models.
\end{abstract}
%Multiple evaluation metrics based on flood-oriented hydrological signatures are also employed to assess the accuracy and robustness of the approach.

%%%%%%%%%%%%%%%%%%%%%%%%%%%%%%%%%%%%%%%%%%%%%%%
%
%  BODY TEXT
%
%%%%%%%%%%%%%%%%%%%%%%%%%%%%%%%%%%%%%%%%%%%%%%%

\section{Introduction}

Irrespective of their type and complexity, hydrological models are more or less empirical and uncertain representations of multiscale coupled hydrological processes whose observability is limited. Hydrological model parameters are in general effective quantities that cannot be directly measured. Instead, they are typically inferred through a calibration procedure aimed primarily at obtaining satisfactory streamflow simulations (e.g., \citeA{Beven_hess_2001,Kirchner2006, gupta2006model,vrugt2008treatment}). In most cases, this optimization problem is a difficult ill-posed inverse problem faced with the equifinality \cite{Beven_hess_2001} of feasible solutions, which can be further interpreted as model structural equifinality and spatial equifinality in the context of spatially sparse observations compared to model controls (see for example discussions in \citeA{GARAMBOIS2020}).
Most calibration approaches enable the estimation of spatially uniform model parameters for a single gauged catchment, but this leads to piecewise constant and discontinuous parameters fields for adjacent catchments. Moreover, parameter sets determined through calibration are not transferable to ungauged locations although the latter represent the majority of the global land surface \cite{fekete2007current, hannah2011large}. Therefore, prediction in ungauged basins remains a ``grand challenge'' \cite{Sivapalan_2003_regio} in hydrology \cite{Hrachowitz_2013}, which hinders the development of effective high-resolution models adapted to the simulation of hydrological extremes in a context of high data uncertainty (e.g., for Mediterranean flash floods in \citeA{garambois2015parameter, jay2024spatially}).

The estimation of hydrological model parameters in ungauged regions is performed with so-called regionalization approaches that exploit and transfer hydrological information from gauged to ungauged catchments using various descriptors of catchments physical properties (see reviews in \citeA{bloschl2013runoff, samaniego2010multiscale, Hrachowitz_2013, beck2020global}). The most widely used approach in early regionalization studies involves independent catchment-by-catchment calibrations, followed by multiple regression or interpolation techniques to transfer the calibrated parameter sets from gauged to ungauged locations \cite{abdulla1997development, seibert1999regionalisation, parajka2005comparison, razavi2013streamflow, parajka2013comparative} and can be called post-regionalization \cite{samaniego2010multiscale}. This approach presumes that the variability of calibrated model parameters through the catchments \rev{is related, for instance, to} spatial proximity \cite{widen2007global, oudin2008spatial}, and physical or climatic similarity \cite{oudin2010seemingly,beck2016global}. 
Statistical learning methods\rev{, including machine learning models and artificial neural networks (ANNs),} have also been applied in post-regionalization to explore the relationships between physical descriptors and calibrated parameter sets at gauged locations (e.g., \citeA{BASTOLA2008188,saadi2019random,wang2023research}). 
However, post-regionalization approaches are limited to lumped parameters, thus ignoring within-catchment variabilities (see reviews in \citeA{samaniego2010multiscale, razavi2013streamflow}),  
except when calibrated parameters correspond to tuning factors of physical pedotranfer functions or hydraulic frictions correspondence tables (see \citeA{garambois2015parameter} for details). The identification of transfer functions in post-regionalization is complicated by the uncertainty of estimated parameter sets, while spatial proximity is mostly applicable to densely gauged river networks and regions \cite{oudin2008spatial,Reichl_2009}. Moreover, incorporating a statistical learning process, especially unsupervised learning approaches, in the post-regionalization step can exacerbate existing issues such as parameter biases induced by data measurement errors \cite{kavetski2006bayesian}. A regionalized calibration simultaneously exploiting the information of multiple gauges, within spatial clusters defined a priori from descriptors, is performed in \citeA{HUANG2019} over Norway using climatic similarity. The  parameters calibrated over multiple gauges of a climatic zone are applied to ungauged catchments of the same zone. This approach does not account for hydrological heterogeneity within the catchments or within the regional clusters determined by physical similarity, which can have a major impact on forecasting, \rev{in particular for extreme floods} \cite{garambois2015parameter,jay2024spatially}.

The simultaneous regionalization approach involves optimizing a transfer function between physical descriptors and model parameters (cf. \citeA{hundecha2004modeling,gotzinger2007comparison,BASTOLA2008188,samaniego2010multiscale}). In this case and contrarily to post-regionalization, the descriptors-to-parameters mapping is the first trainable operator of the forward hydrological model. \rev{This approach} enables overcoming most of the aforementioned problems and has been applied in several studies. For instance, it has been used for regionalizing semi-distributed models such as HBV in \citeA{hundecha2004modeling} or in \citeA{gotzinger2007comparison} who introduced monotonicity and Lipschitz condition into the optimization problem to constrain the inferred spatial fields. %; TOPMODEL in \citeA{BASTOLA2008188} who used an Artificial Neural Network (ANN)-based mapping between catchment descriptors and model parameters (both their value and their uncertainty as quantified by a posterior covariance matrix). 
\rev{A multiscale parameter regionalization (MPR) method, combining descriptors maps, spatial upscalings functions and regionalization transfer functions in the form of multivariate mappings from descriptors, has been proposed by \citeA{samaniego2010multiscale}. This method is implemented within a spatially distributed multiscale hydrological model (mHM) and later applied to over 400 European catchments at $0.25^{\circ}$ spatial resolution in \citeA{Rakovec_2016}. The MPR approach, providing consistent (seamless) parameter and flux fields across scales \cite{Samaniego_hess-2017}, imposes a spatial regularization effect through a strong constraint in the forward model (upscaling laws and regionalization transfer functions). Such a regularization is needed when working with spatially distributed hydrological models and spatially sparse discharge data leading to overparameterized optimization problems. Likewise, \citeA{de2019regularization} discussed regularization and calibration from nested gauges for semi-lumped models and proposed a sequential optimization strategy from upstream sub-basins to downstream basins with upstream parameters relaxation. In the case of a fully distributed model calibrated with a variational data assimilation algorithm, overparameterization issues are typically addressed using classical regularization with a background term \cite{jay2020potential} or with a physiographic regularization term \cite{jay2024spatially} in the cost function. This approach induces weak constraints on the optimization problem and is not sufficient for effective spatial extrapolation to ungauged basins.}

%See discussion on regularization and calibration approach for} semi-lumped model from nested gauges \rev{considering sequential optimizations from upstream sub-basins to downstream with upstream parameters relaxation in \citeA{de2019regularization}, also for overparameterization issues in fully distributed model calibration with a variational data assimilation algorithm and classical regularization with a background term \cite{jay2020potential} or with a physiographic regularization term \cite{jay2024spatially} in the cost function, i.e., weak constraints of the optimization problem insufficient for effective spatial extrapolation to ungauged basins}. 

%\textbf{\\
%NOTE BR: La phrase ci-dessus est trop longue et rédigée comme une suite de bullet points, ci-dessous une proposition à modifier comme vous le sentez:\\
%Likewise, \citeA{de2019regularization} discussed regularization and calibration from nested gauges for semi-lumped models and proposed a sequential optimization strategy from upstream sub-basins to downstream basins with upstream parameters relaxation. In the case of a fully distributed model calibrated with a variational data assimilation algorithm, overparameterization issues are typically addressed using classical regularization with a background term \cite{jay2020potential} or with a physiographic regularization term \cite{jay2024spatially} in the cost function. This approach induces weak constraints on the optimization problem and is not sufficient for effective spatial extrapolation to ungauged basins.
%\\}

The MPR method of \citeA{samaniego2010multiscale} has also been used with other gridded hydrological models in large sample applications. For example, \citeA{mizukami2017towards} calibrate the VIC model at a resolution of $0.125^{\circ}$ over 531 headwater catchments ($\mathrm{area}< 2,000~\mathrm{km}^2$) in the continguous US area, using a lumped regionalization approach. Another example is \citeA{beck2020global}, who calibrate the HBV model at $0.05^{\circ}$ resolution over 4,229 headwater catchments ($\mathrm{area} < 5,000~\mathrm{km}^2$) worldwide. In their study, they categorize the catchments into three climatic groups and perform tenfold cross-validation using $90\%$ of the gauged catchments. While these studies applied MPR deterministically, in \citeA{lane2021incorporating}, the MPR method is applied within the generalized likelihood uncertainty estimation (GLUE) framework, with a high-resolution HRU model (DECIPHeR framework \rev{proposed by \citeA{coxon2019}}) at daily time resolution over a large sample of 437 catchments in the UK. However, the routing module in this study is calibrated separately with a simple random sampling approach. In \citeA{mizukami2017towards}, the runoff routing model is a gamma distribution function with two parameters that are ``directly calibrated for each basin''. The same remark can be made for \citeA{beck2020global}, \rev{with HBV discharge modeling at a daily time step on headwater catchments without any routing modeling.} Therefore, those regionalization \rev{studies} focus on \rev{lumped rainfall-runoff modeling at a daily time step for mostly headwater catchments whose characteristic response time scale might be shorter, or on more complex spatially distributed land surface modeling (LSM, including soil moisture and evaporation modeling in addition to river discharge) applied at regional or country scale still at a daily time step (mHM, e.g., \citeA{Boieng-hess-26-5137-2022}, or VIC)}.

In all the above studies, state of the art optimization \rev{or sampling} algorithms are used, especially the Shuffle Complex Evolution algorithm (SCE) \cite{Duan1992_SCE} in \citeA{mizukami2017towards}, or the Distributed Evolutionary Algorithms (DEAP) \cite{fortin2012_DEAP} in \citeA{beck2020global}, or the GLUE framework with a random sampling approach in \citeA{lane2021incorporating}. Those algorithms are applicable with low-dimensional controls only, which limits the affordable number of descriptors and the spatialization of regional transfer parameters (that are lumped in all methods above), and more importantly, which limits the affordable complexity of the regionalization operator. Nevertheless, gradient-based algorithms are efficient approaches for solving high dimensional inverse problems, and their potential has been demonstrated in optimization of spatial parameters of hydraulic models \cite{MONNIER2016}, or spatially distributed hydrological models \cite{Castaings-2009,jay2020potential, colleoni2022adjoint}. They crucially need accurate estimates of cost gradients, i.e., gradients of the cost (objective) function with respect to the sought parameters, which can be spatialized and of large dimension. Such gradients can be  computed with an adjoint model for example obtained by source code differentiation in \citeA{Castaings-2009,MONNIER2016,jay2020potential,colleoni2022adjoint}. Note that a key property of neural networks is their differentiability, which makes them compatible with variational data assimilation frameworks \rev{based on differentiable models} \cite{MONNIER2016}\rev{. Moreover, the intrinsic capability of neural networks to extract multi-level information from large dataset \cite{LeCun2015} and to perform non-linear multivariate regressions (e.g., \citeA{gemperline1991nonlinear}) makes} them promising candidates to \rev{learn effective non-linear descriptors-to-parameters} regionalization functions \rev{for spatially distributed differentiable hydrological modeling}. %\rev{The combination of NN based regionalization transfer function within a spatially distributed differential hydrological model and variational data assimilation framework has seldom been done and is relevant to extract information from discharge time series from multiple gauges and from physical descriptors maps to provide effective regional maps of model parameters  applicable to gauged and ungauged catchments.}
%In the context of \rev{spatially distributed} hydrological modeling of ungauged basins, achieving such a flexible regionalization is desirable to adequately represent the multi-scale variabilities of the physical system. It would also maximize the extraction of information from large sets of physical descriptors and hydrological response observations, while accounting for data and modeling uncertainties.

A novel approach called HDA-PR (Hybrid Data Assimilation and Parameter Regionalization) is presented in this article. \rev{The word ``hybrid'' refers to the incorporation of machine learning methods into a deterministic hydrological model for parameter regionalization}. HDA-PR relies on seamless regional optimization algorithms for learning complex transfer functions between physical descriptors and conceptual parameters of spatially distributed hydrological models, applicable at high-resolution with spatial constraints of various rigidity to address the spatial equifinality issue. It is designed to exploit the informative content of massive heterogeneous datasets over large spatio-temporal computational domains, and is therefore adapted to solving high-dimensional inverse problems. Our approach leverages information from multi-site river flow observations and high-resolution data on a 1~$\mathrm{km}^2$ and 1~$\mathrm{h}$ resolution grid, relying on the original combination of the following ingredients:
\begin{itemize}
    \item Learnable regionalization functions via the introduction into the direct hydrological model of an explicit tunable mapping between heterogeneous physical descriptors and spatially distributed conceptual parameters. This mapping allows estimating parameter values while imposing a constraint on their spatial variability, via the use of physical descriptors and a priori knowledge. Multivariate polynomial regressions and neural networks are employed to learn such a complex non-linear descriptors-to-parameters mapping.
    \item A differentiable spatially distributed hydrological model into which the regionalization operators have been implemented. This enables the computation of accurate, spatially-distributed gradients of the calibration cost (objective) function, with respect to the sought regionalization parameters, which can be of high dimension. Obtaining accurate gradients for such high-dimensional parameters is crucially needed for optimization algorithms.
\end{itemize}

The original combination of the above ingredients amounts to introducing regionalization transfer functions into a variational data assimilation (VDA) algorithm (similar to the tunable differentiable mappings in hydraulic VDA algorithms \cite{MONNIER2016,GARAMBOIS2020}) dedicated to spatially distributed hydrological modeling and high-dimensional inverse problems. This has seldom been investigated especially for regional hydrological learning from multi-site data. The strength of HDA-PR lies in its capability to learn complex relations between physical descriptors and conceptual parameters of spatially distributed models in the context of structural and spatial parametric equifinality.
Additionally, our approach aims at ensuring that the hybrid data assimilation algorithm, which integrates an explainable learning process, produces results that can be physically interpreted \cite{larnier2020hybrid, hoge2022improving, fablet2021learning, althoff2021addressing}. It is able to enhance calibration scores with deep learning from large heterogeneous datasets while maintaining their physical interpretability.

\rev{The research questions investigated in this article are the following: 
\begin{itemize}
    \item Does the embedding of neural network-based regionalization transfer functions within a spatially distributed differentiable hydrological model enable using variational data assimilation to extract relevant information from physical descriptors and discharge time series at multiple gauges? And does it provide effective spatial constraints to avoid spatial overparameterization in the inverse problem, leading to effective and interpretable regional maps of conceptual hydrological parameters for modeling gauged and ungauged catchments?
    \item How do neural networks compare, in terms of modeling performances in calibration and spatio-temporal extrapolation, to a simple multi-gauge calibration approach with spatially uniform model parameters or to a linear multivariate regression on physical descriptors, i.e., transfer functions of an equivalent complexity to the one used by, for example, \citeA{beck2020global}?% to those of MPR without upscaling laws? 
\end{itemize}
}

\rev{To assess the proposed HDA-PR approach and to study the research questions above, the} evaluation procedure adopted in this work considers \rev{challenging regionalization problems over a relatively large set of 126 flash-flood prone catchments in the French Mediterranean region, for an hourly and kilometric distributed hydrological model. It is a high-resolution hydrological modeling problem compared to existing studies with daily rainfall-runoff or LSM models for regional to continental coverage. Moreover, it is a difficult case because of very sudden and non-linear hydrological responses with significant spatial variabilities (e.g., \citeA{garambois2015parameter}). Performances are assessed by means of }%challenging regionalization problems with multi-gauge settings in flash-flood-prone areas 
multiple evaluation metrics including flood event hydrological signatures \cite{HUYNH2023signatures}. 
We address the following aspects of the HDA-PR approach: (i) performance at gauged and ungauged sites \rev{during calibration and validation time periods}; (ii) factors determining the performance; and (iii) spatial patterns of the regionalized parameters in relation to information extraction from physical descriptors.

The remaining sections of this paper are organized as follows: section~\ref{sec:Forward-inverse algorithms} describes the HDA-PR algorithms and the SMASH spatially distributed hydrological assimilation platform into which they have been implemented. In section~\ref{sec:Data-Num-exp}, we present the case studies based on two \rev{calibration setups} and analyze the performance of HDA-PR using different regionalization mappings. Subsequently, in section~\ref{sec:discuss}, we discuss compelling findings based on the results from the previous section. Finally, in section~\ref{sec:Conclusion}, we conclude our work and outline potential future research directions.

\section{Forward-Inverse Algorithms}\label{sec:Forward-inverse algorithms}

\begin{figure}[ht!]
 \noindent\includegraphics[width=\textwidth]{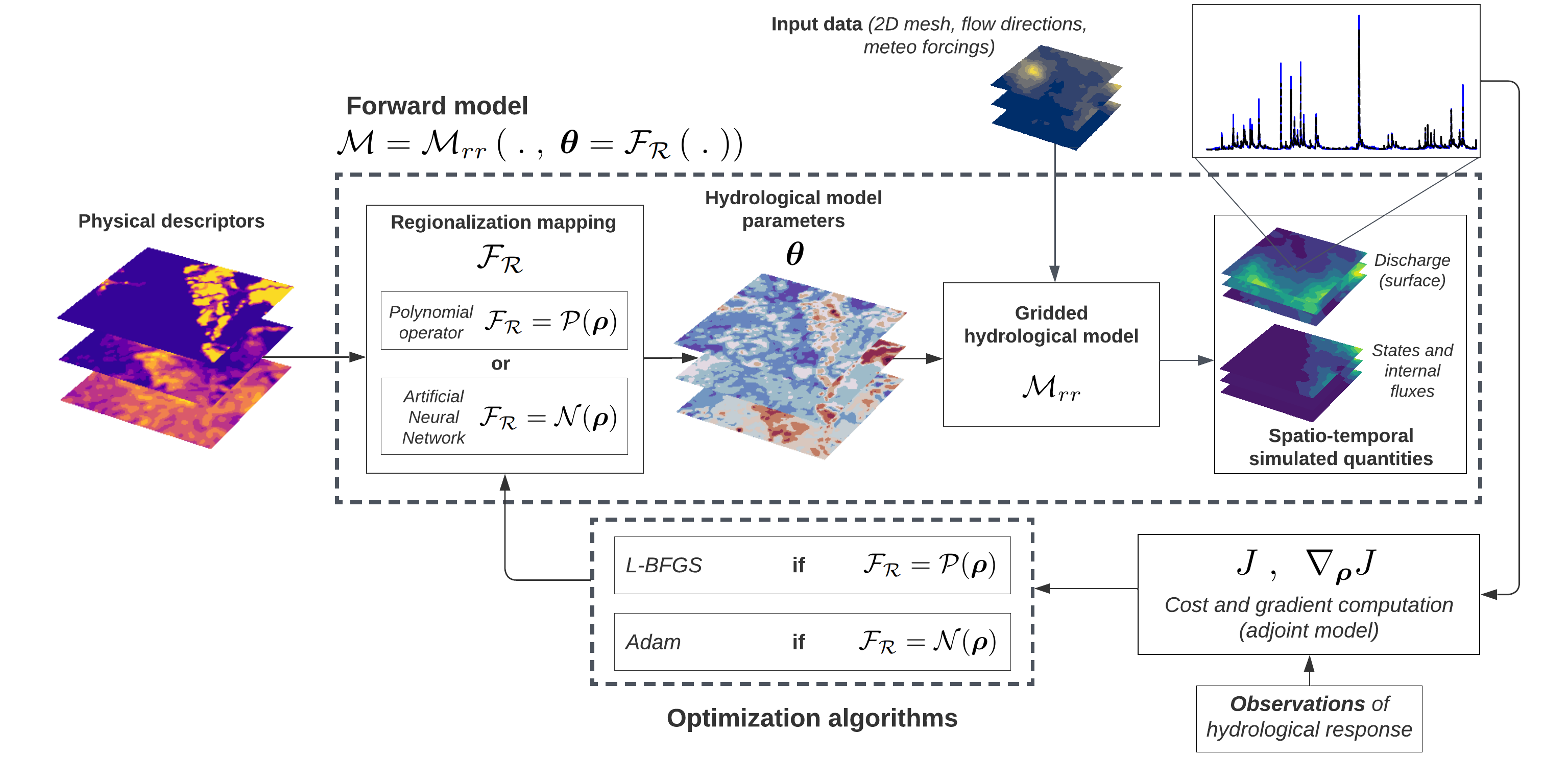}
 \caption{Flowchart of the forward-inverse algorithm used in HDA-PR. The forward hydrological model is a gridded model (spatio-temporal regular grid at 1~$\mathrm{km}^2$ and 1~$\mathrm{h}$) using GR operators \cite{perrin2003improvement}.}
\label{fig:flowchart_regio}
\end{figure}

This section presents the forward model and inverse algorithms of the proposed HDA-PR method. An algorithm flowchart is provided in Figure~\ref{fig:flowchart_regio} to help in global understanding.

First, the differentiable forward model consists in: (i) a parsimonious and robust \rev{spatially distributed} GR-like  conceptual hydrological model structure, \rev{composed of GR hydrological operators (``Génie Rural'' lumped models cf.  \citeA{perrin2003improvement}) applied at pixel scale for runoff generation, and a simple pixels-to-pixel routing scheme, the whole hydrological model being differentiable \cite{jay2020potential, colleoni2022adjoint}}; and (ii) regionalization operators, consisting of either multivariate polynomial regressions or neural networks, for mapping descriptors onto hydrological model parameters.
The calibration cost function adapted to multi-site (and potentially multi-source) observations is then defined. 
\rev{The inverse optimization algorithms, that use the spatially distributed gradients of the cost function with respect to model parameters, are capable of dealing with high-dimensional inverse problems such as encountered with tunable parameters of regionalization descriptors-to-parameters mappings.} 

The core strength of HDA-PR is the use of differentiable descriptors-to-parameters transfer functions, especially in the form of neural networks, and the capability to automatically compute accurate cost gradients. The latter enables the use of gradient-based variational optimization algorithms with high-dimensional regional parameter vectors. The method is applicable to any differentiable forward model as well as to multi-source heterogeneous datasets, and hence constitutes a powerful data assimilation framework.

\subsection{Forward Model with Regionalization}\label{sub:forward_model_preregio}

First, let $\Omega\subset\mathbb{R}^2$ denote a 2D spatial domain   that can contain multiple catchments, both gauged and ungauged, with a minimum of one gauged catchment, and $t>0$ the physical time. In what follows, the vector of spatial coordinates over $\Omega$ is denoted $x$. The number of active cells within the spatial domain $\Omega$ is noted $N_{x}$.
A 2D flow directions map $\mathcal{D}_{\Omega}$ is obtained from terrain elevation processing and will be used for runoff routing, with the only condition that a unique point in the regular mesh $\mathcal{T}_{\Omega}$ has the highest drainage area.

Consider observed discharge time series $\boldsymbol{Q}^*_g(t)$ at $N_{G}$ observation cells of coordinates $x_{g}\in \Omega$, $g=1..N_{G}$ ($N_{G} \geq 1$).
For each observation cell, the corresponding gauged upstream sub-catchment is noted $\Omega_{g}$ so that $\Omega_{ung} = \Omega \setminus \left( \cup_{g=1}^{N_{G}}\Omega_{g} \right)$ is the remaining ungauged part of the whole spatial domain $\Omega$. Note that this definition is suitable for the general regionalization case dealing with spatially independent and/or nested gauged catchments.

Then, the forward model $\mathcal{M}$ can be defined as a multivariate function obtained by partially composing a hydrological model $\mathcal{M}_{rr}$ with a regionalization operator $\mathcal{F}_{R}$ to compute hydrological parameters $\boldsymbol{\theta}$ such that:
\begin{equation}\label{eq:forward model}
\mathcal{M}=\mathcal{M}_{rr}\left(\;.\;,\;\boldsymbol{\theta}=\mathcal{F_{R}}\left(\;.\;\right)\right)
\end{equation}
Let us now introduce and detail the hydrological model and the regionalization operator along with their input variables.

The rainfall and potential evapotranspiration fields are respectively noted $\boldsymbol{P}\left(x,t\right)$ and $\boldsymbol{E}\left(x,t\right)$, $\forall x\in \Omega$.
The hydrological model $\mathcal{M}_{rr}$ is a dynamic operator projecting the input fields $\boldsymbol{P}\left(x,t\right)$ and $\boldsymbol{E}\left(x,t\right)$, given an input drainage plan $\mathcal{D}_{\Omega}\left(x\right)$, onto \rev{an output field $\boldsymbol{U}(x,t)$, which comprises the discharge field $\boldsymbol{Q}\left(x,t\right)$ and state fields $\boldsymbol{h}\left(x,t\right)$. This is such that, for all $(x, t') \in \Omega \times \left[0,t\right]$}:
\begin{equation}\label{eq:forward hydrological model}
\boldsymbol{U}\left(x,t\right)=\left(\boldsymbol{h},\boldsymbol{Q}\right)\left(x,t\right)=\mathcal{M}_{rr}\left[ (\mathcal{D}_{\Omega}, \boldsymbol{\theta})(x); (\boldsymbol{P},\boldsymbol{E})(x,t'),\boldsymbol{h}(x,0), t \right]
\end{equation}
where $\boldsymbol{\theta}$ is the $N_{\theta}$-dimensional vector of model parameters 2D fields that we aim to estimate regionally with the new algorithms proposed below, and $\boldsymbol{h}$ is the $N_{S}$-dimensional vector of internal model states. 
In this study, the distributed hydrological model $\mathcal{M}_{rr}$ is a parsimonious GR4-like conceptual structure \cite{perrin2003improvement}, which is the spatialized ``S-GR4'' structure presented in the documentation of SMASH \cite{colleoni_2024_10991739}. The hydrological parameters vector $\forall x\in \Omega$ is:
\begin{equation*}
    \boldsymbol{\theta}\left(x\right)= \left( c_p (x), c_{t} (x), k_{exc} (x), l_{l_r} (x) \right) ^T
\end{equation*}
where the four spatially varying parameter fields are the capacity of the production reservoir ($c_p$ in [mm]), the capacity of the transfer reservoir ($c_{t}$ in [mm]), the parameter ($k_{exc}$ in [mm/dt]) of the non-conservative water exchange flux, and the linear routing parameter ($l_{l_r}$ in [min]).

In order to constrain and explain these spatial fields of conceptual model parameters $\boldsymbol{\theta}(x)$ from descriptors $\boldsymbol{D}(x)$, we introduce a regionalization operator $\mathcal{F}_{R}$ that is a descriptors-to-parameters mapping  such that:
\begin{equation}\label{eq:regio general eq}
    \boldsymbol{\theta}\left(x\right)=\mathcal{F}_{R}(\boldsymbol{D}\left(x\right),\boldsymbol{\rho}),\,\forall x \in \Omega
\end{equation}
with $\boldsymbol{D}$ the $N_{D}$-dimensional vector of physical descriptor maps covering $\Omega$, and $\boldsymbol{\rho}$ the vector of tunable regionalization parameters that is defined below. 

Two types of regionalization operators are used in HDA-PR (see Figure~\ref{fig:flowchart_regio}):

\begin{enumerate}
   \item A set $\mathcal{P}$ of multivariate polynomial regression operators for each parameter of the forward hydrological model (Equation~\ref{eq:forward hydrological model}):
   \begin{equation}\label{eq:prereg_lphp}
   \begin{aligned}
       \boldsymbol{\theta}(x,\boldsymbol{D},\boldsymbol{\rho}) \coloneqq \mathcal{P}\left(\boldsymbol{D}(x), \boldsymbol{\rho}\right) \equiv \left[
           \left(\theta_{k}(x,\boldsymbol{D},\boldsymbol{\rho}_k)\right)_{k=1}^{N_\theta}
       \right]^T,\, \forall x \in \Omega; \\
       \theta_{k}(x,\boldsymbol{D},\boldsymbol{\rho}_k)\coloneqq s_{k}\left(\alpha_{k,0}+\sum_{d=1}^{N_{D}}\alpha_{k,d}D_{d}^{\beta_{k,d}}(x)\right),\,\forall k \in[1..N_{\theta}] 
   \end{aligned}
   \end{equation}
    with $s_{k}(z)=l_{k}+(u_{k}-l_{k})/\left(1+e^{- z}\right),\,\forall z\in\mathbb{R}$,
    a transformation based on a Sigmoid function with values in $\left]l_k,u_k\right[$, thus imposing bound constraints in the direct hydrological model such that  $l_{k}<\theta_{k}(x)<u_{k},\,\forall x\in\Omega$. The lower and upper bounds $l_{k}$ and $u_{k}$, associated to each parameter field $\theta_{k}$ of the hydrological model (Equation~\ref{eq:forward hydrological model}) are assumed spatially uniform for simplicity here. 
    The regional control vector to be estimated in this case is: 
    \begin{equation}\label{eq:control_lphp}
    \boldsymbol{\rho} \equiv \left[\left(\boldsymbol{\rho}_k\right)_{k=1}^{N_\theta}\right]^{T} \equiv \left[\left(\alpha_{k,0},\left(\alpha_{k,d},\beta_{k,d}\right)_{d=1}^{N_D}\right)_{k=1}^{N_\theta}\right]^{T} 
   \end{equation}

   \item An ANN denoted $\mathcal{N}$, consisting of a multilayer perceptron, aimed at learning the descriptors-to-parameters mapping 
   such that:
   \begin{equation}\label{eq:neural_net}
       \boldsymbol{\theta}(x,\boldsymbol{D},\boldsymbol{\rho})\coloneqq \mathcal{N}\left(\boldsymbol{D}(x), \boldsymbol{W}, \boldsymbol{b}\right),\forall x \in \Omega
   \end{equation}
   where $\boldsymbol{W}$ and $\boldsymbol{b}$ are respectively weights and biases of the neural network composed of $N_L$ dense layers. The architecture of the neural network and the forward propagation is detailed in Appendix~\ref{appd:ann} and Equation~\ref{eq:forward_propa}.
   Note that an output layer consisting of a scaling transformation based on the Sigmoid function (cf. Equation~\ref{app:eq:scaling}) enables the imposition of bound constraints on the $k^{th}$-hydrological parameters, i.e., $l_{k}<\theta_{k}(x)<u_{k},\,\forall x\in\Omega$.
   The regional control vector in this case is: 
   \begin{equation}\label{eq:control_net}
       \boldsymbol{\rho} \equiv \left[\boldsymbol{W}, \boldsymbol{b}\right]^T \equiv \left[\left(\boldsymbol{W}_j, \boldsymbol{b}_j\right)_{j=1}^{N_L}\right]^T
   \end{equation} 
\end{enumerate}

For each regionalization operator (Equation~\ref{eq:prereg_lphp} or \ref{eq:neural_net}), the regional calibration problem consists in optimizing (in a sense defined below) the regionalization control $\boldsymbol{\rho}$ (Equation~\ref{eq:control_lphp} or \ref{eq:control_net}) that can be of relatively high dimension since it is proportional to the number of descriptors ($N_D$), the number of model parameters ($N_{\theta}$), and the degree of spatialization of the regional controls.
Optimization algorithms adapted to the high-dimensional problems of interest, taking advantage of accurate spatially distributed gradients computation with the adjoint of the forward model, are detailed thereafter. Importantly, note that by definition of the mathematical model \rev{(differentiable hydrological model, cf. \citeA{jay2020potential}, combined with embedded neural networks or polynomial regionalization functions which are differentiable)} and  given the numerical implementation rules followed, the forward \rev{numerical} model is differentiable. This is a necessary condition for computing cost gradients with respect to spatially distributed hydrological parameters and obtain those of regional controls, as needed for solving the optimization problem. This is a key idea and property of our proposed algorithms.

The numerical resolution of the ordinary differential equation (ODE)-based operator of the forward hydrological model (Equation~\ref{eq:forward hydrological model}) relies on an explicit expression of its solution, approximated on the regular mesh $\mathcal{T}_{\Omega}$  of constant step $dx$ with a fixed time step $dt$. All physical descriptors are mapped onto model grid for simplicity here.

Note that adding an upscaling operator after the regionalization scheme (as done in \citeA{samaniego2010multiscale}) is feasible in HDA-PR under the condition that it is differentiable (at least numerically), and is a potentially interesting topic for further research, as is improving observation operators. In both cases one could use algebraic expressions or neural networks in the HDA-PR assimilation framework.

\subsection{Calibration Cost Function}

A calibration cost function is defined to measure the misfit between simulated and observed discharge time series, respectively noted $Q_{g}(t)$ and $Q_{g}^{*}(t)$, for $g\in 1.. N_{G}$ gauged cells. In order to measure the discrepancy between observed and simulated quantities from multiple observation sites, we consider the cost function:
\begin{equation}\label{eq:cost-definition}
    J=\sum_{g=1}^{N_{G}}w_{g}J_{g}^{*}
\end{equation}
with $w_{g}$ a weighting function explained afterwards, $J_{g}^{*}$ a local quadratic metric ``at the station'', here $1-NSE$ or $1-KGE_{2}$ \rev{(the latter being only used in validation in the case study of section~\ref{sec:Data-Num-exp},} see Appendix~\ref{appd:metrics}). This \rev{NSE-based calibration} cost function is \rev{a quadratic} differentiable and convex function, involving the response of the direct model. It depends on the control vector $\boldsymbol{\rho}$ through the direct model $\mathcal{M}$ (Equation~\ref{eq:forward model}) composed of the regionalization operator $\mathcal{F}_{R}$ (Equation~\ref{eq:regio general eq}) and the direct hydrological model $\mathcal{M}_{rr}$ (Equation~\ref{eq:forward hydrological model}).

The multi-site calibration corresponds to $N_{G}>1$ while $N_{G}=1$ is the classical single-gauge calibration. For $N_{G}>1$, the weighting $w_{g}$ is defined such that $\sum_{g=1}^{N_{G}}w_{g}=1$.

\subsection{\rev{The Variationnal Data Assimilation Algorithm}}\label{subsec:inverse-algo}

\rev{The VDA aims at estimating the unknown input parameter $\boldsymbol{\rho}$ of the descriptors-to-parameters transfer functions $\mathcal{F}_R$ embedded into the hydrological model $\mathcal{M}_{rr}$ and predicting hydrological parameters $\boldsymbol{\theta}(x),\forall x\in\Omega$ from physical desciptors maps $\boldsymbol{D}(x)$, by minimizing the discrepancy between the modeled and observed discharges at multiple gauges. The cost function $J\left(\boldsymbol{U}\left(\boldsymbol{\rho}\right)\right)$ to optimize depends on the regional control $\boldsymbol{\rho}$ through the response $\ensuremath{\boldsymbol{U}\left(\boldsymbol{\rho}\right)=\mathcal{M}_{rr}\left(.,\;\boldsymbol{\theta}=\mathcal{F}_{R}\left(\boldsymbol{\rho}\right)\right)}$ of the forward model that combines the hydrological model $\mathcal{M}_{rr}$ and the regionalization operator $\mathcal{F}_R$.}
The \rev{VDA} inverse problem is written as the following convex optimization problem of the control vector $\boldsymbol{\rho}$:
\begin{equation}\label{eq:general inv pb}
\boldsymbol{\hat{\rho}}=\arg\min_{\boldsymbol{\rho}}J\left(\boldsymbol{U}\left(\boldsymbol{\rho}\right)\right)
\end{equation}
The calibration \rev{of this regional control $\boldsymbol{\rho}$} aims to (i) reduce the misfit between observed and simulated discharges at spatially sparse gauging stations, as evaluated by Equation~\ref{eq:cost-definition}, while (ii) determining the hydrological parameter maps $\boldsymbol{\theta}(x)$ for discharge modeling at any ungauged sites, \rev{thereby benefiting from the information extracted from physical descriptors $\boldsymbol{D}(x)$, and spatial constraints induced by the regional transfer functions, whose parameters $\boldsymbol{\rho}$ are being optimized.} 
The regionalization operator can be expressed as either (i) a multi-polynomial mapping $\mathcal{F}_R \equiv \mathcal{P}$ (Equation~\ref{eq:prereg_lphp}), or (ii) an artificial neural network $\mathcal{F}_R \equiv \mathcal{N}$ (Equation~\ref{eq:neural_net}). \rev{For both formulations of the regionalization operator}, the regional control vector $\boldsymbol{\rho}$ to optimize is large, and gradient-based optimization methods adapted to high-dimensional inverse problems are employed \rev{and detailed thereafter}.

\rev{Note that the inverse problem \ref{eq:general inv pb} is a variational data assimilation optimization problem in the sense that it seeks to optimally combines observations (here discharge series $Q^*_{g=1..N_G}$ at $N_G$ multiple gauges) with a model (here of Equation~\ref{eq:forward model}). While the NSE-based cost function defined earlier is not built upon explicit probabilistic assumptions, it can be shown \cite{kavetski2006bayesian} that the minimization in Equation~\ref{eq:general inv pb} is equivalent to maximizing the posterior density $p\left(\boldsymbol{\rho}|\boldsymbol{Q}^{*}\right)$ of the parameter $ \boldsymbol{\rho}$ given observations $\boldsymbol{Q}^{*}$ under the assumptions of independent and identically distributed Gaussian errors at each gauge, spatially independent errors and no prior information. This probabilistic interpretation highlights a set of assumptions that could certainly be improved upon, and opens the way for more advanced probabilistic models that would explicitly recognize the various sources of uncertainty affecting the model and the surrounding data (forcings and responses). This is further discussed in section~\ref{sec:discuss}.}
%We do not consider uncertainty on model atmospheric forcings, physical descriptors or discharge time series which represents an interesting subject for further research.}

\subsubsection{Optimization Algorithm for Polynomial Regionalization}\label{deterministic_regio}

In this case, the forward model includes the polynomial descriptors-to-parameters mapping (Equation~\ref{eq:prereg_lphp}), i.e., $\mathcal{F}_R \equiv \mathcal{P}$ and the regional control vector is:
\begin{equation*}
    \boldsymbol{\rho} \coloneqq \left[\alpha_{k,0},\left(\alpha_{k,d},\beta_{k,d}\right)\right]^{T},\forall(k,d)\in[1..N_{\theta}]\times[1..N_{D}]
\end{equation*}

The optimization problem, represented in Equation~\ref{eq:general inv pb}, is solved using the L-BFGS-B algorithm (limited-memory Broyden–Fletcher–Goldfarb–Shanno bound-constrained) \cite{Zhu1997}. This algorithm is specially adapted to the high-dimensional parameter space, and in this study, there are no bound constraints on the values of $\alpha_{k,.}$, whereas the exponents $\beta_{k,d}$ are simply sought between 0.5 and 2. This algorithm requires the gradient of the cost function with respect to the sought parameters $\nabla_{\boldsymbol{\rho}} J $. This gradient is computed by \rev{a single run of} the adjoint model, which is obtained by automatic differentiation using the Tapenade engine \cite{hascoet2013tapenade}. The entire process is implemented in the SMASH Fortran source code, where the full forward model $\mathcal{M}\equiv \mathcal{M}_{rr}\left(.,\mathcal{P}\left(.\right)\right)$ is a composition of both the hydrological model and the polynomial descriptors-to-parameters mapping.

The background value $\boldsymbol{\rho}^{*}$, used as a starting point for the optimization, is set using a spatially uniform solution $\bar{\boldsymbol{\theta}}^*$, which is obtained by a simple global optimization algorithm \cite{Michel1989} of the inverse problem (Equation~\ref{eq:general inv pb}) where $\mathcal{M}\equiv \mathcal{M}_{rr}$ and $\boldsymbol{\rho} \coloneqq\bar{\boldsymbol{\theta}}$, as follows:
\begin{equation*}
    \boldsymbol{\rho}^*\equiv\left[ \alpha_{k,0} = s^{-1}_{k}\left( \bar{\theta_{k}}^* \right),\left( \alpha_{k,d} = 0, \beta_{k,d} = 1\right)\right]^{T},\forall(k,d)\in[1..N_{\theta}]\times[1..N_{D}]
\end{equation*}
where $s^{-1}_{k}(z)= \ln\left(\frac{z-l_{k}}{u_{k}-z}\right)$ is the inverse Sigmoid.

The termination criterion is determined based on the satisfaction of at least one of the following criteria:

\begin{itemize}
    \item Maximum number of iterations;
    \item Cost function criterion: $\frac{J^{(i)} - J^{(i+1)}}{\max \left(\lvert{J^{(i)}} \rvert, \lvert{J^{(i+1)}}\rvert, 1 \right)} \leq \epsilon \times 10^{6}$ (e.g., $\epsilon \approx 2.22\times10^{-16}$);
    \item Gradient criterion: $||\nabla_{\boldsymbol{\rho}} J^{(i)}||_{\infty} \leq 10^{-12}$
\end{itemize}
where $J^{(i)}$, $||\nabla_{\boldsymbol{\rho}} J^{(i)}||_{\infty}$ are respectively the cost value and its projected gradient at iteration $i$, and $\epsilon$ represents the machine precision.

\subsubsection{Optimization Algorithm for Neural Network-based Regionalization}\label{ann_regio}

In this case, the forward model includes a descriptors-to-parameters mapping performed with a neural network, i.e., $\mathcal{F}_R \equiv \mathcal{N}$ and the regional control vector is $\boldsymbol{\rho} \coloneqq \left[\boldsymbol{W},\boldsymbol{b}\right]^T$. 
The optimization problem (Equation~\ref{eq:general inv pb}) can typically be solved using the Adam optimization algorithm \cite{kingma2014adam}, an efficient stochastic gradient descent algorithm able to adapt the learning rate based upon the first and the approximation of the second moments of the gradients for fast convergence, and only requiring the first order gradients of the cost function. 
In the present case, the cost function writes as:
\begin{equation}\label{eq:cost-ann}
    J\left(\boldsymbol{U}\left(\boldsymbol{\rho}\right)\right)=J\big(\boldsymbol{Q}^{*},\mathcal{M}_{rr}(.\; ,\; \boldsymbol{\theta}=\mathcal{N}(\boldsymbol{D},\boldsymbol{\rho}))\big)
\end{equation}

This formulation of the cost function highlights its dependency on the forward model $\mathcal{M}\equiv \mathcal{M}_{rr}\left(.,\mathcal{N}\left(.\right)\right)$, which is composed of two components in its numerical implementation: (i) an ANN implemented in Python, which produces the output $\boldsymbol{\theta}$ used as input by (ii) the hydrological model $\mathcal{M}_{rr}$ implemented in Fortran. In order to optimize $J$, its gradients with respect to $\boldsymbol{\rho}$ are required. The main technical difficulty here is to achieve a ``seamless flow of gradients'' through back-propagation. To overcome this, we divide the gradients into two parts and apply the chain rule with analytical derivation and numerical code differentiation (cf. hybrid VDA course in \citeA{monnier:hal-03040047} and references therein). First, $\nabla_{\boldsymbol{\theta}} J$ can be computed via the automatic differentiation applied to the Fortran code corresponding to $\mathcal{M}_{rr}$. Then, $\nabla_{\boldsymbol{\rho}} \boldsymbol{\theta}$ is simply obtained by analytical calculus applicable given the explicit architecture of the ANN, consisting of a multilayer perceptron. Finally, the two gradients can be combined as $\nabla_{\boldsymbol{\rho}} J = \nabla_{\boldsymbol{\theta}} J . \nabla_{\boldsymbol{\rho}} \boldsymbol{\theta}$. \rev{The background value $\boldsymbol{\rho}^*$ in this case is randomly initialized using a specific method, which will be discussed later.} The termination criterion is determined by a specified number of training epochs in the optimization algorithm. A detailed explanation of the network architecture, backward propagation, and the optimization process can be found in Appendix~\ref{appd:ann}.

\section{Data and Numerical Experiment} \label{sec:Data-Num-exp}
\subsection{Study Area and Experimental Design}

%The performance of HDA-PR \rev{in terms of discharge modeling but also in terms of its capability to extract information from data to estimate conceptual model parameters is analyzed over a relatively large catchment sample, and compared with the performance of other regionalization methods: a simple multi-gauge calibration approach for spatially uniform model parameters (Uniform) or a linear multivariate regression (Multi-linear) on physical descriptors, i.e., transfer functions of the same complexity as those of MPR but without upscaling laws.} 

The performance of HDA-PR \rev{in terms of discharge modeling but also in terms of its capability to extract information from data to estimate conceptual model parameters is analyzed over the French Mediterranean region. The performance of three regionalization methods is compared: a simple multi-gauge calibration approach for spatially uniform model parameters (Uniform), and two variants of HDA-PR. This first variant (Multi-linear) uses linear multivariate regression  on physical descriptors, i.e., transfer functions of the same complexity as those of \citeA{beck2020global}. The second variant (ANN) uses a multilayer perceptron, which corresponds to a more complex mapping.}

\begin{figure}[ht!]
 \noindent\includegraphics[width=\textwidth]{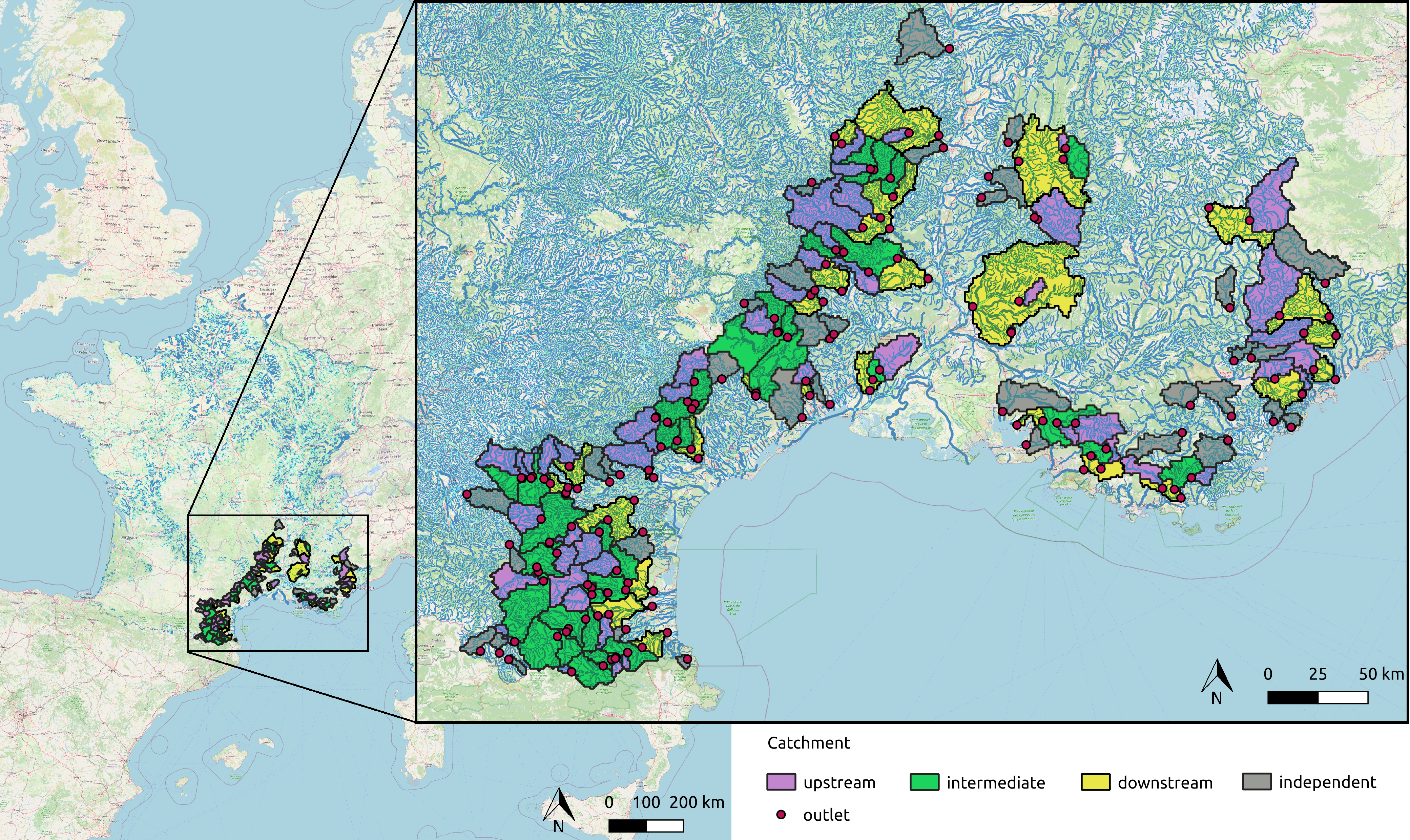}
 \caption{\rev{Map of France highlighting the ArcMed study area, covering 150,000~$\mathrm{km}^2$ (100,000~$\mathrm{km}^2$ excluding sea), comprising 126 catchments categorized as 38 catchments located upstream, 33 intermediate catchments, 24 catchments positioned downstream, and 31 independent catchments, representing a total drainage area of 26,000~$\mathrm{km^2}$.}}
 \label{fig:study-zones}
\end{figure}

The SMASH model is run on a $dx = 1~\mathrm{km}$ spatial grid at $dt = 1~\mathrm{h}$ time step. It is forced by: (i) observed rainfall grids based on hourly ANTILOPE J+1 radar-gauge rainfall reanalysis from Météo-France \cite{champeaux2009mesures}; (ii) potential evapotranspiration (PET) estimated using the formula of \citeA{oudin2005potential}; and (iii) temperature data from SAFRAN reanalysis produced by Météo-France on a $8\times8$~$\mathrm{km}^2$ spatial grid \cite{Quintana2008} downscaled to a $1\times1$~$\mathrm{km}^2$ spatial grid. \rev{The evaluation is performed on a flash-flood-prone area known as the Mediterranean arc (and called ArcMed hereafter), situated in the South of France, as indicated in Figure~\ref{fig:study-zones}. Covering an approximate land area of 100,000~$\mathrm{km}^2$, ArcMed comprises 126 flash-flood-prone catchments, including both nested and independent ones, representing about 26,000~$\mathrm{km}^2$ of combined drainage area. These catchments have been selected based on the availability of long time series with high-quality observed flow data and minimal anthropogenic impacts. ArcMed is known for its diverse hydrological properties and contrasted catchment behaviors, including steep topography and very heterogeneous soils and bedrock (e.g., \citeA{garambois2015parameter}). The area is affected by intense rainfall events that trigger non-linear flash flood responses, presenting a very challenging modeling scenario, in particular due to the significant presence of karstic zones. This evaluation dataset is quite extensive considering the high-resolution, which is made necessary by the fast and small-scale non-linear processes characterizing the flash floods occurring in the area. The resulting spatio-temporal computational domain is quite large ($>26,000 \text{ pixels} \times 35,000$ \text{ time steps}). By comparison, regionalization studies in the literature often make use of larger regions and more numerous catchments, but also of much coarser models typically running at a daily time step and coarser spatial resolution. %Such coarse models may not meaningful for flash-flood prone regions.
}

A set of 7 physical descriptors (Table~\ref{tab:descriptors}) available over the whole French territory is used following \citeA{Odry2017} and \citeA{jay2024spatially}. Note that this setup is sufficient to assess the regionalization performance of the proposed algorithms while keeping the present article concise. \rev{The issue of selecting the most relevant information for multi-source observations of hydrological responses and the most adequate descriptors layers is intentionally left for future research since it requires additional complementary modules to the proposed framework.} It is worth noting that prior to the optimization process, all descriptors are standardized between 0 and 1 through min-max scaling.

\begin{table}[ht!]
\caption{Descriptors used as input data for regionalization methods.}
\label{tab:descriptors}
%\scalebox{0.9}{
\begin{tabular}{c c >{\centering\arraybackslash}p{0.35\textwidth} c c}
    \hline
    Notation & Type & Description & Unit & Source \\
    \hline
     $d_1$ & Topography & Slope & $^{\circ}$ & \citeA{Odry2017} \\
     $d_2$ & Morphology & Drainage density & - & \citeA{organde2013regionalisation} \\
     $d_3$ & Influence & Percentage of basin area in karst zone & \% & \citeA{caruso2013notice} \\
     $d_4$ & Land use & Forest cover rate & \% & \rev{CORINE Land Cover} (2012)\\
     $d_5$ & Land use & Urban cover rate \rev{(including artificial and non-vegetated areas)} & \% & \rev{CORINE Land Cover} (2012)\\
     $d_6$ & Hydrogeology & Potential available water reserve & mm & \citeA{ponce2016} \\
     $d_7$ & Hydrogeology & High storage capacity basin rate & \% & \citeA{finke1998geo} \\
    \hline
\end{tabular}
%}
\end{table}

\rev{To evaluate the performance of HDA-PR, we employ two spatial cross-validation setups which involve (i) calibrating the model on 38 upstream gauges (upstream calibration) and (ii) calibrating the model on 24 downstream gauges (downstream calibration). The upstream calibration poses a greater challenge due to the smaller catchments' areas, resulting in a lower integrative effect on non-linear hydrological processes and potentially more hydrological variabilities within and between basins. Consequently, this setup represents a more demanding interpolation scenario compared to the downstream calibration, which involves larger basins, as demonstrated later with calibration results. These two calibration setups, along with a temporal validation scheme based on a two-period split, will facilitate the study of information extraction from discharge time series and from physical descriptors.
For each calibration setup, we apply the different multi-site regional calibration methods using the set of gauges from the calibration catchments, while the remaining gauges from the validation catchments are used for spatial and spatio-temporal validation purposes. The weighting in Equation~\ref{eq:cost-definition} can be set as $w_{g}=\frac{1}{N_{G}}$, representing the average cost over multiple gauges, since the gauges used for calibration in each setup, have the same number of observations (up to sporadic missing values), and share the same nature (either downstream or upstream), thus have similar information accumulation along the flow paths. The chosen calibration metric is the NSE, computed using data from multiple gauges over a four-year period P1 (August 2016 to July 2020) with a one-year warm-up period (August 2016 to July 2017). The following calibration methods are compared:
\begin{itemize}
    \item Local calibrations for each gauge, both with spatially uniform (i.e., $\boldsymbol{\rho} \equiv \bar{\boldsymbol{\theta}}$) and full spatially distributed controls (i.e., $\boldsymbol{\rho} \equiv \boldsymbol{\theta}\left(x\right)$), which are respectively under- and over-parameterized hydrological optimization problems. These approaches represent reference or benchmark performances, denoted as ``Uniform (loc)'' and ``Distributed (loc)''.
\item Multi-gauge regional calibration approaches with:
\begin{itemize}
    \item lumped model parameters (i.e., $\boldsymbol{\rho} \equiv \bar{\boldsymbol{\theta}}$), representing ``level 0'' regionalization, denoted as ``Uniform (reg)'';
    \item a multivariate linear mapping (i.e., $\boldsymbol{\rho} \equiv \left[\alpha_{k, 0}, (\alpha_{k, d}, 1)\right]^T$), referred to as ``Multi-linear (reg)'', which represents the classical regionalization mapping with the same complexity as the transfer functions of \citeA{beck2020global} and without upscaling laws;
    % \item a multivariate polynomial mapping (i.e., $\boldsymbol{\rho} \equiv \left[\alpha_{k, 0}, (\alpha_{k, d}, \beta_{k, d})\right]^T$), denoted ``Multi-polynomial (reg)'';
    \item a multilayer perceptron (i.e., $\boldsymbol{\rho} \equiv \left[\boldsymbol{W}, \boldsymbol{b}\right]^T$), denoted as ``ANN (reg)'', representing the core novelty of the HDA-PR framework.
\end{itemize}
\end{itemize}
}
\noindent \rev{Note that the multivariate linear mapping above is a particular case of the polynomial mapping presented in section~\ref{sub:forward_model_preregio}, obtained by forcing exponents $\beta_{k,d}$ to one. General polynomial mappings have been studied (see \citeA{huynh2023learning}) but are not further considered in this paper for the sake of conciseness.}
To ensure robust validation, model performances are assessed in terms of spatial, temporal, and spatio-temporal validations, with a temporal validation period P2 covering two years from \rev{August 2020 to July 2022}. Various evaluation metrics, including multiple hydrological signature-based metrics presented in \citeA{HUYNH2023signatures}, are also employed.

\subsection{Regional Learning Performance and Computational Efficiency}
\label{subsec:perf}

\begin{figure}[ht!]
 \noindent\includegraphics[width=\textwidth]{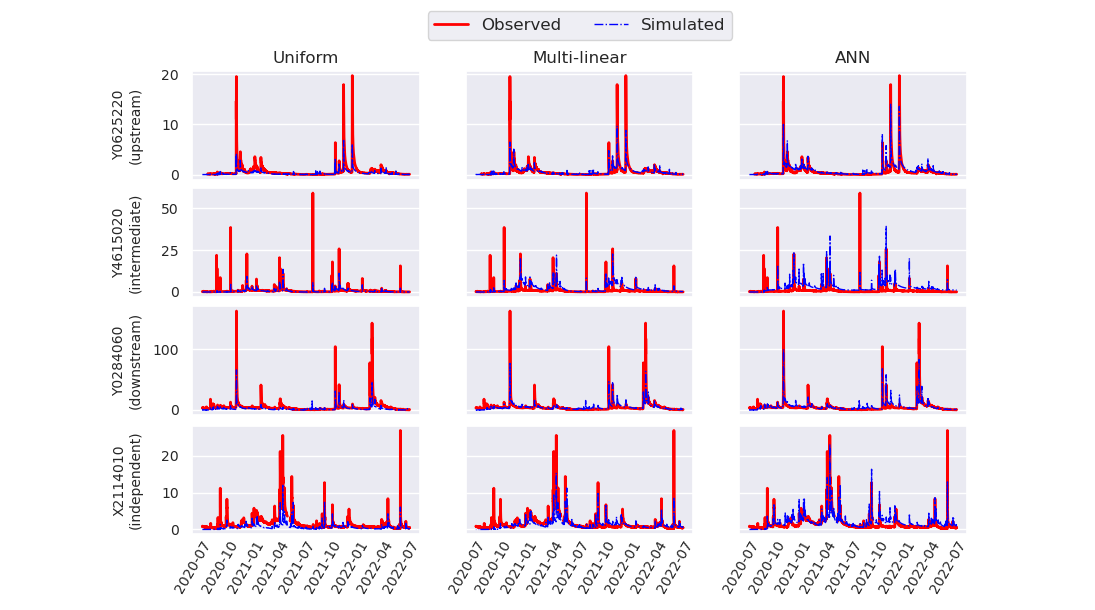}
 \caption{\rev{Observed and simulated discharges (in $\mathrm{m}^3/\mathrm{s}$) at several locations during the validation period P2, using three multi-gauge regional calibration methods (columns) within the first calibration setup, where upstream catchments are used for calibration on P1. The first row hence corresponds to temporal validation assessments, while the subsequent three rows correspond to spatio-temporal validation assessments.}}
 \label{fig:hydrogram-p1}
\end{figure}

In this section, the regional learning performance is analyzed \rev{over the whole ArcMed region. Before a thorough evaluation of performance is performed, Figure~\ref{fig:hydrogram-p1} provides several examples of observed and simulated discharges at randomly selected gauges, and compares them to observations. For these examples,} using lumped model parameters $\bar{\boldsymbol{\theta}}$ leads to a poor performance in simulating discharges. \rev{By contrast,} the other \rev{two} regional learning methods result in improved performance, in both calibration and validation catchments. Given the complexity and heterogeneity of the region, it is unsurprising that lumped model parameters regionalization is unable to accurately reproduce such contrasted hydrological responses.

\begin{figure}[ht!]
 \noindent\includegraphics[width=\textwidth]{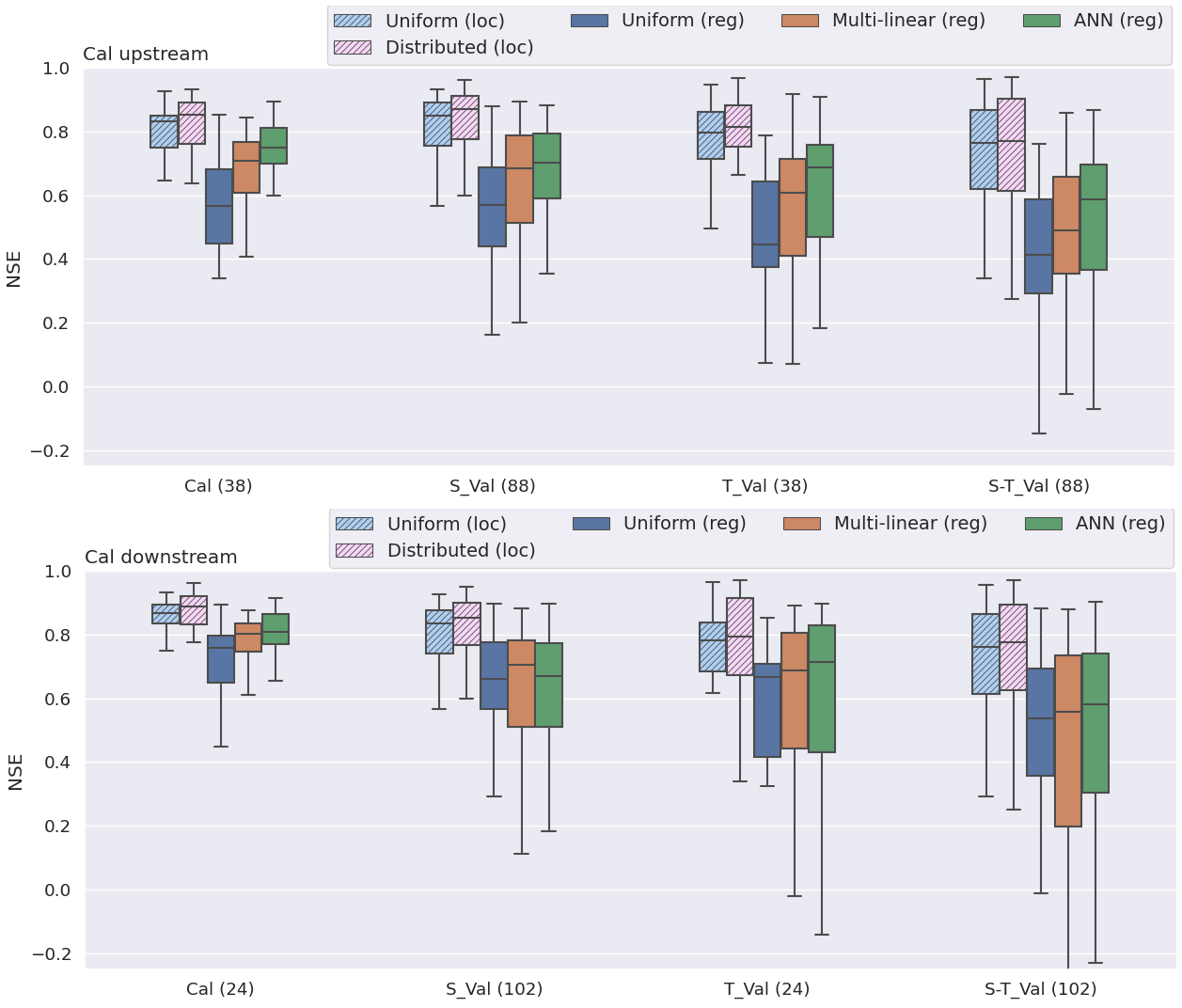}
 \caption{\rev{Boxplots of NSE scores (optimal value = 1) across calibration and validation catchments for both calibration setups which are upstream (top) and downstream (bottom), compared to reference solutions obtained by local calibration methods (Uniform (loc) and Distributed (loc)). From left to right: results are displayed for calibration catchments on period P1 (``Cal''), validation catchments on P1 for spatial validation (``S\_Val''), calibration catchments on P2 for temporal validation (``T\_Val''), and validation catchments on P2 for spatio-temporal validation (``S-T\_Val''). The numbers in parentheses indicate the count of catchments included in each boxplot.}}
 \label{fig:res-all-nse}
\end{figure}

\begin{figure}[ht!]
 \noindent\includegraphics[width=\textwidth]{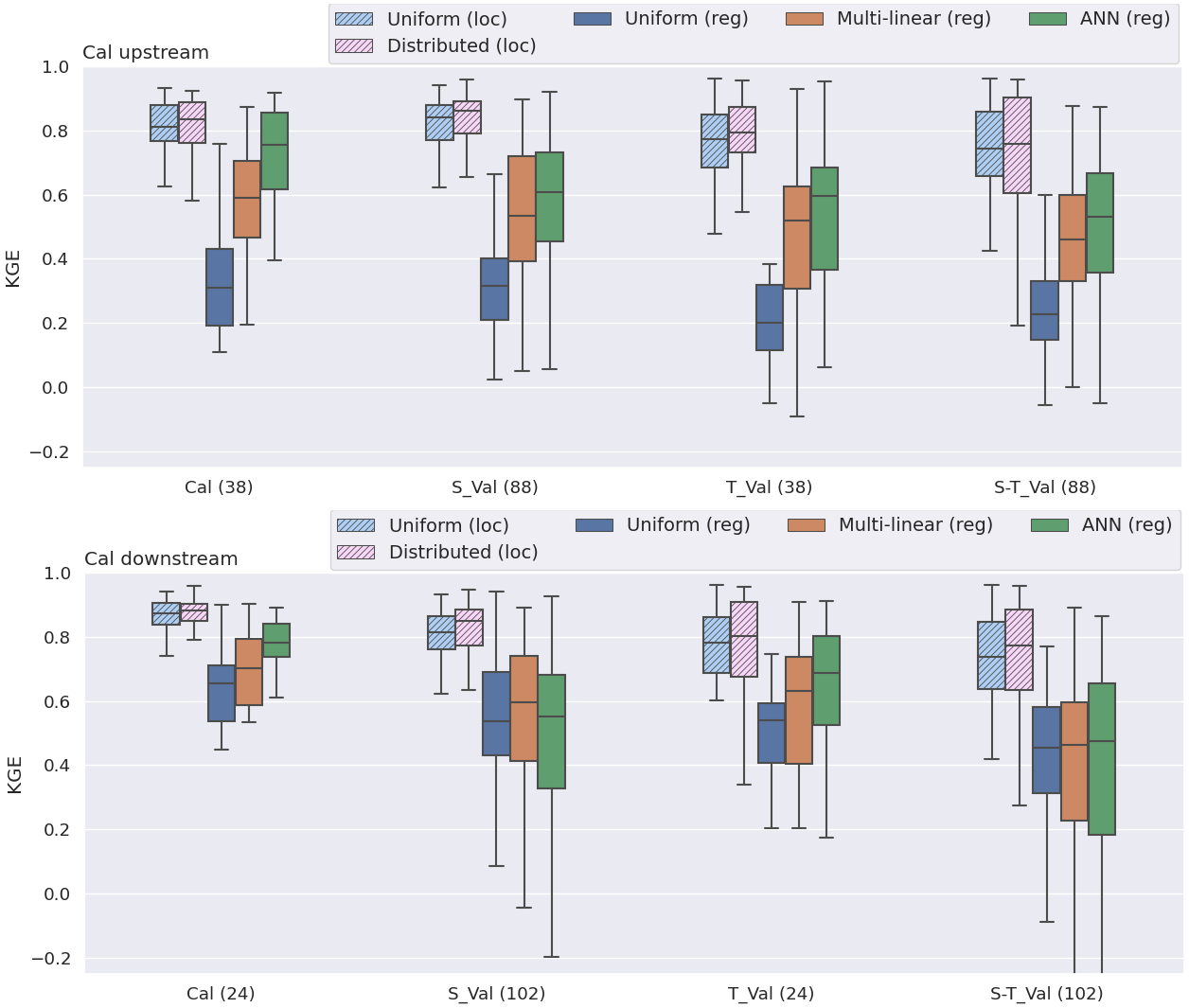}
 \caption{\rev{Boxplots of KGE scores (optimal value = 1) across calibrated (with NSE cost function) and validation catchments for both calibration setups (upstream (top) and downstream (bottom)), compared to reference solutions obtained by local calibration methods (Uniform (loc) and Distributed (loc)). From left to right: results are displayed for calibration catchments on period P1 (``Cal''), validation catchments on P1 for spatial validation (``S\_Val''), calibration catchments on P2 for temporal validation (``T\_Val''), and validation catchments on P2 for spatio-temporal validation (``S-T\_Val''). The numbers in parentheses indicate the count of catchments included in each boxplot.}}
 \label{fig:res-all-kge}
\end{figure}

\rev{Figure~\ref{fig:res-all-nse} focuses on global NSE scores (see Figure~\ref{fig:res-all-kge} for KGE scores), i.e., the calibration metric, over the whole ArcMed region, for the two local calibration methods (spatially uniform and spatially distributed calibrations) and the three regionalization methods (Uniform, Multi-linear, and ANN). Overall, it suggests that the regionalization methods incorporating information from physical descriptors and imposing model parameters spatialization (Multi-linear and ANN) lead to superior performance when compared to the spatially Uniform baseline. This trend is particularly noticeable in the more challenging calibration-validation scenario using only upstream gauges for calibration. Furthermore, the enhanced efficiency of ANN-based regionalization is evident from global performance comparisons against the Multi-Linear approach, considering both upstream and downstream basin sets used in regional optimization.} 

\noindent \rev{In calibration (``Cal''), median NSE scores for the upstream calibration on 38 catchments are 0.7 and 0.75 for Multi-linear and ANN regionalization methods, representing a marked improvement compared to 0.57 for the Uniform approach. 
By contrast, for the downstream calibration, the median NSE scores are much closer: 0.8 and 0.81 for Multi-linear and ANN regionalization methods versus 0.76 for the Uniform approach. 
This indicates that the improvement obtained with the former two regionalization methods is smaller for the downstream calibration than for the upstream one. Moreover, the NSE performance for the Uniform approach is already quite close to that of the reference local calibrations in the downstream case, leaving little room for improvement with Multi-linear and ANN regionalization methods. 
Overall, these results suggest that the multi-gauge calibration problem poses a more challenging interpolation issue on upstream gauges than on downstream gauges. Thus, the regional calibration problem incorporating descriptors is more demanding on smaller-sized catchments with lower integrative effects and potentially higher non-linear responses across basins.
In terms of temporal validation (``T\_Val'') for both upstream and downstream cases, the same ranking as in calibration is observed with the three regionalization methods. Specifically, the ANN achieves a median NSE score of 0.69 (resp. 0.71) compared to 0.6 and 0.44 (resp. 0.69 and 0.68) for Multi-linear and Uniform in the upstream calibration setup (resp. downstream calibration setup). 
In spatial validation (``S\_Val''), which assesses the potential of regionalization methods for ungauged basins, the ANN demonstrates the best performance in the upstream calibration setup, while similar performances (similar medians and interquartile ranges) of the three regionalization methods are observed in the downstream setup. 
Moving on to the most complex evaluation for the extrapolation capability of HDA-PR, spatio-temporal validation (``S-T\_Val''), results consistently show the efficiency of the ANN, improving the NSE scores by 0.11 (resp. 0.18) in upstream setup and by 0.02 (resp. 0.04) in downstream setup compared to Multi-linear (resp. Uniform). 
In summary, multi-linear regression leads to slightly better median performances in spatial validation, while ANN performs best in temporal and spatio-temporal validations in the downstream calibration setup. 
However, in the upstream calibration case, the ANN demonstrates superior performance in calibration and for all validation scenarios. This result highlights the capability of ANN to outperform (at least in terms of scoring metrics up to this point) classical regionalization approaches using lumped model parameters and multi-linear regression in the most challenging extrapolation scenario (i.e., upstream calibration).}

\begin{figure}[ht!]
 \noindent\includegraphics[width=\textwidth]{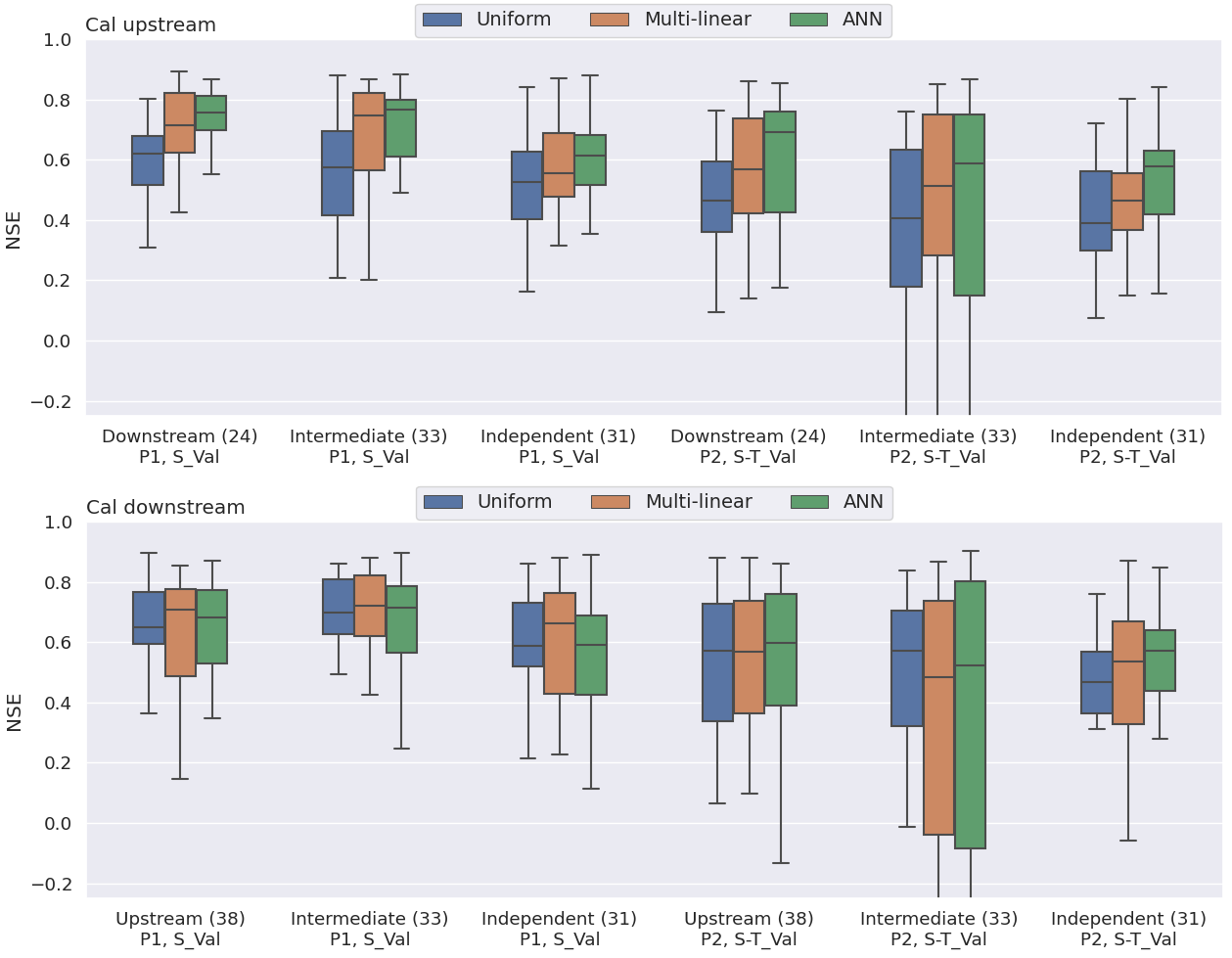}
 \caption{\rev{Comparison of NSE scores (optimal value = 1) across validation catchments categorized by their nature (upstream, downstream, intermediate, or independent) for both calibration setups (upstream (top) and downstream (bottom)). From left to right: results are displayed for validation catchments on period P1 for spatial validation (``S\_Val'') and validation catchments on P2 for spatio-temporal validation (``S-T\_Val''). The numbers in parentheses indicate the count of catchments included in each boxplot.}}
 \label{fig:res-score-by-nature-nse}
\end{figure}

\begin{figure}[ht!]
 \noindent\includegraphics[width=\textwidth]{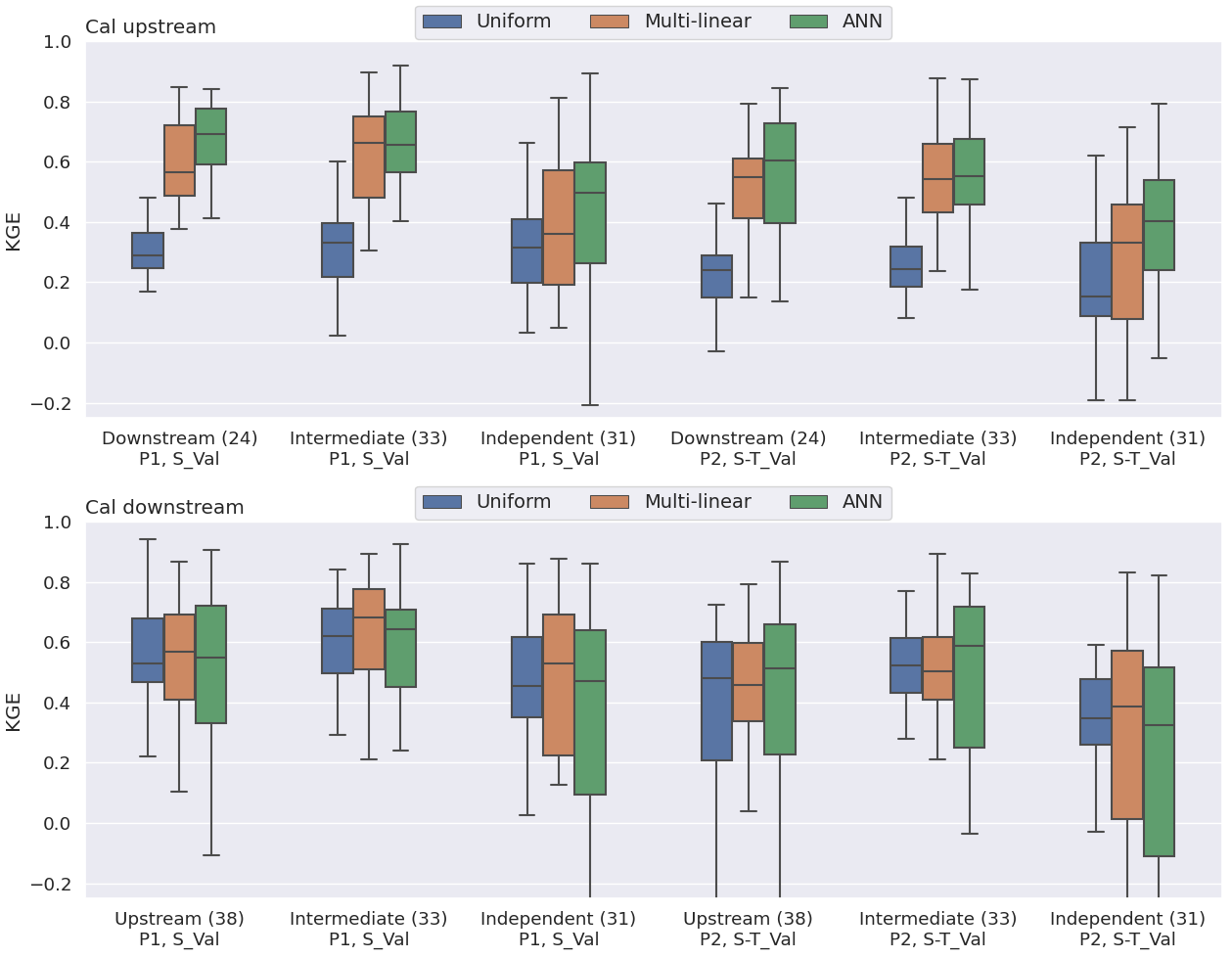}
 \caption{\rev{Comparison of KGE scores (optimal value = 1) across validation catchments categorized by their nature (upstream, downstream, intermediate, or independent) for both calibration setups (upstream (top) and downstream (bottom)). From left to right: results are displayed for validation catchments on period P1 for spatial validation (``S\_Val'') and validation catchments on P2 for spatio-temporal validation (``S-T\_Val''). The numbers in parentheses indicate the count of catchments included in each boxplot.}}
 \label{fig:res-score-by-nature-kge}
\end{figure}

\rev{We now delve into a more detailed analysis of NSE performances in validation catchments categorized by their nature (upstream, downstream, intermediate, or independent) for both calibration setups: upstream and downstream, as illustrated in Figure~\ref{fig:res-score-by-nature-nse} (see Figure~\ref{fig:res-score-by-nature-kge} for KGE scores). In the downstream calibration setup, no clear ranking appears. %for spatial validation (resp. spatio-temporal validation), the median NSE scores are slightly lower for ANN compared to Multi-linear, but with a narrower spread for ANN (resp. median NSE slightly higher with ANN, still with less spread except for intermediate catchments). 
Regarding the upstream calibration setup, the median NSE is systematically higher with ANN regionalization, with less spread (interquartile range) in all basin classes except for the 33 intermediate catchments, which show a slightly smaller first quartile compared to Multi-linear. Moreover, significantly higher performances are achieved with ANN in spatio-temporal validation. This is particularly notable on a relatively limited number of smaller-sized catchments in the calibration, indicating an effective extraction of information from both discharge data and descriptors. This effectiveness becomes evident when the model is tested on the validation catchments, which are more numerous, may or may not be nested and have varied areas. Overall, the performances obtained with both upstream and downstream descriptor-based calibrations show the relevance of regionalization with ANN. Indeed, ANN performs best in the upstream setup. ANN performance is not always the highest in the downstream setup, but it remains very similar to other regionalization approaches. The complex and highly-parameterized nature of the ANN hence does not seem to result in a deterioration of performances in validation, even when this complexity may not be necessary.}


\begin{figure}[ht!]
 \noindent\includegraphics[width=\textwidth]{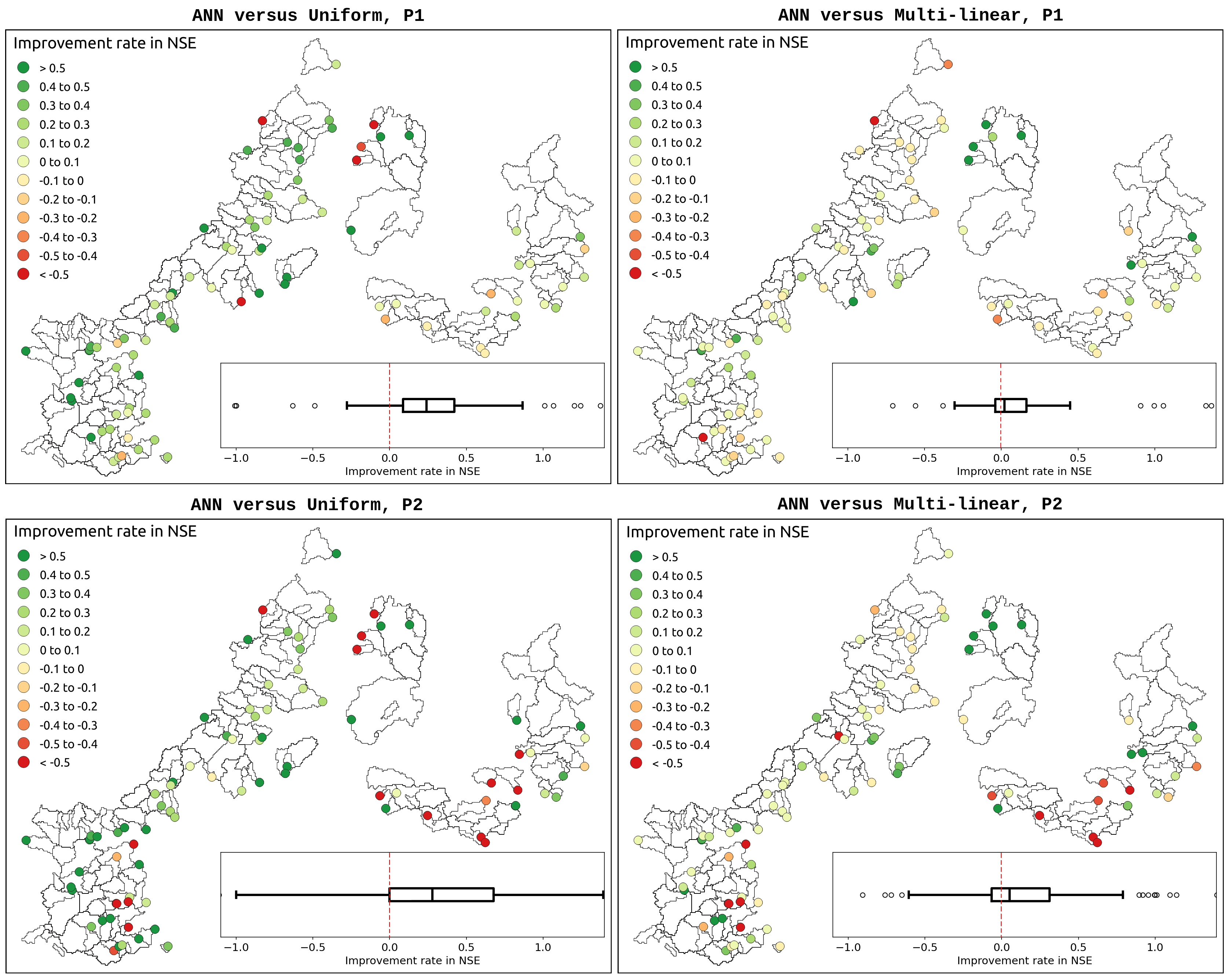}
 \caption{\rev{Comparison of NSE improvement rates in spatial and spatio-temporal validation for HDA-PR with ANN mapping versus uniform (left) and multi-linear (right) mappings, utilizing models calibrated with upstream catchments. Results are shown across validation catchments positioned downstream, intermediate, and independent for spatial validation on period P1 (top) and spatio-temporal validation on period P2 (bottom). The improvement rate at a catchment $i$ is calculated as $r^i=\frac{NSE^i_{\text{ANN}}-NSE^i_{\square}}{|NSE^i_{\square}|}$, where $\square$ represents either multi-linear or uniform mapping.}}
 \label{fig:map-scores}
\end{figure}

\rev{
To quantify how much the ANN-based regionalization approach improves the model performance compared to multi-linear regression and spatially uniform parameters regionalization, we compute the improvement rate in terms of NSE of the ANN relative to the other methods. We focus on the upstream calibration setup and evaluate improvement rates of ANN versus the two simpler approaches (Uniform and Multi-linear) in spatial and spatio-temporal validation, as illustrated in the maps of Figure~\ref{fig:map-scores}. Comparing to Uniform, the ANN yields positive improvement rates at more than 80\% of validation catchments on P1, with a median improvement rate of around 0.25 (see the boxplot in the top left panel of Figure~\ref{fig:map-scores}). Transitioning to P2, the variation of improvement rates among catchments becomes more significant. Catchments that exhibited positive improvement rates during P1 generally maintain positive or enhanced rates, whereas some of those with negative rates either worsen or remain unchanged. Although the median improvement rate of NSE reaches 0.3, the interquartile range and whiskers visibly expand, as indicated by the boxplot in the bottom left panel of Figure~\ref{fig:map-scores}. Regarding the improvement rates of ANN versus Multi-linear, we observe similar trends in the variation of improvement rates among catchments across the two periods. While the number of catchments with high improvement rates is generally reduced, more than 65\% of catchments still exhibit positive rates on both periods. Nevertheless, it is apparent that some catchments (situated in the southeast region of the map and known for their complex hydrological behaviours including karstic effects and more difficulties in hydrological modeling) display negative improvement rates of the ANN versus both Uniform and Multi-linear approaches.
This discrepancy may stem from difficulties in extrapolating effective model parameters from the considered physical descriptors at these locations with the ANN. This could be improved by using other descriptors and/or stronger constraints, for example, on the sensitive exchange parameter $k_{exc}$ as discussed later.}

\begin{figure}[ht!]
 \noindent\includegraphics[width=\textwidth]{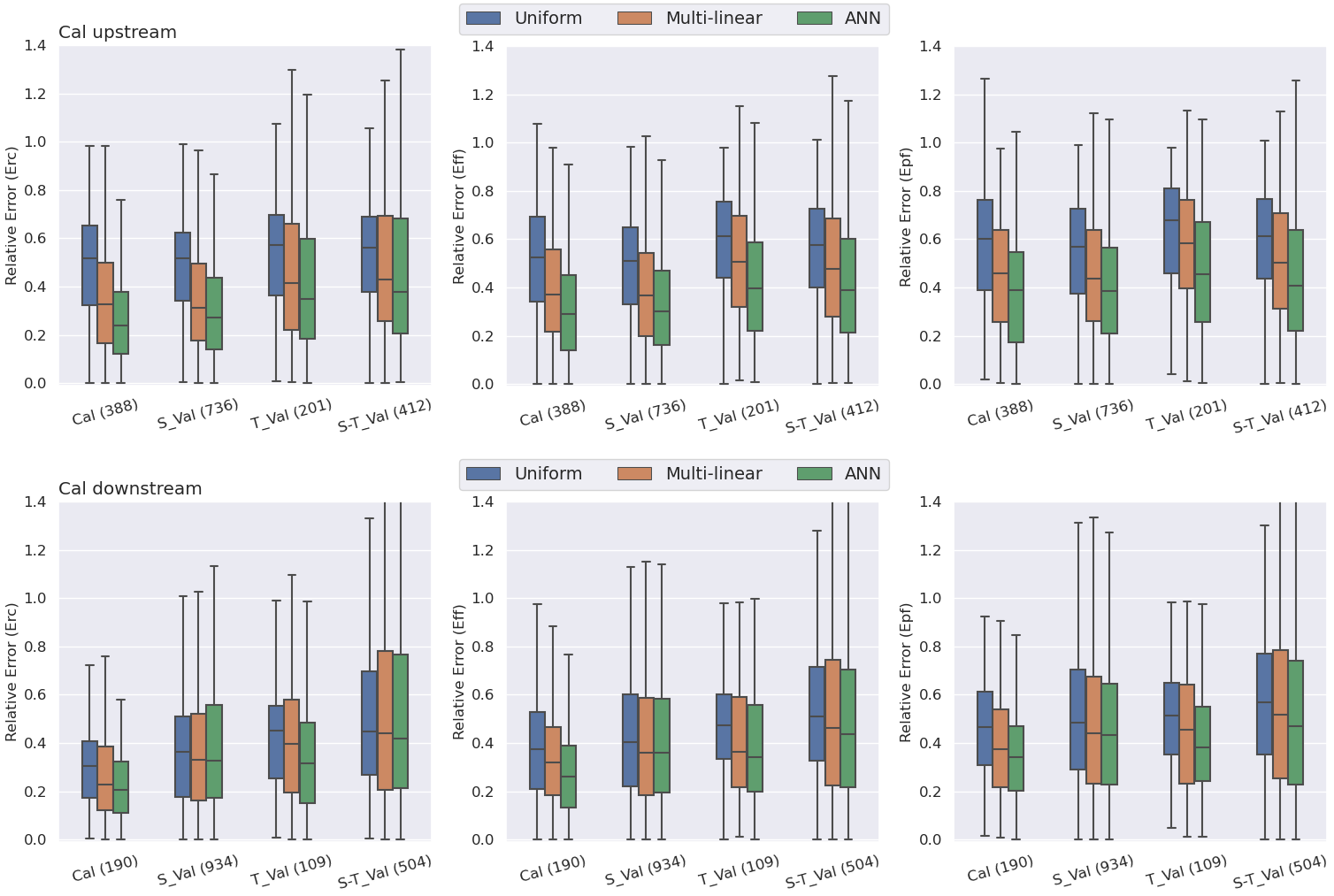}
 \caption{\rev{Relative error (optimal value = 0) of three flood event signatures—flood runoff coefficient (Erc), flood flow (Eff), and peak flow (Epf)—for both calibration setups: upstream (top) and downstream (bottom). The evaluation is based on a total of 1,737 flood events, with 1,124 selected during the calibration period P1 and 613 selected during the validation period P2, across 126 catchments. From left to right: results are displayed for calibration catchments on P1 (``Cal''), validation catchments on P1 for spatial validation (``S\_Val''), calibration catchments on P2 for temporal validation (``T\_Val''), and validation catchments on P2 for spatio-temporal validation (``S-T\_Val''). The numbers in parentheses indicate the count of flood events included in each boxplot.}}
 \label{fig:sign-val}
\end{figure}

In order to obtain a more robust evaluation criterion adapted to flood modeling, we consider validation in terms of multiple evaluation metrics based on hydrological signatures for flood events, which are computed via an automated segmentation algorithm proposed by \citeA{HUYNH2023signatures}. \rev{This evaluation is based on the relative error of three flood event signatures: flood runoff coefficient (Erc), flood flow (Eff), and peak flow (Epf), computed on a total of 1,737 flood events selected from the entire study area over a six-year period from August 2016 to July 2022. In the upstream calibration setup, the two non-uniform regionalization methods, and in particular the ANN, demonstrate their ability to outperform the uniform regionalization (using lumped model parameters) in both calibration and validation, as illustrated in Figure~\ref{fig:sign-val}. For example, the relative errors (median over flood events) of simulated peak flows using ANN are around 0.4, compared to over 0.6 (for the uniform baseline) and 0.5 (for Multi-linear) in spatio-temporal validation (upstream calibration setup). In the downstream calibration setup, although the ANN does not markedly outperform the Multi-linear and Uniform regionalizations, it still achieves the best scores in terms of flood event signatures in both calibration and validation. For example, in spatio-temporal validation of the peak flow, ANN achieves a median relative error of 0.47, while Uniform and Multi-linear have median relative errors of 0.58 and 0.52. It is noteworthy that these flood signatures-based metrics were not included in the cost function during the calibration process, which further supports the robustness and power of the regionalization methods, particularly the one based on ANN. It thus underscores the potential of HDA-PR for enhancing flash flood forecasting systems.}

\rev{Finally, we discuss the computational efficiency of HDA-PR in terms of memory allocation and computation time. The SMASH numerical code, which incorporates the proposed HDA-PR algorithms, enables parallel computations using the openMP library \cite{dagum98} and simple decomposition of the spatial domain, thus reducing CPU computational time. Additionally, we have implemented sparse rainfall storage to mitigate memory consumption, which is crucial for running the adjoint model (whose memory usage has also been improved using a checkpointing technique) at each evaluation of the gradient, especially across expansive spatio-temporal domains. Note that the gridded atmospheric data constitutes a significant portion of memory usage across all methods. %, alongside the adjoint model for the two gradient-based optimization methods employed, namely L-BFGS-B for Multi-linear and Adam for ANN. 
The CPU and memory computational costs of the different calibration runs over a relatively large computational domain are given in Table~\ref{tab:number_param_mapping}, demonstrating that the application of the method on larger domains such as at a country scale is feasible.}

\begin{table}[ht!]
\caption{
\rev{Computation time for optimizing different regionalization mappings, where $N_{\theta}=4$ and $N_{D}=7$. Each calibration approach is performed on an AMD Opteron(TM) Processor 6276 at 2.5 GHz, running in parallel across 6 CPUs. The values are presented for the downstream calibration setup, with similar performance for the upstream case.}}
\label{tab:number_param_mapping}
%\scalebox{0.9}{
\begin{tabular}{c >{\centering\arraybackslash}p{0.15\textwidth} >{\centering\arraybackslash}p{0.15\textwidth} >{\centering\arraybackslash}p{0.15\textwidth} c c}
    \hline
    Mapping & Number of parameters & Memory usage (GB) & Optimization algorithm & Iterations & Time (h) \\
    \hline
    Uniform & 4 & 10.97 & SBS & 11 & 16.94 \\
    Multi-linear & 32 & 10.98 & L-BFGS-B & 250 & 157.24 \\
    ANN & 6276 & 11.29 & Adam & 350 & 184.45 \\
\end{tabular}
%}
\end{table}

\subsection{\rev{Interpretation of the Learning Process and Inferred Parameters}}

\begin{figure}[ht!]
 \noindent\includegraphics[width=\textwidth]{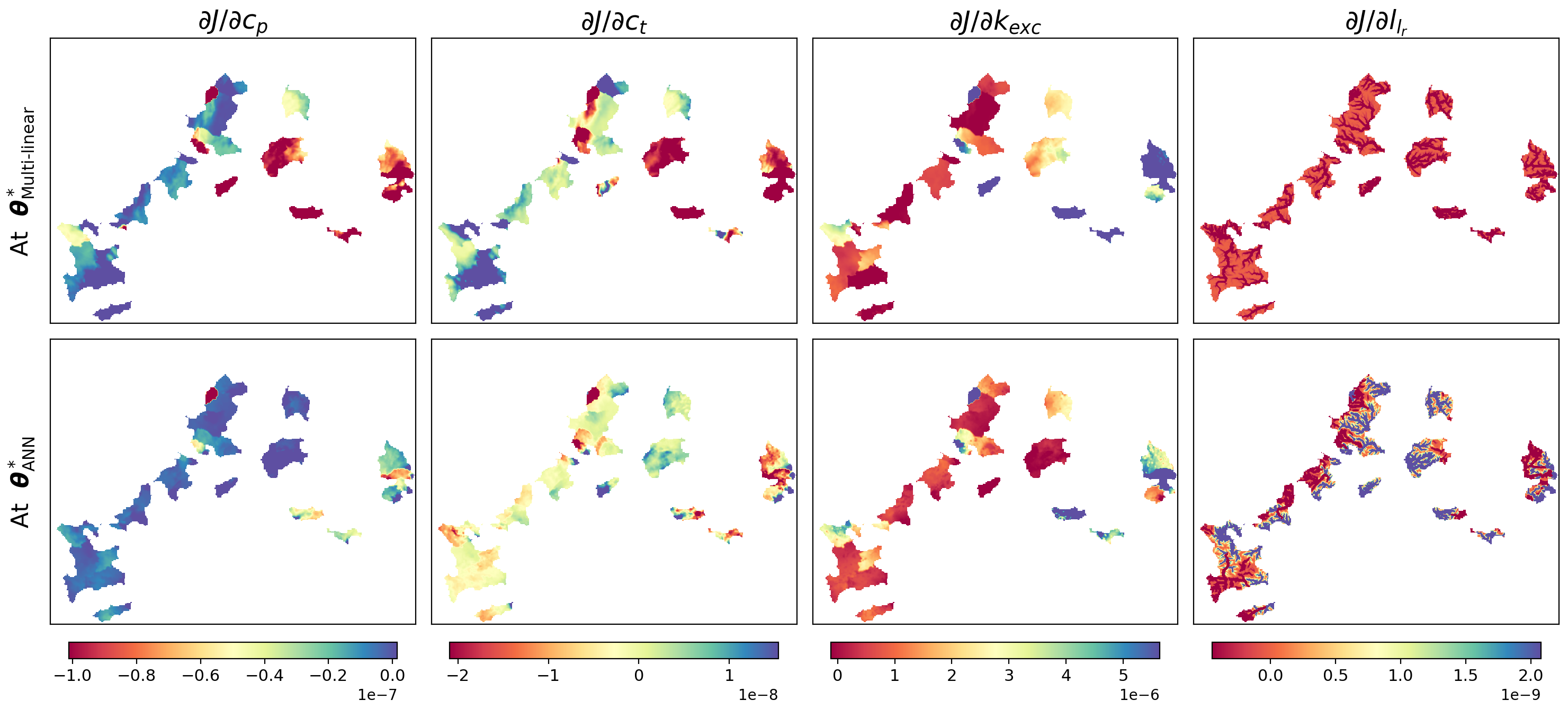}
 \caption{\rev{Spatially distributed cost gradients $ \nabla_{\boldsymbol{\theta}}J \left(\boldsymbol{\theta}_\square^*\right)$ to conceptual hydrological parameters used in the optimization process (of $\boldsymbol{\rho}$ using $\nabla_{\boldsymbol{\rho}} J = \nabla_{\boldsymbol{\theta}} J . \nabla_{\boldsymbol{\rho}} \boldsymbol{\theta}$) showed for downstream calibration (for the sake of visibility) at initial points $\boldsymbol{\theta}^{*}_{\text{Multi-linear}}\coloneqq\mathcal{P}(., \boldsymbol{\rho}^{*}_\text{Multi-linear})$ and $\boldsymbol{\theta}^{*}_{\text{ANN}}\coloneqq\mathcal{N}(., \boldsymbol{\rho}^{*}_{\text{ANN}})$, where $\boldsymbol{\rho}^{*}_\text{Multi-linear}$ (resp. $\boldsymbol{\rho}^{*}_\text{ANN}$) is the background value of the multi-linear regression (resp. ANN) mapping. Note that the cost gradient values are obtained for the active cells within the gauged partition, i.e., cells contributing to the simulated discharge at gauges (downstream here) where the observation cost function $J$ is evaluated.}}
 \label{fig:grad_j_theta}
\end{figure}

\rev{A key feature of the proposed HDA-PR algorithm pertains to the differentiability of the forward spatially distributed model, including the ANN or Multi-Linear regionalization mappings (cf. section~\ref{sub:forward_model_preregio}), which allows obtaining the gradient $\nabla_{\boldsymbol{\theta}} J$ of the cost function with respect to the hydrological model parameters. This gradient facilitates the computation of $\nabla_{\boldsymbol{\rho}} J$ necessary for optimizing the regional control vector $\boldsymbol{\rho}$ using algorithms tailored for high-dimensional inverse problems (cf. section~\ref{subsec:inverse-algo}), while maintaining a ``physical'' interpretation. Spatial maps of the cost gradient with respect to hydrological model parameters (which also represents the local sensitivity of the model parameters), at a background value $\boldsymbol{\theta}^*=\mathcal{F}_{R}(., \boldsymbol{\rho}^*)$ corresponding to the control value $\boldsymbol{\rho}^*$ at the first iteration of the optimization process (cf. section~\ref{deterministic_regio} and section~\ref{ann_regio}), are shown in Figure~\ref{fig:grad_j_theta}: $\nabla_{\boldsymbol{\theta}}J(\boldsymbol{\theta}^{*})(x)=\left(\frac{\partial J(\boldsymbol{\theta}^{*})}{\partial c_{p}},\frac{\partial J(\boldsymbol{\theta}^{*})}{\partial c_{t}},\frac{\partial J(\boldsymbol{\theta}^{*})}{\partial k_{exc}},\frac{\partial J(\boldsymbol{\theta}^{*})}{\partial l_{l_r}}\right)(x)$). The figure displays the maps for the downstream calibration setup, chosen for enhanced visualization compared to smaller basin sizes utilized in the upstream calibration setup. For both regionalization mappings (ANN or Multi-Linear), the initial cost gradient (sensitivity) at different points $\boldsymbol{\theta}^*$ in model parameter space for each mapping, indicates a high sensitivity to the non-conservative exchange parameter $k_{exc}$, which is coherent with global and local sensitivity analysis from \citeA{HUYNH2023signatures} on a comparable model structure without regionalization mapping. Interestingly, the gradient maps (for the first optimization iteration) of the forward model containing descriptors-to-parameters mappings show the spatial trends that will be obtained in the following optimization iteration, since a positive gradient should result in a parameter increase and conversely (if $\nabla_{\boldsymbol{\rho}}\boldsymbol{\theta}$ is of the same sign as $\nabla_{\boldsymbol{\theta}}J$ plotted here since $\nabla_{\boldsymbol{\rho}} J = \nabla_{\boldsymbol{\theta}} J . \nabla_{\boldsymbol{\rho}} \boldsymbol{\theta}$). Note that due to the high dimension of $\boldsymbol{\rho}$ and its non-physical meaning for ANN weights and biases, $\nabla_{\boldsymbol{\rho}}\boldsymbol{\theta}$ cannot be plotted in a simple manner and, more importantly, cannot be ``physically'' interpreted like $\nabla_{\boldsymbol{\theta}}J$. Moreover, these maps of $\nabla_{\boldsymbol{\theta}}J$ interestingly contain the footprint of physical descriptor patterns (along with the footprint of model structure and multi-site discharge observations and cost function formulation); they indicate sensitivity ``hot spots'', which could be useful in future works for tailoring descriptor processing and choice, as well as for studying other information sources, tailoring cost functions, parameter bounds, and regularizations of the forward-inverse problem.}

\begin{figure}[ht!]
 \noindent\includegraphics[width=\textwidth]{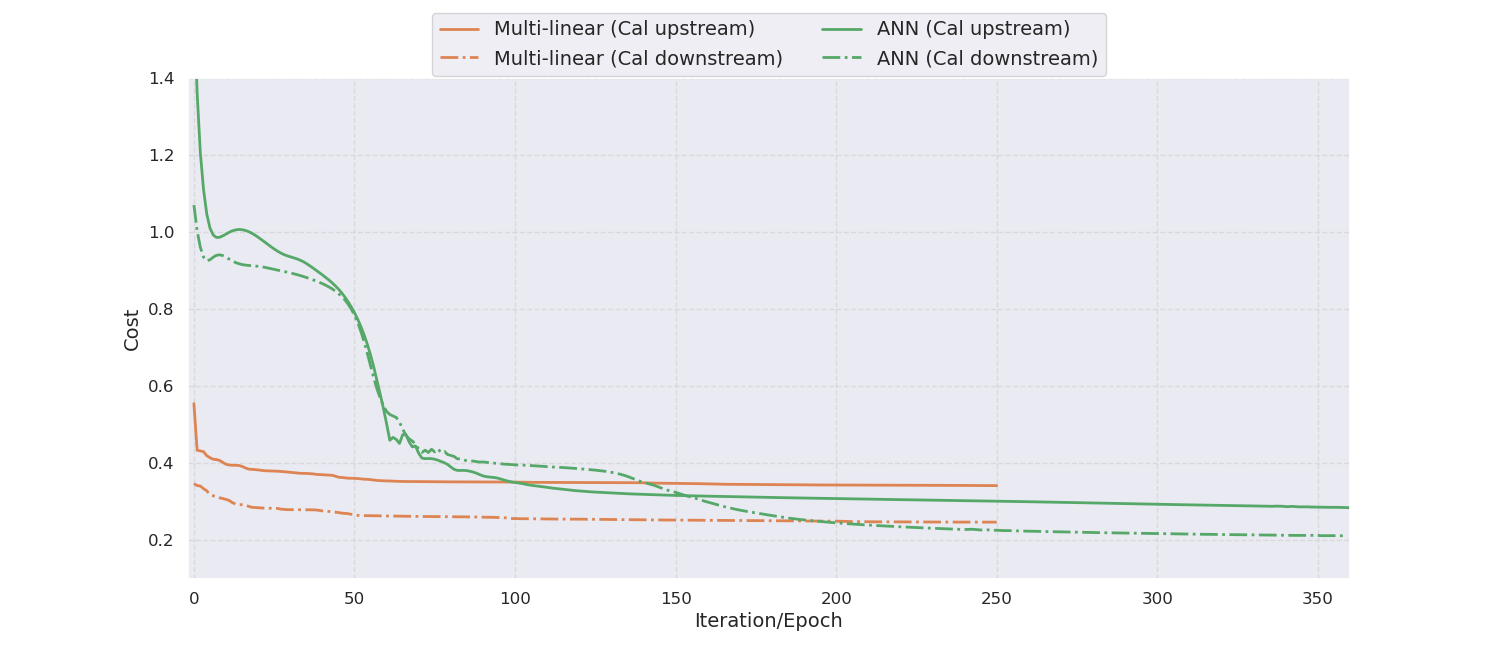}
 \caption{\rev{The descent of the cost function $J=1-NSE$ for both Multi-linear and ANN approaches across both calibration setups.}}
 \label{fig:cost-descent}
\end{figure}

When calibrating a model with gradient-based optimization algorithms, it is also important to discuss the descent of the cost function. This analysis enables understanding how optimization algorithms converge towards the global or local minimum of the cost function, and identifying potential trade-offs between model flexibility and overparameterization, thereby completing validation results. \rev{The descent of the cost function $J$ is represented in Figure~\ref{fig:cost-descent}.}

\noindent It is apparent that the cost functions in \rev{the Multi-linear case start from a more optimal point than the ANN (approximately 0.37 in downstream calibration setup and 0.58 in upstream calibration setup)} since they use a uniform background solution obtained by a global optimization method, as mentioned in section~\ref{deterministic_regio}. Furthermore, they converge after around 200 iterations and remain monotonous throughout the optimization process. \rev{The ANN, with a significantly larger number of parameters (Table~\ref{tab:number_param_mapping}), can achieve a lower cost despite starting from a higher cost.} Moreover, the ANN cost function in Figure~\ref{fig:cost-descent} is not monotonous throughout training epochs: it shows localized increases for several instances within the first 100 epochs. 
\rev{This result can be attributed to the complexity of the regionalization mapping in each case, suggesting that the surface of the cost function is more complex for ANN than for multi-linear mapping. 
%The variability of the projected cost gradient shown in the right panels of Figure~\ref{fig:cost-descent} suggests that the surface of the cost function is more complex for ANN than for the \rev{multi-linear regression}. This is confirmed by the gradient values in the right panels of Figure~\ref{fig:cost-descent}: multi-linear regression leads to projected gradients \rev{that are much higher than those of ANN.} This discrepancy can be attributed to the complexity of the regionalization mapping in each case, with the ANN being the more complex. 
Moreover, \rev{the Adam algorithm used to optimize the ANN enables the exploration of} different paths to locate the optimum, therefore avoiding getting stuck in local minima, where significant changes occur in the control vector space (biases and weights).} This property could be essential in tackling equifinality and reach robust global optimum even with different starting points. 

\begin{figure}[ht!]
 \noindent\includegraphics[width=\textwidth]{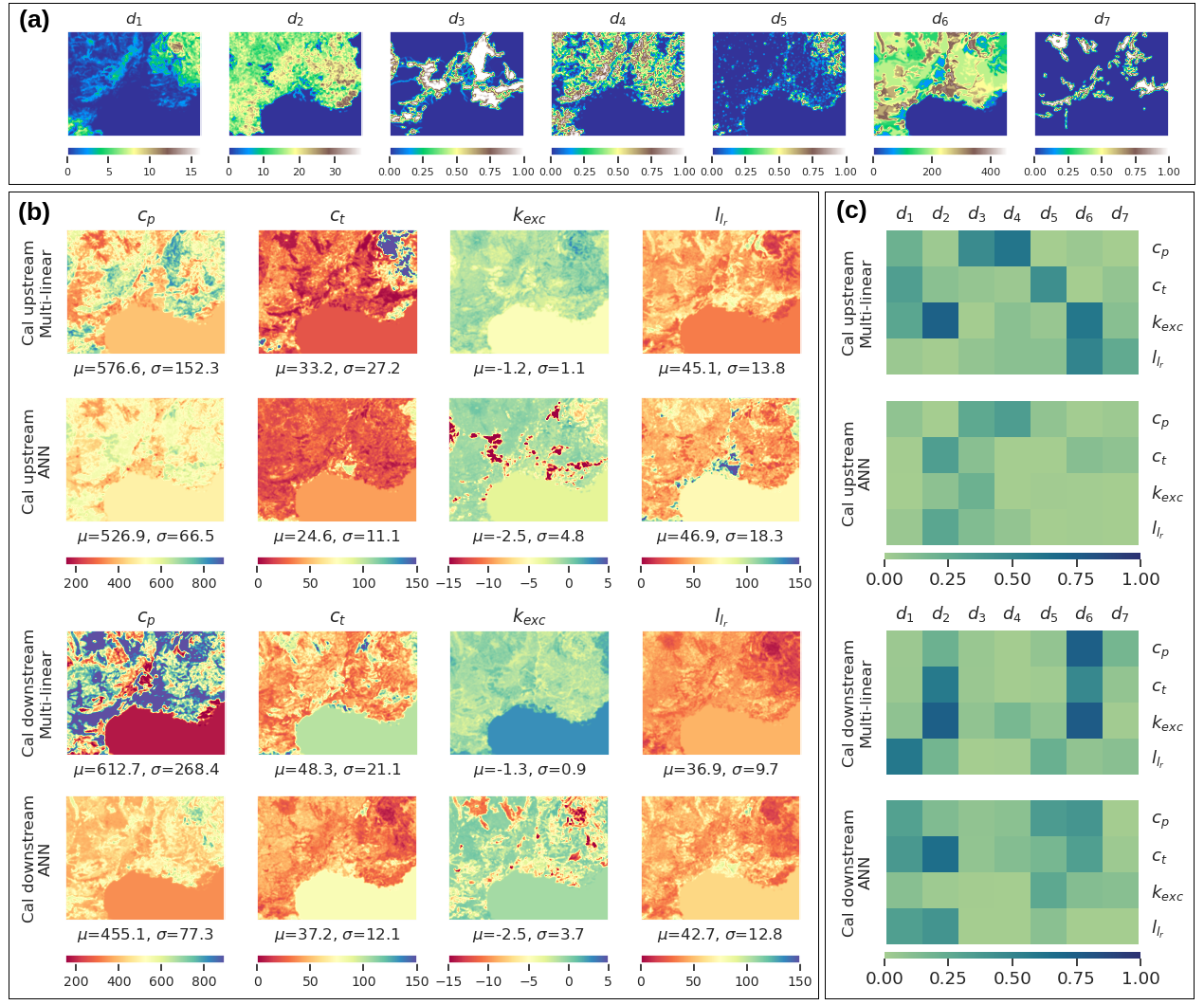}
 \caption{\rev{Analysis of input descriptors and output model parameters for two regionalization methods (Multi-linear and ANN) across both calibration setups (upstream and downstream): (a) Spatial distribution of physical descriptors ($d_1$-$d_7$), details provided in Table~\ref{tab:descriptors}; (b) Spatial distribution of calibrated hydrological parameters ($c_p$, $c_t$, $k_{exc}$, $l_{l_r}$) with $\mu$ and $\sigma$ indicating their spatial average and standard deviation over the active cells within the spatial domain $\Omega$; (c) Linear correlation between descriptors and parameters over the active cells within the spatial domain $\Omega$.}}
 \label{fig:param-map}
\end{figure}

\rev{Finally, Figure~\ref{fig:param-map}-b shows the optimized parameter maps obtained with the two non-uniform regionalization methods across both calibration setups, that can be physically interpreted. In the downstream calibration setup, the spatial patterns of the drainage density descriptor ($d_2$) and the potential water reserve descriptor ($d_6$) in Figure~\ref{fig:param-map}-a are easily observable in the distributed parameter maps in the multi-linear regression case, as evidenced by strong correlations between these descriptors and the capacity of the production store ($c_p$), the capacity of the transfer store ($c_t$), and the non-conservative water exchange ($k_{exc}$). Such correlations can be quantified using one-to-one parameter-descriptor correlation matrices as depicted in Figure~\ref{fig:param-map}-c. 
%The relations between the model parameters and physical descriptors are not limited to a multivariate linear exploitation of descriptors but interestingly enable non-linear multivariate extraction of information in the ANN case, as shown by correlation matrices.
Correlations are smaller in the ANN case than in the Multi-linear case: this suggests that the parameter maps derived with the former go beyond a simple linear combination of descriptors, and hence include some degree of non-linearity. 
%This leads to parameter maps that are quite distinct from the multi-linear regression case. Furthermore, the correlation patterns shown in Figure~\ref{fig:param-map}-c are strongly different between the ANN and Multi-linear approaches. This indicates that the ANN identifies correlations between input descriptors and model parameters that are distinct from the ones identified by multi-linear regression.
}

\rev{Overall, the obtained regional hydrological parameter maps, together with the performances in interpolation and spatio-temporal extrapolation cases (as shown in section~\ref{subsec:perf}), demonstrate the relevance of the intrinsic spatial constraint introduced by the descriptors-to-parameters mappings. These mappings result in a regularizing effect for the overparameterized calibration problem in distributed hydrological modeling from sparse discharge data.}

%The following section focuses on results more specific to the optimization process and provide insights on the use of multivariate regression and ANN-based methods \rev{in the study area}. 

\section{Discussions}
\label{sec:discuss}

Learning the spatial variability of conceptual hydrological parameters may be difficult to achieve with simple regionalization methods (e.g., based on multi-linear regression). However, while complex regional mappings can reduce the misfit between observed and simulated hydrological responses, their ability to produce physically interpretable results may be questioned. %This section presents compelling findings and insightful observations obtained from the calibration using HDA-PR, with a particular focus on the learning process in the case of ANN.
The various regionalization mappings define the way information is transferred from physical descriptors to hydrological parameters. It is important to ensure that HDA-PR effectively uses the physical patterns provided by descriptor maps to constrain the estimation of meaningful parameter maps. To address this question, it is worth noting that the ``safest'' approach is to use multi-linear regression, which corresponds to a simple weighted average of the descriptor patterns. The case of multi-polynomial regression \rev{was not included in the case study of this paper, but was tested by \citeA{huynh2023learning} in a sub-region of the French Mediterranean area}. In this case, the risk of losing physical properties may arise when the polynomial degree is unbounded. To mitigate this risk, \rev{we propose imposing} bounds on the polynomial degree, $0.5\leq\beta_{k,d}\leq2$. ANNs, however, pose the most complicated scenario, where the control vector (that is, the weights and biases) consists of numerous parameters that are difficult to physically constrain. Our hands-on experience indicates that a multilayer perceptron with two or three hidden layers is sufficient for learning the parameters of a parsimonious conceptual distributed hydrological model without under- or over-extracting the physical information of the input descriptors. Note that the number of neurons in each layer must be reasonable, and should not exceed $\sqrt{N_D . N_x}$ based on our experiments. Next, to alleviate the vanishing gradient problem inherent to the ANN, we employed several techniques commonly used in the machine learning community. First, we applied Xavier initialization \cite{glorot2010understanding} to the weights, maintaining a reasonable magnitude of the gradients. Second, we utilized the ReLU activation function or its variants in the hidden layers, enabling the gradient to flow more freely through the network. Third, we varied the number of hidden layers between 2 and 4, striking a balance between network flexibility and exacerbation of the vanishing gradient problem. Ultimately, we employed a relatively high initial learning rate (e.g., \rev{from 0.003 to 0.005}) to prevent the gradients from shrinking excessively during training.

%\textbf{As mentioned earlier, Figure~\ref{fig:param-map}-b suggests that the three regionalization methods are able to produce meaningful parameter maps, notably when looking at their correlations with the physical descriptors (Figure~\ref{fig:param-map}-c). Each method results in different parameter maps, in contrast to the similar ones observed in Ardèche, where the problem is less challenging. For Ardèche, we observe that the ANN can identify relations between descriptors and parameters which are similar to those found in both regression cases (see Figure~\ref{fig:param-map-ardeche}-c). This finding suggests that the ANN can effectively approximate the multi-linear regression when dealing with areas like Ardèche, where it does not necessitate more intricate mappings. This demonstrates the robustness and flexibility of the ANN, which proves to be efficient not only in complex study areas but also in simpler ones. // As previously, I would delete the whole paragraph: the first sentence just repeats a basic result, and I think the rest of the paragraph would better fit at the end of section 3.2, as suggested earlier. //}

\rev{While the ANN approach showed superior performance in the upstream calibration setup, its performance did not exhibit significant improvement in the downstream scenario, where the Uniform approach already yielded quite satisfactory results compared to local calibration references. It is worth noting that while ``extreme'' setups such as upstream-only and downstream-only calibrations are useful in the academic context of this paper, mixed setups that combine both, as well as include independent catchments, are likely to be employed in practice. For example, a calibration setup with a random selection of catchments was tested in \citeA{huynh2023learning, huynh2023multigauge} for a subset of the 126 catchments presented in this paper. Furthermore, the performance of calibration setups that use nested gauges could be improved by adapting the cost function for better selecting and weighting information between calibration gauges and flow signal parts (e.g., with differentiable flood signatures-based algorithm for calibration \cite{HUYNH2023signatures}), as well as by accounting for data and structural uncertainties in this cost function \cite{kuczera2010, renard2010}.}

Regarding regionalization over larger areas, such as for large basins or at country scales, an increased flexibility in the regional mapping might be needed \rev{to deal with significant physical heterogeneity. For instance, one may wish to use different transfer functions for different clusterings of the spatial domain (e.g., into sub-regions or hydrological response units (HRU)). This can be achieved through the use of spatialized regional controls, for example as done in regional calibration for catchment clusters determined with a similarity measure \cite{HUANG2019}. In the proposed HDA-PR framework, mappings that vary with regions or HRU can be obtained by including indicators of such regions/HRUs in the list of descriptors. 
%Such spatialized regional controls can be through masked descriptor maps, for each hydrological parameter independently or jointly. 
This additional flexibility would certainly be necessary to circumvent the rigidity of the multi-linear mapping, but maybe not for a flexible mapping such as the ANN one. 
Another interesting research avenue is to develop an automatic identification of effective physical descriptors from large databases as well as identification of optimal spatial flexibility for constructing effective regional data assimilation approaches.}

%In this work, the capabilities of HDA-PR have been successfully demonstrated in a high-dimensional and challenging high-resolution flash-flood modeling context. Determining effective physical descriptor sets from large databases as well as finding optimal spatial flexibility represent interesting research avenues for constructing \rev{effective} regional data assimilation approaches. \textbf{2BeDevelopped}

\rev{Finally, an important research avenue is the development of a probabilistic framework to quantify the uncertainties surrounding the application of HDA-PR. Several challenges need to be addressed for this purpose, some of them being related to the spatialized nature of the hydrologic model, while others are more specific to the large dimensionality induced by the regionalization mappings, in particular the ANN one. First, accounting for the uncertainty affecting the forcing and response data used in calibration requires a modification of the cost function based on data error models. Such models need to be parsimonious (e.g., to represent spatialized rainfall errors using a controlled number of parameters, \citeA{mustafa_estimation_2018}), yet recognize the specificities of some data acquisition procedures (for instance, the partly systematic nature of rating curve errors affecting streamflow, \citeA{horner_impact_2018}). Second, since the structural errors made by the hydrologic model are varying in space, a probabilistic model describing structural uncertainty would need to be regionalized, as is the hydrologic model itself. Third, estimation tools need to be able to cope with the high dimensionality of the estimation problem. For instance, the use of Markov Chain Monte Carlo methods will certainly require methods that can take advantage of the differentiable implementation of HDA-PR \cite{hoffman_no-u-turn_2014} and of parallell computing capabilities \cite{laloy_high-dimensional_2012,syed_non-reversible_2022}. Alternatively, the search of regionalization mappings that retain the flexibility provided by ANNs while being more parsimonious is another promising perspective.}

%\textbf{Last but not least, we want to highlight that the proposed ANN scheme in the HDA-PR algorithms has been constructed incrementally and that we were largely inspired by our mathematical/numerical understanding and know-hows built with the regression methods studied in a known VDA context. // I don't understand the point you want to make here, needs rewording I believe //}

\section{Conclusion}
\label{sec:Conclusion}

A Hybrid Data Assimilation and Parameter Regionalization (HDA-PR) approach has been introduced in this study. We investigated the potential of incorporating learnable regionalization mappings, including multivariate polynomial regressions and neural networks, into a differentiable high-resolution hydrological model. To the best of our knowledge, we present the first implementation of ANNs within this context, enabling a seamless regionalization in hydrology. Effective optimization algorithms capable of performing high-dimensional optimizations from multi-source data have been obtained with:
\begin{itemize}
    \item effective regional transfer functions of adaptable complexity, enabling the use of information from heterogeneous data sources, \rev{with learning of a non-linear multivariate mapping in the case of the ANN, }and %with flexible formulations 
    \rev{providing effective spatial constraint of} varied \rev{flexibility};
    \item a differentiable forward hydrological model, embedding the regional mappings, that enables accurate computation of spatially distributed gradients of the multi-gauge cost function - which is crucially needed in the context of sparse observations (i.e., cost evaluation locations), and relatively small gradient values;
    \item optimization algorithms, adapted to high dimensional problems, with seamless flow of cost gradients, especially when combined with physical descriptors and spatial gradients, which efficiently enhance the transferability of geophysical properties from gauged to ungauged locations.
\end{itemize}

HDA-PR has been thoroughly evaluated \rev{with two calibration setups in the French Mediterranean region with a high-resolution spatio-temporal hydrological modeling approach using multi-gauge discharge and descriptors maps. The results obtained on both calibration setups, and especially on the most challenging calibration scenario using only upstream gauges, highlight the effectiveness of HDA-PR that utilizes physical descriptors, surpassing the performance of a uniform regionalization method with lumped model parameters. Moreover, the ANN exhibited superior performance compared with multi-linear regression, that is, an approach with the same complexity as transfer functions used in previous studies on descriptors-based regionalization with a seamless calibration scheme. The NSE scores of HDA-PR with ANN in temporal validation surpass 0.7 for both calibration setups, which is a fairly good performance compared to the reference benchmarks of around 0.78 (obtained in local calibration), thereby establishing its remarkable capability in challenging modeling scenarios (i.e., upstream calibration) as well as its capacity to collapse to a simpler mapping in less challenging ones (i.e., downstream calibration). Various flood event signatures are also used as validation metrics to demonstrate the robustness of HDA-PR, where the two regionalization methods using descriptors outperform the uniform regionalization method. Particularly, the ANN-based mapping yields the best performance for all validation scenarios based on flood event signatures. Interestingly, the ANN enables extracting information from physical descriptors and learning non-linear multivariate descriptors-to-parameters mappings while providing better model controllability than a linear mapping for complex calibration cases. The exploration of more complex dataset and neural networks architecture, on top of other differentiable flow models, represent a very promising axis for further research.}

This research and the proposed algorithms open several perspectives. Immediate work focuses on: (i) the testing and improvement of HDA-PR for application at national scales and on other continents, \rev{such as its operational application to estimate the regional hydrological model (at a 15-min temporal resolution) for the national flash flood early warning system called Vigicrues Flash in France \cite{javelle2019flash, piotte2020}}; (ii) study of effective descriptor selection along with multi-gauge cost functions explicitly accounting for data uncertainties, and optimal spatial clustering of regional controls, for example into HRU; (iii) %study of a global Bayesian estimator to improve the first-guess determination, especially with the multi-polynomial mapping.
\rev{information selection with a signature-based cost function \cite{HUYNH2023signatures} and  spatial weighting strategies for multi-site discharge over river networks; (iv) implementation of a framework to quantify predictive uncertainty at both gauged and ungauged sites}.

HDA-PR can be extended to state and composite parameters-states optimization which could be very interesting for multi-scale data assimilation and real time model correction from multi-source and multi-site data. Adding a learnable \rev{feature extraction layer fom images (CNN) or time series (LSTM)} on top of the regionalization transfer functions would enable the exploration of larger \rev{and richer} databases including time \rev{ varying} data.
Finally, HDA-PR is transposable to regionalization of differentiable integrated hydrological-hydraulic networks models (e.g., \citeA{Pujol-gmd-2022}) and could be used to explore regionalization potential from cocktails of in-situ and satellite data, including the forthcoming SWOT (Surface Water and Ocean Topography satellite mission) observations of water surfaces variabilities of worldwide larger rivers. In general, its applicability extends beyond hydrological models and can be adapted to other geophysical models.

\appendix

\section{Metrics}
\label{appd:metrics}

Let $\boldsymbol{Q}(t)$ and $\boldsymbol{Q}^*(t)$ be the simulated and observed discharge time series. The hydrological cost functions studied are:

\begin{itemize}
    \item observation cost function based on the Nash-Sutcliffe Efficiency (NSE):
    \begin{equation*}
    1-NSE = \frac{{\sum_{t=t^{*}}^{T}\left(\boldsymbol{Q}^*(t)-\boldsymbol{Q}(t)\right)^{2}}}{{\sum_{t=t^{*}}^{T}\left(\boldsymbol{Q}^*(t)-\bar{\boldsymbol{Q}}^*\right)^{2}}}
    \end{equation*}
    \item observation cost function based on the Kling-Gupta Efficiency (KGE):
    \begin{equation*}
    1-KGE_2 = a_{1}\left(r\left(\boldsymbol{Q}^*(t),\boldsymbol{Q}(t)\right)-1\right)^{2}+a_{2}\left(\beta\left(\boldsymbol{Q}^*(t),\boldsymbol{Q}(t)\right)-1\right)^{2}+a_{3}\left(\alpha\left(\boldsymbol{Q}^*(t),\boldsymbol{Q}(t)\right)-1\right)^{2} 
    \end{equation*}
    with $r$, $\beta$ and $\alpha$ being respectively measures of the correlation, bias and variability of observation with respect to simulated discharge time series; $\sum_{i=1}^{3} a_i=1$. This function is quadratic and differentiable.
\end{itemize}

\section{Incorporating ANN into the Differentiable Hydrological Model}
\label{appd:ann}

This appendix details the neural network design and the derivation of hydrological cost gradients for the ANN-based regionalization algorithm.

\begin{figure}[ht!]
\noindent\includegraphics[width=\textwidth]{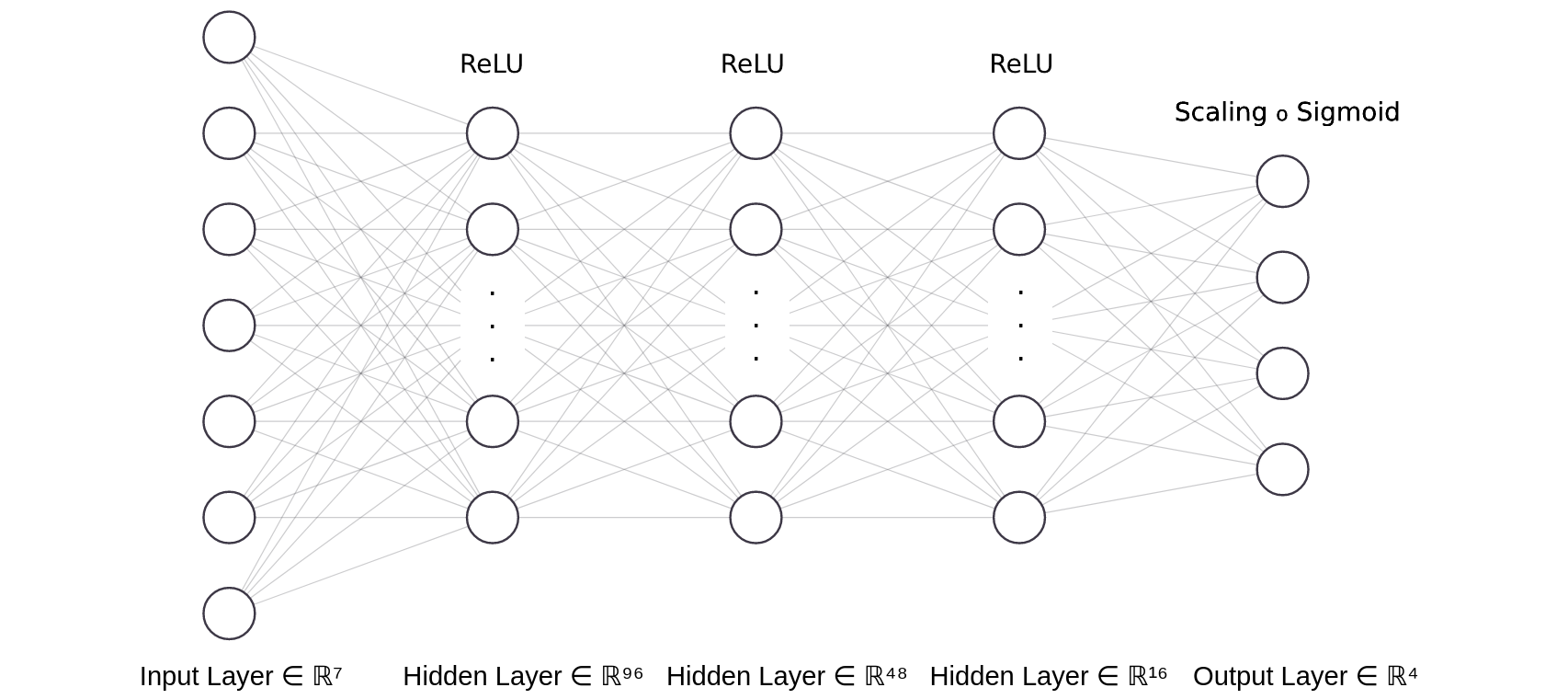}
\caption{The architecture of the ANN consists of three hidden layers followed by the ReLU activation function and an output layer that uses the Sigmoid activation function in combination with a scaling function. In this particular case, we have $N_D=7$, $N_L=4$ and $N_{\theta}=4$. The total number of trainable parameters is calculated in Table~\ref{tab:cal_number_param_ann}.}
\label{fig:ann}
\end{figure}

A simple ANN denoted $\mathcal{N}$, consisting of $N_L$ fully connected (dense) layers, intends to learn the descriptors-to-parameters field mapping in the 2D spatial domain, from $\boldsymbol{D}(x) \in \mathbb{R}^{N_D}$ to $\boldsymbol{\theta}(x) \in \mathbb{R}^{N_{\theta}}, \forall x \in \Omega$ (Figure~\ref{fig:ann}).

\begin{table}[ht!]
\caption{Number of parameters of the ANN where $N_D=7$, $N_L=4$ and $N_{\theta}=4$.}
\label{tab:cal_number_param_ann}
%\scalebox{0.9}{
\begin{tabular}{ccccc}
    \hline
    {} & Hidden layer 1 & Hidden layer 2 & Hidden layer 3 & Output layer\\
    \hline
     Input shape & $(N_D, \;)$ & $(96, \;)$ & $(48, \;)$ & $(16, \;)$\\
     Number of neurons & $96$ & $48$ & $16$ & $N_{\theta}$\\
     Number of parameters & $N_D.96 + 96 = 768$ & $96.48 + 48 = 4656$ & $48.16+16 = 784$ & $16.N_{\theta} + N_{\theta} = 68$\\
    \hline
    \multicolumn{5}{r}{Total parameters: $6276$.}
\end{tabular}
%}
\end{table}

Let us consider an ensemble of layers where each layer is associated with its weight $\boldsymbol{W}_j$ and bias $\boldsymbol{b}_j$. Then, an input $I$ of each layer is mapped to the input of the next layer by a linear function $\phi_j(I) = \boldsymbol{W}_j I + \boldsymbol{b}_j$, and followed by the ReLU activation function denoted $\delta$, except for the last layer, which is followed by the Sigmoid activation function denoted $\sigma$, ensuring that its outputs are between 0 and 1.
Now an output $\boldsymbol{O}_x=\sigma\circ\phi_{N_L}(., x) \in [0,1]^{N_{\theta}}$ of the last layer is mapped to the range of the hydrological model parameters by a differentiable scaling function $s$:
\begin{equation}\label{app:eq:scaling}
    \boldsymbol{\theta}(x) 
    = s(\boldsymbol{O}_x) 
    = l + (u - l) \odot \boldsymbol{O}_x
\end{equation}
where $l=\left(l_1,...,l_{N_\theta}\right)$ and $u=\left(u_1,...,u_{N_\theta}\right)$ with the lower and upper bounds $l_{k}\in\mathbb{R}$ and $u_{k}\in\mathbb{R}$, assumed spatially uniform, defining the bound constraints of $\theta_k(x), \forall (k, x) \in [1..N_{\theta}] \times \Omega$, in the direct hydrological model. The notation ``$\odot$'' denotes the Hadamard product. 
Noting $\Psi_j \equiv 
\begin{cases}
    \delta \circ \phi_j \text{, } j=1..N_L-1 \\
    \sigma \circ \phi_j \text{, } j=N_L
\end{cases}$, the forward propagation of the neural network $\mathcal{N}$ is defined as Equation~\ref{eq:forward_propa}.
\begin{equation}\label{eq:forward_propa}
    \boldsymbol{\theta}(x) = \mathcal{N}(\boldsymbol{D}(x), .) = s \circ \Psi_{N_L} \circ \Psi_{N_L-1} \circ ... \circ \Psi_{1}(\boldsymbol{D}(x)), \forall x \in \Omega.  
\end{equation}
Here, the notation ``$\circ$'' denotes the function composition operator.

Recall that our objective is the calibration problem of Equation~\ref{eq:general inv pb} with respect to the regional control vector $\boldsymbol{\rho} \coloneqq \left[\boldsymbol{W},\boldsymbol{b}\right]$, using the cost function of Equation~\ref{eq:cost-ann}. In such manner, different variants of stochastic gradient descent algorithm are used and thus require the gradients of the cost function with respect to the weights and biases $\frac{\partial J}{\partial \boldsymbol{\rho}_j}$ for each layer, where $\boldsymbol{\rho}_j \coloneqq [\boldsymbol{W}_j, \boldsymbol{b}_j]$. Since the forward model $\mathcal{M}\equiv \mathcal{M}_{rr}\left(.,\mathcal{N}\left(.\right)\right)$ with $\boldsymbol{\theta}$ being both the output of $\mathcal{N}$ and the input of $\mathcal{M}_{rr}$, we can write $\frac{\partial J}{\partial \boldsymbol{\rho}_j}= \frac{\partial J}{\partial \boldsymbol{\theta}} \frac{\partial \boldsymbol{\theta}}{\partial \boldsymbol{\rho}_j}$. Then these two gradients are obtained as follows:
\begin{itemize}
    \item The gradients $\frac{\partial J}{\partial \boldsymbol{\theta}}$ of the cost function with respect to the hydrological model parameters, computed by solving the numerical adjoint model of $\mathcal{M}_{rr}$; 
    \item The gradients $\frac{\partial \boldsymbol{\theta}}{\partial \boldsymbol{\rho}_j}$ of the network output with respect to the weight and bias, computed using the chain rule of composite functions of $\mathcal{N}$.
\end{itemize}
Eventually, the backward propagation for updating the weights and biases, using for instance Adam optimizer, is described in Algorithm~\ref{algo-back-prog}.

\begin{algorithm}[ht!]
\caption{Adapted back-propagation using Adam optimizer}
\label{algo-back-prog}
\begin{flushleft}
$\vartriangleright$ Randomly initialized weights and biases 
$\boldsymbol{\rho}^{(0)}=\left(\boldsymbol{\rho}^{(0)}_1, ..., \boldsymbol{\rho}^{(0)}_{N_L}\right)$ \\
$\vartriangleright$ Number of training epochs $N_{epo}$ 
\end{flushleft}
\begin{algorithmic}
\FOR {$i=1..N_{epo}$}
  \STATE $\vartriangleright$ Forward propagation over the spatial domain
  $
  \boldsymbol{\theta} \gets \left[\left(\mathcal{N}\left(\boldsymbol{D}(x), \boldsymbol{\rho}^{(i-1)}\right)\right)_{x \in \Omega}\right]^T
  $
  \STATE $\vartriangleright$ Initial gradient accumulation
  $
    \nabla A \gets \nabla_{\boldsymbol{\theta}} J = \left(\frac{\partial J}{\partial \theta_1}, ...,  \frac{\partial J}{\partial \theta_{N_{\theta}}}\right)
  $
  \FOR {$j=N_L..1$}
    \STATE $\vartriangleright$ Gradient computation
    $
    \frac{\partial J}{\partial \boldsymbol{\rho}_j} \gets \left(\frac{\partial \boldsymbol{\theta}}{\partial \boldsymbol{\rho}_j}\right)^T \nabla A
    $
    \STATE $\vartriangleright$ Updated gradient accumulation
    $
    \nabla A \gets \nabla A . \left[W_j^{(i-1)}\right]^T
    $
    \STATE $\vartriangleright$ Updated weights and biases
    $
        \boldsymbol{\rho}^{(i)}_j \gets \boldsymbol{\rho}^{(i-1)}_j - \eta \frac{m^{(i)}}{(1-\beta_1)\left(\sqrt{\frac{v^{(i)}}{1-\beta_2}}+\epsilon\right)}
    $ 
    where:\\
    $m^{(i)} \gets \beta_1 m^{(i-1)} + (1-\beta_1)\frac{\partial J}{\partial \boldsymbol{\rho}_j}\left(\boldsymbol{\rho}_j^{(i-1)}\right)$ \\
    $v^{(i)} \gets \beta_2 v^{(i-1)} + (1-\beta_2)\left(\frac{\partial J}{\partial \boldsymbol{\rho}_j}\left(\boldsymbol{\rho}_j^{(i-1)}\right)\right)^2$ \\ 
    $\beta_1=0.9$ and $\beta_2=0.999$ are the decay rates for first and second moments of gradients \\ 
    $\epsilon=10^{-8}$ is a small scalar \\
    $\eta$ is the learning rate that is a tuning parameter determining the step size of the optimization problem
  \ENDFOR
\ENDFOR 
\end{algorithmic}
\end{algorithm}

%%%%%%%%%%%%%%%%%%%%%%%%%%%%%%%%%%%%%%%%%%%%%%%
% Optional Glossary, Notation or Acronym section goes here:
%
% Glossary is only allowed in Reviews of Geophysics
%  \begin{glossary}
%  \term{Term}
%   Term Definition here
%  \term{Term}
%   Term Definition here
%  \term{Term}
%   Term Definition here
%  \end{glossary}

%%%%%%%%%%%%%%%%%%%%%%%%%%%%%%%%%%%%%%%%%%%%%%%
% Acronyms
%% NOTE that acronyms in the final published version will be spelled out when used in figure captions.
%   \begin{acronyms}
%   \acro{Acronym}
%   Definition here
%   \acro{EMOS}
%   Ensemble model output statistics
%   \acro{ECMWF}
%   Centre for Medium-Range Weather Forecasts
%   \end{acronyms}

%%%%%%%%%%%%%%%%%%%%%%%%%%%%%%%%%%%%%%%%%%%%%%%
% Notation
%   \begin{notation}
%   \notation{$a+b$} Notation Definition here
%   \notation{$e=mc^2$}
%   Equation in German-born physicist Albert Einstein's theory of special
%  relativity that showed that the increased relativistic mass ($m$) of a
%  body comes from the energy of motion of the body—that is, its kinetic
%  energy ($E$)—divided by the speed of light squared ($c^2$).
%   \end{notation}

% \newpage
% ~\newpage

%%%%%%%%%%%%%%%%%%%%%%%%%%%%%%%%%%%%%%%%%%%%%%%
%
% DATA SECTION and ACKNOWLEDGMENTS
%
%%%%%%%%%%%%%%%%%%%%%%%%%%%%%%%%%%%%%%%%%%%%%%%

\section*{Open Research}
\noindent\textit{Data Availability Statement.}
The dataset, \rev{Version 0.2}, that supports this study comprise preprocessed data sourced from SCHAPI-DGPR and Météo-France, and available at \url{https://doi.org/10.5281/zenodo.10901472} \cite{huynh_2024_10901472}. \\
\noindent\textit{Software Availability Statement.}
The proposed algorithms in the study were implemented into the SMASH source code, \rev{Version 1.0.0-rc2.21}, which is preserved at \url{https://doi.org/10.5281/zenodo.8219280} \cite{colleoni_2024_10991739}, available via GNU-3 license and developed openly at \url{https://github.com/DassHydro/smash}. Additionally, the code for conducting the numerical experiments and analysis, \rev{Version 0.2}, is preserved at \url{https://doi.org/10.5281/zenodo.11049351} \cite{ngo_nghi_truyen_huynh_2024_11049351}, available via MIT license and developed openly at \url{https://github.com/nghi-truyen/Regionalization-Learning}.

\acknowledgments
The authors greatly acknowledge Killian Pujol-Nicolas for his contribution in the preliminary stage of this work; 
Etienne Leblois from INRAE Riverly (Lyon) for fine terrain elevation processing at multiple scale over French territory.
This work was supported by funding from SCHAPI-DGPR, ANR grant ANR-21-CE04-0021-01 (MUFFINS project, ``MUltiscale Flood Forecasting with INnovating Solutions''), and NEPTUNE European project DG-ECO.

%%%%%%%%%%%%%%%%%%%%%%%%%%%%%%%%%%%%%%%%%%%%%%%
% REFERENCES and BIBLIOGRAPHY
%
% \bibliography{<name of your .bib file>} don't specify the file extension
% don't specify bibliographystyle
%
%%%%%%%%%%%%%%%%%%%%%%%%%%%%%%%%%%%%%%%%%%%%%%%

\bibliography{references}

\begin{thebibliography}{}

\bibitem [\protect \citeauthoryear {%
Abdulla%
\ \BBA {} Lettenmaier%
}{%
Abdulla%
\ \BBA {} Lettenmaier%
}{%
{\protect \APACyear {1997}}%
}]{%
abdulla1997development}
\APACinsertmetastar {%
abdulla1997development}%
\begin{APACrefauthors}%
Abdulla, F\BPBI A.%
\BCBT {}\ \BBA {} Lettenmaier, D\BPBI P.%
\end{APACrefauthors}%
\unskip\
\newblock
\APACrefYearMonthDay{1997}{}{}.
\newblock
{\BBOQ}\APACrefatitle {Development of regional parameter estimation equations
  for a macroscale hydrologic model} {Development of regional parameter
  estimation equations for a macroscale hydrologic model}.{\BBCQ}
\newblock
\APACjournalVolNumPages{Journal of hydrology}{197}{1-4}{230--257}.
\PrintBackRefs{\CurrentBib}

\bibitem [\protect \citeauthoryear {%
Althoff%
, Rodrigues%
\BCBL {}\ \BBA {} da Silva%
}{%
Althoff%
\ \protect \BOthers {.}}{%
{\protect \APACyear {2021}}%
}]{%
althoff2021addressing}
\APACinsertmetastar {%
althoff2021addressing}%
\begin{APACrefauthors}%
Althoff, D.%
, Rodrigues, L\BPBI N.%
\BCBL {}\ \BBA {} da Silva, D\BPBI D.%
\end{APACrefauthors}%
\unskip\
\newblock
\APACrefYearMonthDay{2021}{}{}.
\newblock
{\BBOQ}\APACrefatitle {Addressing hydrological modeling in watersheds under
  land cover change with deep learning} {Addressing hydrological modeling in
  watersheds under land cover change with deep learning}.{\BBCQ}
\newblock
\APACjournalVolNumPages{Advances in Water Resources}{154}{}{103965}.
\PrintBackRefs{\CurrentBib}

\bibitem [\protect \citeauthoryear {%
Bastola%
, Ishidaira%
\BCBL {}\ \BBA {} Takeuchi%
}{%
Bastola%
\ \protect \BOthers {.}}{%
{\protect \APACyear {2008}}%
}]{%
BASTOLA2008188}
\APACinsertmetastar {%
BASTOLA2008188}%
\begin{APACrefauthors}%
Bastola, S.%
, Ishidaira, H.%
\BCBL {}\ \BBA {} Takeuchi, K.%
\end{APACrefauthors}%
\unskip\
\newblock
\APACrefYearMonthDay{2008}{}{}.
\newblock
{\BBOQ}\APACrefatitle {Regionalisation of hydrological model parameters under
  parameter uncertainty: A case study involving TOPMODEL and basins across the
  globe} {Regionalisation of hydrological model parameters under parameter
  uncertainty: A case study involving topmodel and basins across the
  globe}.{\BBCQ}
\newblock
\APACjournalVolNumPages{Journal of Hydrology}{357}{3}{188-206}.
\newblock
\begin{APACrefDOI} \doi{10.1016/j.jhydrol.2008.05.007} \end{APACrefDOI}
\PrintBackRefs{\CurrentBib}

\bibitem [\protect \citeauthoryear {%
Beck%
\ \protect \BOthers {.}}{%
Beck%
\ \protect \BOthers {.}}{%
{\protect \APACyear {2020}}%
}]{%
beck2020global}
\APACinsertmetastar {%
beck2020global}%
\begin{APACrefauthors}%
Beck, H\BPBI E.%
, Pan, M.%
, Lin, P.%
, Seibert, J.%
, van Dijk, A\BPBI I.%
\BCBL {}\ \BBA {} Wood, E\BPBI F.%
\end{APACrefauthors}%
\unskip\
\newblock
\APACrefYearMonthDay{2020}{}{}.
\newblock
{\BBOQ}\APACrefatitle {Global fully distributed parameter regionalization based
  on observed streamflow from 4,229 headwater catchments} {Global fully
  distributed parameter regionalization based on observed streamflow from 4,229
  headwater catchments}.{\BBCQ}
\newblock
\APACjournalVolNumPages{Journal of Geophysical Research:
  Atmospheres}{125}{17}{e2019JD031485}.
\PrintBackRefs{\CurrentBib}

\bibitem [\protect \citeauthoryear {%
Beck%
\ \protect \BOthers {.}}{%
Beck%
\ \protect \BOthers {.}}{%
{\protect \APACyear {2016}}%
}]{%
beck2016global}
\APACinsertmetastar {%
beck2016global}%
\begin{APACrefauthors}%
Beck, H\BPBI E.%
, van Dijk, A\BPBI I.%
, De~Roo, A.%
, Miralles, D\BPBI G.%
, McVicar, T\BPBI R.%
, Schellekens, J.%
\BCBL {}\ \BBA {} Bruijnzeel, L\BPBI A.%
\end{APACrefauthors}%
\unskip\
\newblock
\APACrefYearMonthDay{2016}{}{}.
\newblock
{\BBOQ}\APACrefatitle {Global-scale regionalization of hydrologic model
  parameters} {Global-scale regionalization of hydrologic model
  parameters}.{\BBCQ}
\newblock
\APACjournalVolNumPages{Water Resources Research}{52}{5}{3599--3622}.
\PrintBackRefs{\CurrentBib}

\bibitem [\protect \citeauthoryear {%
Beven%
}{%
Beven%
}{%
{\protect \APACyear {2001}}%
}]{%
Beven_hess_2001}
\APACinsertmetastar {%
Beven_hess_2001}%
\begin{APACrefauthors}%
Beven, K.%
\end{APACrefauthors}%
\unskip\
\newblock
\APACrefYearMonthDay{2001}{}{}.
\newblock
{\BBOQ}\APACrefatitle {How far can we go in distributed hydrological
  modelling?} {How far can we go in distributed hydrological modelling?}{\BBCQ}
\newblock
\APACjournalVolNumPages{Hydrology and Earth System Sciences}{5}{1}{1--12}.
\newblock
\begin{APACrefDOI} \doi{10.5194/hess-5-1-2001} \end{APACrefDOI}
\PrintBackRefs{\CurrentBib}

\bibitem [\protect \citeauthoryear {%
Bl{\"o}schl%
, Sivapalan%
, Wagener%
, Savenije%
\BCBL {}\ \BBA {} Viglione%
}{%
Bl{\"o}schl%
\ \protect \BOthers {.}}{%
{\protect \APACyear {2013}}%
}]{%
bloschl2013runoff}
\APACinsertmetastar {%
bloschl2013runoff}%
\begin{APACrefauthors}%
Bl{\"o}schl, G.%
, Sivapalan, M.%
, Wagener, T.%
, Savenije, H.%
\BCBL {}\ \BBA {} Viglione, A.%
\end{APACrefauthors}%
\unskip\
\newblock
\APACrefYear{2013}.
\newblock
\APACrefbtitle {Runoff prediction in ungauged basins: synthesis across
  processes, places and scales} {Runoff prediction in ungauged basins:
  synthesis across processes, places and scales}.
\newblock
\APACaddressPublisher{}{Cambridge University Press}.
\PrintBackRefs{\CurrentBib}

\bibitem [\protect \citeauthoryear {%
Boeing%
\ \protect \BOthers {.}}{%
Boeing%
\ \protect \BOthers {.}}{%
{\protect \APACyear {2022}}%
}]{%
Boieng-hess-26-5137-2022}
\APACinsertmetastar {%
Boieng-hess-26-5137-2022}%
\begin{APACrefauthors}%
Boeing, F.%
, Rakovec, O.%
, Kumar, R.%
, Samaniego, L.%
, Schr\"on, M.%
, Hildebrandt, A.%
\BDBL {}Marx, A.%
\end{APACrefauthors}%
\unskip\
\newblock
\APACrefYearMonthDay{2022}{}{}.
\newblock
{\BBOQ}\APACrefatitle {High-resolution drought simulations and comparison to
  soil moisture observations in Germany} {High-resolution drought simulations
  and comparison to soil moisture observations in germany}.{\BBCQ}
\newblock
\APACjournalVolNumPages{Hydrology and Earth System
  Sciences}{26}{19}{5137--5161}.
\newblock
\begin{APACrefURL} \url{https://hess.copernicus.org/articles/26/5137/2022/}
  \end{APACrefURL}
\newblock
\begin{APACrefDOI} \doi{10.5194/hess-26-5137-2022} \end{APACrefDOI}
\PrintBackRefs{\CurrentBib}

\bibitem [\protect \citeauthoryear {%
Caruso%
, Guillot%
\BCBL {}\ \BBA {} Arnaud%
}{%
Caruso%
\ \protect \BOthers {.}}{%
{\protect \APACyear {2013}}%
}]{%
caruso2013notice}
\APACinsertmetastar {%
caruso2013notice}%
\begin{APACrefauthors}%
Caruso, A.%
, Guillot, A.%
\BCBL {}\ \BBA {} Arnaud, P.%
\end{APACrefauthors}%
\unskip\
\newblock
\APACrefYearMonthDay{2013}{}{}.
\newblock
{\BBOQ}\APACrefatitle {Notice sur les indices de confiance de la m{\'e}thode
  Shyreg-D{\'e}bit--D{\'e}finitions et Calculs} {Notice sur les indices de
  confiance de la m{\'e}thode shyreg-d{\'e}bit--d{\'e}finitions et
  calculs}.{\BBCQ}
\newblock
\BIn{} \APACrefbtitle {Aix en Provence: IRSTEA, Convention DGPR/SNRH.} {Aix en
  provence: Irstea, convention dgpr/snrh.}
\PrintBackRefs{\CurrentBib}

\bibitem [\protect \citeauthoryear {%
Castaings%
, Dartus%
, Le~Dimet%
\BCBL {}\ \BBA {} Saulnier%
}{%
Castaings%
\ \protect \BOthers {.}}{%
{\protect \APACyear {2009}}%
}]{%
Castaings-2009}
\APACinsertmetastar {%
Castaings-2009}%
\begin{APACrefauthors}%
Castaings, W.%
, Dartus, D.%
, Le~Dimet, F\BHBI X.%
\BCBL {}\ \BBA {} Saulnier, G\BHBI M.%
\end{APACrefauthors}%
\unskip\
\newblock
\APACrefYearMonthDay{2009}{}{}.
\newblock
{\BBOQ}\APACrefatitle {Sensitivity analysis and parameter estimation for
  distributed hydrological modeling: potential of variational methods}
  {Sensitivity analysis and parameter estimation for distributed hydrological
  modeling: potential of variational methods}.{\BBCQ}
\newblock
\APACjournalVolNumPages{Hydrology and Earth System Sciences}{13}{4}{503--517}.
\newblock
\begin{APACrefDOI} \doi{10.5194/hess-13-503-2009} \end{APACrefDOI}
\PrintBackRefs{\CurrentBib}

\bibitem [\protect \citeauthoryear {%
Champeaux%
\ \protect \BOthers {.}}{%
Champeaux%
\ \protect \BOthers {.}}{%
{\protect \APACyear {2009}}%
}]{%
champeaux2009mesures}
\APACinsertmetastar {%
champeaux2009mesures}%
\begin{APACrefauthors}%
Champeaux, J\BHBI L.%
, Dupuy, P.%
, Laurantin, O.%
, Soulan, I.%
, Tabary, P.%
\BCBL {}\ \BBA {} Soubeyroux, J\BHBI M.%
\end{APACrefauthors}%
\unskip\
\newblock
\APACrefYearMonthDay{2009}{}{}.
\newblock
{\BBOQ}\APACrefatitle {Les mesures de pr{\'e}cipitations et l'estimation des
  lames d'eau {\`a} M{\'e}t{\'e}o-France: {\'e}tat de l'art et perspectives}
  {Les mesures de pr{\'e}cipitations et l'estimation des lames d'eau {\`a}
  m{\'e}t{\'e}o-france: {\'e}tat de l'art et perspectives}.{\BBCQ}
\newblock
\APACjournalVolNumPages{La Houille Blanche}{}{5}{28--34}.
\PrintBackRefs{\CurrentBib}

\bibitem [\protect \citeauthoryear {%
Colleoni%
, Garambois%
, Javelle%
, Jay-Allemand%
\BCBL {}\ \BBA {} Arnaud%
}{%
Colleoni%
\ \protect \BOthers {.}}{%
{\protect \APACyear {2022}}%
}]{%
colleoni2022adjoint}
\APACinsertmetastar {%
colleoni2022adjoint}%
\begin{APACrefauthors}%
Colleoni, F.%
, Garambois, P\BHBI A.%
, Javelle, P.%
, Jay-Allemand, M.%
\BCBL {}\ \BBA {} Arnaud, P.%
\end{APACrefauthors}%
\unskip\
\newblock
\APACrefYearMonthDay{2022}{}{}.
\newblock
{\BBOQ}\APACrefatitle {Adjoint-based spatially distributed calibration of a
  grid GR-based parsimonious hydrological model over 312 French catchments with
  SMASH platform} {Adjoint-based spatially distributed calibration of a grid
  gr-based parsimonious hydrological model over 312 french catchments with
  smash platform}.{\BBCQ}
\newblock
\APACjournalVolNumPages{EGUsphere}{2022}{}{1--37}.
\newblock
\begin{APACrefDOI} \doi{10.5194/egusphere-2022-506} \end{APACrefDOI}
\PrintBackRefs{\CurrentBib}

\bibitem [\protect \citeauthoryear {%
Colleoni%
\ \protect \BOthers {.}}{%
Colleoni%
\ \protect \BOthers {.}}{%
{\protect \APACyear {2024}}%
}]{%
colleoni_2024_10991739}
\APACinsertmetastar {%
colleoni_2024_10991739}%
\begin{APACrefauthors}%
Colleoni, F.%
, Huynh, N\BPBI N\BPBI T.%
, Benjamin, R.%
, De~Fournas, T.%
, El~Baz, A.%
\BCBL {}\ \BBA {} Garambois, P\BHBI A.%
\end{APACrefauthors}%
\unskip\
\newblock
\APACrefYearMonthDay{2024}{{\APACmonth{04}}}{}.
\newblock
\APACrefbtitle {SMASH: Version 1.0.0-rc2.21.} {Smash: Version 1.0.0-rc2.21.}
\newblock
\APACaddressPublisher{}{Zenodo}.
\newblock
\begin{APACrefURL} \url{https://doi.org/10.5281/zenodo.8219280}
  \end{APACrefURL}
\newblock
\begin{APACrefDOI} \doi{10.5281/zenodo.8219280} \end{APACrefDOI}
\PrintBackRefs{\CurrentBib}

\bibitem [\protect \citeauthoryear {%
Coxon%
\ \protect \BOthers {.}}{%
Coxon%
\ \protect \BOthers {.}}{%
{\protect \APACyear {2019}}%
}]{%
coxon2019}
\APACinsertmetastar {%
coxon2019}%
\begin{APACrefauthors}%
Coxon, G.%
, Freer, J.%
, Lane, R.%
, Dunne, T.%
, Knoben, W\BPBI J\BPBI M.%
, Howden, N\BPBI J\BPBI K.%
\BDBL {}Woods, R.%
\end{APACrefauthors}%
\unskip\
\newblock
\APACrefYearMonthDay{2019}{}{}.
\newblock
{\BBOQ}\APACrefatitle {DECIPHeR v1: Dynamic fluxEs and ConnectIvity for
  Predictions of HydRology} {Decipher v1: Dynamic fluxes and connectivity for
  predictions of hydrology}.{\BBCQ}
\newblock
\APACjournalVolNumPages{Geoscientific Model Development}{12}{6}{2285--2306}.
\newblock
\begin{APACrefDOI} \doi{10.5194/gmd-12-2285-2019} \end{APACrefDOI}
\PrintBackRefs{\CurrentBib}

\bibitem [\protect \citeauthoryear {%
Dagum%
\ \BBA {} Menon%
}{%
Dagum%
\ \BBA {} Menon%
}{%
{\protect \APACyear {1998}}%
}]{%
dagum98}
\APACinsertmetastar {%
dagum98}%
\begin{APACrefauthors}%
Dagum, L.%
\BCBT {}\ \BBA {} Menon, R.%
\end{APACrefauthors}%
\unskip\
\newblock
\APACrefYearMonthDay{1998}{}{}.
\newblock
{\BBOQ}\APACrefatitle {OpenMP: an industry standard API for shared-memory
  programming} {Openmp: an industry standard api for shared-memory
  programming}.{\BBCQ}
\newblock
\APACjournalVolNumPages{IEEE Computational Science and
  Engineering}{5}{1}{46-55}.
\newblock
\begin{APACrefDOI} \doi{10.1109/99.660313} \end{APACrefDOI}
\PrintBackRefs{\CurrentBib}

\bibitem [\protect \citeauthoryear {%
De~Lavenne%
, Andr{\'e}assian%
, Thirel%
, Ramos%
\BCBL {}\ \BBA {} Perrin%
}{%
De~Lavenne%
\ \protect \BOthers {.}}{%
{\protect \APACyear {2019}}%
}]{%
de2019regularization}
\APACinsertmetastar {%
de2019regularization}%
\begin{APACrefauthors}%
De~Lavenne, A.%
, Andr{\'e}assian, V.%
, Thirel, G.%
, Ramos, M\BHBI H.%
\BCBL {}\ \BBA {} Perrin, C.%
\end{APACrefauthors}%
\unskip\
\newblock
\APACrefYearMonthDay{2019}{}{}.
\newblock
{\BBOQ}\APACrefatitle {A regularization approach to improve the sequential
  calibration of a semidistributed hydrological model} {A regularization
  approach to improve the sequential calibration of a semidistributed
  hydrological model}.{\BBCQ}
\newblock
\APACjournalVolNumPages{Water Resources Research}{55}{11}{8821--8839}.
\PrintBackRefs{\CurrentBib}

\bibitem [\protect \citeauthoryear {%
Duan%
, Sorooshian%
\BCBL {}\ \BBA {} Gupta%
}{%
Duan%
\ \protect \BOthers {.}}{%
{\protect \APACyear {1992}}%
}]{%
Duan1992_SCE}
\APACinsertmetastar {%
Duan1992_SCE}%
\begin{APACrefauthors}%
Duan, Q.%
, Sorooshian, S.%
\BCBL {}\ \BBA {} Gupta, V.%
\end{APACrefauthors}%
\unskip\
\newblock
\APACrefYearMonthDay{1992}{}{}.
\newblock
{\BBOQ}\APACrefatitle {Effective and efficient global optimization for
  conceptual rainfall-runoff models} {Effective and efficient global
  optimization for conceptual rainfall-runoff models}.{\BBCQ}
\newblock
\APACjournalVolNumPages{Water Resources Research}{28}{4}{1015-1031}.
\newblock
\begin{APACrefDOI} \doi{10.1029/91WR02985} \end{APACrefDOI}
\PrintBackRefs{\CurrentBib}

\bibitem [\protect \citeauthoryear {%
Fablet%
\ \protect \BOthers {.}}{%
Fablet%
\ \protect \BOthers {.}}{%
{\protect \APACyear {2021}}%
}]{%
fablet2021learning}
\APACinsertmetastar {%
fablet2021learning}%
\begin{APACrefauthors}%
Fablet, R.%
, Chapron, B.%
, Drumetz, L.%
, M{\'e}min, E.%
, Pannekoucke, O.%
\BCBL {}\ \BBA {} Rousseau, F.%
\end{APACrefauthors}%
\unskip\
\newblock
\APACrefYearMonthDay{2021}{}{}.
\newblock
{\BBOQ}\APACrefatitle {Learning variational data assimilation models and
  solvers} {Learning variational data assimilation models and solvers}.{\BBCQ}
\newblock
\APACjournalVolNumPages{Journal of Advances in Modeling Earth
  Systems}{13}{10}{e2021MS002572}.
\PrintBackRefs{\CurrentBib}

\bibitem [\protect \citeauthoryear {%
Fekete%
\ \BBA {} V{\"o}r{\"o}smarty%
}{%
Fekete%
\ \BBA {} V{\"o}r{\"o}smarty%
}{%
{\protect \APACyear {2007}}%
}]{%
fekete2007current}
\APACinsertmetastar {%
fekete2007current}%
\begin{APACrefauthors}%
Fekete, B\BPBI M.%
\BCBT {}\ \BBA {} V{\"o}r{\"o}smarty, C\BPBI J.%
\end{APACrefauthors}%
\unskip\
\newblock
\APACrefYearMonthDay{2007}{}{}.
\newblock
{\BBOQ}\APACrefatitle {The current status of global river discharge monitoring
  and potential new technologies complementing traditional discharge
  measurements} {The current status of global river discharge monitoring and
  potential new technologies complementing traditional discharge
  measurements}.{\BBCQ}
\newblock
\APACjournalVolNumPages{IAHS publ}{309}{}{129--136}.
\PrintBackRefs{\CurrentBib}

\bibitem [\protect \citeauthoryear {%
Finke%
\ \protect \BOthers {.}}{%
Finke%
\ \protect \BOthers {.}}{%
{\protect \APACyear {1998}}%
}]{%
finke1998geo}
\APACinsertmetastar {%
finke1998geo}%
\begin{APACrefauthors}%
Finke, P.%
, Hartwich, R.%
, Dudal, R.%
, Ibanez, J.%
, Jamagne, M.%
, King, D.%
\BDBL {}Yassoglou, N.%
\end{APACrefauthors}%
\unskip\
\newblock
\APACrefYear{1998}.
\newblock
\APACrefbtitle {Geo-referenced soil database for Europe. Manual of procedures,
  version 1.0} {Geo-referenced soil database for europe. manual of procedures,
  version 1.0}.
\newblock
\APACaddressPublisher{}{European Communities}.
\PrintBackRefs{\CurrentBib}

\bibitem [\protect \citeauthoryear {%
Fortin%
, Rainville%
, Gardner%
, Parizeau%
\BCBL {}\ \BBA {} Gagn{{\'e}}%
}{%
Fortin%
\ \protect \BOthers {.}}{%
{\protect \APACyear {2012}}%
}]{%
fortin2012_DEAP}
\APACinsertmetastar {%
fortin2012_DEAP}%
\begin{APACrefauthors}%
Fortin, F\BHBI A.%
, Rainville, F\BHBI M\BPBI D.%
, Gardner, M\BHBI A.%
, Parizeau, M.%
\BCBL {}\ \BBA {} Gagn{{\'e}}, C.%
\end{APACrefauthors}%
\unskip\
\newblock
\APACrefYearMonthDay{2012}{}{}.
\newblock
{\BBOQ}\APACrefatitle {DEAP: Evolutionary Algorithms Made Easy} {Deap:
  Evolutionary algorithms made easy}.{\BBCQ}
\newblock
\APACjournalVolNumPages{Journal of Machine Learning
  Research}{13}{70}{2171--2175}.
\newblock
\begin{APACrefURL} \url{http://jmlr.org/papers/v13/fortin12a.html}
  \end{APACrefURL}
\PrintBackRefs{\CurrentBib}

\bibitem [\protect \citeauthoryear {%
Garambois%
\ \protect \BOthers {.}}{%
Garambois%
\ \protect \BOthers {.}}{%
{\protect \APACyear {2020}}%
}]{%
GARAMBOIS2020}
\APACinsertmetastar {%
GARAMBOIS2020}%
\begin{APACrefauthors}%
Garambois, P\BHBI A.%
, Larnier, K.%
, Monnier, J.%
, Finaud-Guyot, P.%
, Verley, J.%
, Montazem, A\BHBI S.%
\BCBL {}\ \BBA {} Calmant, S.%
\end{APACrefauthors}%
\unskip\
\newblock
\APACrefYearMonthDay{2020}{}{}.
\newblock
{\BBOQ}\APACrefatitle {Variational estimation of effective channel and ungauged
  anabranching river discharge from multi-satellite water heights of different
  spatial sparsity} {Variational estimation of effective channel and ungauged
  anabranching river discharge from multi-satellite water heights of different
  spatial sparsity}.{\BBCQ}
\newblock
\APACjournalVolNumPages{Journal of Hydrology}{581}{}{124409}.
\newblock
\begin{APACrefDOI} \doi{10.1016/j.jhydrol.2019.124409} \end{APACrefDOI}
\PrintBackRefs{\CurrentBib}

\bibitem [\protect \citeauthoryear {%
Garambois%
, Roux%
, Larnier%
, Labat%
\BCBL {}\ \BBA {} Dartus%
}{%
Garambois%
\ \protect \BOthers {.}}{%
{\protect \APACyear {2015}}%
}]{%
garambois2015parameter}
\APACinsertmetastar {%
garambois2015parameter}%
\begin{APACrefauthors}%
Garambois, P\BHBI A.%
, Roux, H.%
, Larnier, K.%
, Labat, D.%
\BCBL {}\ \BBA {} Dartus, D.%
\end{APACrefauthors}%
\unskip\
\newblock
\APACrefYearMonthDay{2015}{}{}.
\newblock
{\BBOQ}\APACrefatitle {Parameter regionalization for a process-oriented
  distributed model dedicated to flash floods} {Parameter regionalization for a
  process-oriented distributed model dedicated to flash floods}.{\BBCQ}
\newblock
\APACjournalVolNumPages{Journal of Hydrology}{525}{}{383--399}.
\PrintBackRefs{\CurrentBib}

\bibitem [\protect \citeauthoryear {%
Gemperline%
, Long%
\BCBL {}\ \BBA {} Gregoriou%
}{%
Gemperline%
\ \protect \BOthers {.}}{%
{\protect \APACyear {1991}}%
}]{%
gemperline1991nonlinear}
\APACinsertmetastar {%
gemperline1991nonlinear}%
\begin{APACrefauthors}%
Gemperline, P\BPBI J.%
, Long, J\BPBI R.%
\BCBL {}\ \BBA {} Gregoriou, V\BPBI G.%
\end{APACrefauthors}%
\unskip\
\newblock
\APACrefYearMonthDay{1991}{}{}.
\newblock
{\BBOQ}\APACrefatitle {Nonlinear multivariate calibration using principal
  components regression and artificial neural networks} {Nonlinear multivariate
  calibration using principal components regression and artificial neural
  networks}.{\BBCQ}
\newblock
\APACjournalVolNumPages{Analytical Chemistry}{63}{20}{2313--2323}.
\PrintBackRefs{\CurrentBib}

\bibitem [\protect \citeauthoryear {%
Glorot%
\ \BBA {} Bengio%
}{%
Glorot%
\ \BBA {} Bengio%
}{%
{\protect \APACyear {2010}}%
}]{%
glorot2010understanding}
\APACinsertmetastar {%
glorot2010understanding}%
\begin{APACrefauthors}%
Glorot, X.%
\BCBT {}\ \BBA {} Bengio, Y.%
\end{APACrefauthors}%
\unskip\
\newblock
\APACrefYearMonthDay{2010}{}{}.
\newblock
{\BBOQ}\APACrefatitle {Understanding the difficulty of training deep
  feedforward neural networks} {Understanding the difficulty of training deep
  feedforward neural networks}.{\BBCQ}
\newblock
\BIn{} \APACrefbtitle {Proceedings of the thirteenth international conference
  on artificial intelligence and statistics} {Proceedings of the thirteenth
  international conference on artificial intelligence and statistics}\ (\BPGS\
  249--256).
\PrintBackRefs{\CurrentBib}

\bibitem [\protect \citeauthoryear {%
G{\"o}tzinger%
\ \BBA {} B{\'a}rdossy%
}{%
G{\"o}tzinger%
\ \BBA {} B{\'a}rdossy%
}{%
{\protect \APACyear {2007}}%
}]{%
gotzinger2007comparison}
\APACinsertmetastar {%
gotzinger2007comparison}%
\begin{APACrefauthors}%
G{\"o}tzinger, J.%
\BCBT {}\ \BBA {} B{\'a}rdossy, A.%
\end{APACrefauthors}%
\unskip\
\newblock
\APACrefYearMonthDay{2007}{}{}.
\newblock
{\BBOQ}\APACrefatitle {Comparison of four regionalisation methods for a
  distributed hydrological model} {Comparison of four regionalisation methods
  for a distributed hydrological model}.{\BBCQ}
\newblock
\APACjournalVolNumPages{Journal of Hydrology}{333}{2-4}{374--384}.
\PrintBackRefs{\CurrentBib}

\bibitem [\protect \citeauthoryear {%
Gupta%
, Beven%
\BCBL {}\ \BBA {} Wagener%
}{%
Gupta%
\ \protect \BOthers {.}}{%
{\protect \APACyear {2006}}%
}]{%
gupta2006model}
\APACinsertmetastar {%
gupta2006model}%
\begin{APACrefauthors}%
Gupta, H\BPBI V.%
, Beven, K\BPBI J.%
\BCBL {}\ \BBA {} Wagener, T.%
\end{APACrefauthors}%
\unskip\
\newblock
\APACrefYearMonthDay{2006}{}{}.
\newblock
{\BBOQ}\APACrefatitle {Model calibration and uncertainty estimation} {Model
  calibration and uncertainty estimation}.{\BBCQ}
\newblock
\APACjournalVolNumPages{Encyclopedia of hydrological sciences}{}{}{}.
\PrintBackRefs{\CurrentBib}

\bibitem [\protect \citeauthoryear {%
Hannah%
\ \protect \BOthers {.}}{%
Hannah%
\ \protect \BOthers {.}}{%
{\protect \APACyear {2011}}%
}]{%
hannah2011large}
\APACinsertmetastar {%
hannah2011large}%
\begin{APACrefauthors}%
Hannah, D\BPBI M.%
, Demuth, S.%
, van Lanen, H\BPBI A.%
, Looser, U.%
, Prudhomme, C.%
, Rees, G.%
\BDBL {}others%
\end{APACrefauthors}%
\unskip\
\newblock
\APACrefYearMonthDay{2011}{}{}.
\newblock
{\BBOQ}\APACrefatitle {Large-scale river flow archives: importance, current
  status and future needs.} {Large-scale river flow archives: importance,
  current status and future needs.}{\BBCQ}
\newblock
\APACjournalVolNumPages{Hydrological Processes}{25}{7}{1191--1200}.
\PrintBackRefs{\CurrentBib}

\bibitem [\protect \citeauthoryear {%
Hascoet%
\ \BBA {} Pascual%
}{%
Hascoet%
\ \BBA {} Pascual%
}{%
{\protect \APACyear {2013}}%
}]{%
hascoet2013tapenade}
\APACinsertmetastar {%
hascoet2013tapenade}%
\begin{APACrefauthors}%
Hascoet, L.%
\BCBT {}\ \BBA {} Pascual, V.%
\end{APACrefauthors}%
\unskip\
\newblock
\APACrefYearMonthDay{2013}{}{}.
\newblock
{\BBOQ}\APACrefatitle {The Tapenade automatic differentiation tool: principles,
  model, and specification} {The tapenade automatic differentiation tool:
  principles, model, and specification}.{\BBCQ}
\newblock
\APACjournalVolNumPages{ACM Transactions on Mathematical Software
  (TOMS)}{39}{3}{1--43}.
\PrintBackRefs{\CurrentBib}

\bibitem [\protect \citeauthoryear {%
Hoffman%
\ \BBA {} Gelman%
}{%
Hoffman%
\ \BBA {} Gelman%
}{%
{\protect \APACyear {2014}}%
}]{%
hoffman_no-u-turn_2014}
\APACinsertmetastar {%
hoffman_no-u-turn_2014}%
\begin{APACrefauthors}%
Hoffman, M\BPBI D.%
\BCBT {}\ \BBA {} Gelman, A.%
\end{APACrefauthors}%
\unskip\
\newblock
\APACrefYearMonthDay{2014}{}{}.
\newblock
{\BBOQ}\APACrefatitle {The {No}-{U}-{Turn} {Sampler}: {Adaptively} {Setting}
  {Path} {Lengths} in {Hamiltonian} {Monte} {Carlo}} {The {No}-{U}-{Turn}
  {Sampler}: {Adaptively} {Setting} {Path} {Lengths} in {Hamiltonian} {Monte}
  {Carlo}}.{\BBCQ}
\newblock
\APACjournalVolNumPages{Journal of Machine Learning
  Research}{15}{}{1593--1623}.
\PrintBackRefs{\CurrentBib}

\bibitem [\protect \citeauthoryear {%
H{\"o}ge%
, Scheidegger%
, Baity-Jesi%
, Albert%
\BCBL {}\ \BBA {} Fenicia%
}{%
H{\"o}ge%
\ \protect \BOthers {.}}{%
{\protect \APACyear {2022}}%
}]{%
hoge2022improving}
\APACinsertmetastar {%
hoge2022improving}%
\begin{APACrefauthors}%
H{\"o}ge, M.%
, Scheidegger, A.%
, Baity-Jesi, M.%
, Albert, C.%
\BCBL {}\ \BBA {} Fenicia, F.%
\end{APACrefauthors}%
\unskip\
\newblock
\APACrefYearMonthDay{2022}{}{}.
\newblock
{\BBOQ}\APACrefatitle {Improving hydrologic models for predictions and process
  understanding using Neural ODEs} {Improving hydrologic models for predictions
  and process understanding using neural odes}.{\BBCQ}
\newblock
\APACjournalVolNumPages{Hydrology and Earth System Sciences
  Discussions}{}{}{1--29}.
\PrintBackRefs{\CurrentBib}

\bibitem [\protect \citeauthoryear {%
Horner%
\ \protect \BOthers {.}}{%
Horner%
\ \protect \BOthers {.}}{%
{\protect \APACyear {2018}}%
}]{%
horner_impact_2018}
\APACinsertmetastar {%
horner_impact_2018}%
\begin{APACrefauthors}%
Horner, I.%
, Renard, B.%
, Le~Coz, J.%
, Branger, F.%
, McMillan, H\BPBI K.%
\BCBL {}\ \BBA {} Pierrefeu, G.%
\end{APACrefauthors}%
\unskip\
\newblock
\APACrefYearMonthDay{2018}{}{}.
\newblock
{\BBOQ}\APACrefatitle {Impact of {Stage} {Measurement} {Errors} on {Streamflow}
  {Uncertainty}} {Impact of {Stage} {Measurement} {Errors} on {Streamflow}
  {Uncertainty}}.{\BBCQ}
\newblock
\APACjournalVolNumPages{Water Resources Research}{54}{3}{1952--1976}.
\newblock
\begin{APACrefDOI} \doi{doi:10.1002/2017WR022039} \end{APACrefDOI}
\PrintBackRefs{\CurrentBib}

\bibitem [\protect \citeauthoryear {%
Hrachowitz%
\ \protect \BOthers {.}}{%
Hrachowitz%
\ \protect \BOthers {.}}{%
{\protect \APACyear {2013}}%
}]{%
Hrachowitz_2013}
\APACinsertmetastar {%
Hrachowitz_2013}%
\begin{APACrefauthors}%
Hrachowitz, M.%
, Savenije, H.%
, Blöschl, G.%
, McDonnell, J.%
, Sivapalan, M.%
, Pomeroy, J.%
\BDBL {}Cudennec, C.%
\end{APACrefauthors}%
\unskip\
\newblock
\APACrefYearMonthDay{2013}{}{}.
\newblock
{\BBOQ}\APACrefatitle {A decade of Predictions in Ungauged Basins (PUB)—a
  review} {A decade of predictions in ungauged basins (pub)—a review}.{\BBCQ}
\newblock
\APACjournalVolNumPages{Hydrological Sciences Journal}{58}{6}{1198-1255}.
\newblock
\begin{APACrefDOI} \doi{10.1080/02626667.2013.803183} \end{APACrefDOI}
\PrintBackRefs{\CurrentBib}

\bibitem [\protect \citeauthoryear {%
Huang%
\ \protect \BOthers {.}}{%
Huang%
\ \protect \BOthers {.}}{%
{\protect \APACyear {2019}}%
}]{%
HUANG2019}
\APACinsertmetastar {%
HUANG2019}%
\begin{APACrefauthors}%
Huang, S.%
, Eisner, S.%
, Magnusson, J\BPBI O.%
, Lussana, C.%
, Yang, X.%
\BCBL {}\ \BBA {} Beldring, S.%
\end{APACrefauthors}%
\unskip\
\newblock
\APACrefYearMonthDay{2019}{}{}.
\newblock
{\BBOQ}\APACrefatitle {Improvements of the spatially distributed hydrological
  modelling using the HBV model at 1 km resolution for Norway} {Improvements of
  the spatially distributed hydrological modelling using the hbv model at 1 km
  resolution for norway}.{\BBCQ}
\newblock
\APACjournalVolNumPages{Journal of Hydrology}{577}{}{123585}.
\newblock
\begin{APACrefDOI} \doi{10.1016/j.jhydrol.2019.03.051} \end{APACrefDOI}
\PrintBackRefs{\CurrentBib}

\bibitem [\protect \citeauthoryear {%
Hundecha%
\ \BBA {} B{\'a}rdossy%
}{%
Hundecha%
\ \BBA {} B{\'a}rdossy%
}{%
{\protect \APACyear {2004}}%
}]{%
hundecha2004modeling}
\APACinsertmetastar {%
hundecha2004modeling}%
\begin{APACrefauthors}%
Hundecha, Y.%
\BCBT {}\ \BBA {} B{\'a}rdossy, A.%
\end{APACrefauthors}%
\unskip\
\newblock
\APACrefYearMonthDay{2004}{}{}.
\newblock
{\BBOQ}\APACrefatitle {Modeling of the effect of land use changes on the runoff
  generation of a river basin through parameter regionalization of a watershed
  model} {Modeling of the effect of land use changes on the runoff generation
  of a river basin through parameter regionalization of a watershed
  model}.{\BBCQ}
\newblock
\APACjournalVolNumPages{Journal of hydrology}{292}{1-4}{281--295}.
\PrintBackRefs{\CurrentBib}

\bibitem [\protect \citeauthoryear {%
Huynh%
}{%
Huynh%
}{%
{\protect \APACyear {2024}}%
}]{%
ngo_nghi_truyen_huynh_2024_11049351}
\APACinsertmetastar {%
ngo_nghi_truyen_huynh_2024_11049351}%
\begin{APACrefauthors}%
Huynh, N\BPBI N\BPBI T.%
\end{APACrefauthors}%
\unskip\
\newblock
\APACrefYearMonthDay{2024}{{\APACmonth{04}}}{}.
\newblock
\APACrefbtitle {Regionalization-Learning: Version 0.2.}
  {Regionalization-learning: Version 0.2.}
\newblock
\APACaddressPublisher{}{Zenodo}.
\newblock
\begin{APACrefURL} \url{https://doi.org/10.5281/zenodo.11049351}
  \end{APACrefURL}
\newblock
\begin{APACrefDOI} \doi{10.5281/zenodo.11049351} \end{APACrefDOI}
\PrintBackRefs{\CurrentBib}

\bibitem [\protect \citeauthoryear {%
Huynh%
\ \BBA {} Colleoni%
}{%
Huynh%
\ \BBA {} Colleoni%
}{%
{\protect \APACyear {2024}}%
}]{%
huynh_2024_10901472}
\APACinsertmetastar {%
huynh_2024_10901472}%
\begin{APACrefauthors}%
Huynh, N\BPBI N\BPBI T.%
\BCBT {}\ \BBA {} Colleoni, F.%
\end{APACrefauthors}%
\unskip\
\newblock
\APACrefYearMonthDay{2024}{{\APACmonth{03}}}{}.
\newblock
\APACrefbtitle {Regionalization Learning Data: Version 0.2.} {Regionalization
  learning data: Version 0.2.}
\newblock
\APACaddressPublisher{}{Zenodo}.
\newblock
\begin{APACrefURL} \url{https://doi.org/10.5281/zenodo.10901472}
  \end{APACrefURL}
\newblock
\begin{APACrefDOI} \doi{10.5281/zenodo.10901472} \end{APACrefDOI}
\PrintBackRefs{\CurrentBib}

\bibitem [\protect \citeauthoryear {%
Huynh%
, Garambois%
, Colleoni%
\BCBL {}\ \BBA {} Javelle%
}{%
Huynh%
, Garambois%
, Colleoni%
\BCBL {}\ \BBA {} Javelle%
}{%
{\protect \APACyear {2023}}%
}]{%
HUYNH2023signatures}
\APACinsertmetastar {%
HUYNH2023signatures}%
\begin{APACrefauthors}%
Huynh, N\BPBI N\BPBI T.%
, Garambois, P\BHBI A.%
, Colleoni, F.%
\BCBL {}\ \BBA {} Javelle, P.%
\end{APACrefauthors}%
\unskip\
\newblock
\APACrefYearMonthDay{2023}{}{}.
\newblock
{\BBOQ}\APACrefatitle {Signatures-and-sensitivity-based multi-criteria
  variational calibration for distributed hydrological modeling applied to
  Mediterranean floods} {Signatures-and-sensitivity-based multi-criteria
  variational calibration for distributed hydrological modeling applied to
  mediterranean floods}.{\BBCQ}
\newblock
\APACjournalVolNumPages{Journal of Hydrology}{625}{}{129992}.
\newblock
\begin{APACrefDOI} \doi{10.1016/j.jhydrol.2023.129992} \end{APACrefDOI}
\PrintBackRefs{\CurrentBib}

\bibitem [\protect \citeauthoryear {%
Huynh%
, Garambois%
, Colleoni%
, Renard%
\BCBL {}\ \BBA {} Roux%
}{%
Huynh%
, Garambois%
, Colleoni%
, Renard%
\BCBL {}\ \BBA {} Roux%
}{%
{\protect \APACyear {2023}}%
}]{%
huynh2023multigauge}
\APACinsertmetastar {%
huynh2023multigauge}%
\begin{APACrefauthors}%
Huynh, N\BPBI N\BPBI T.%
, Garambois, P\BHBI A.%
, Colleoni, F.%
, Renard, B.%
\BCBL {}\ \BBA {} Roux, H.%
\end{APACrefauthors}%
\unskip\
\newblock
\APACrefYearMonthDay{2023}{}{}.
\newblock
{\BBOQ}\APACrefatitle {Multi-gauge Hydrological Variational Data Assimilation:
  Regionalization Learning with Spatial Gradients using Multilayer Perceptron
  and Bayesian-Guided Multivariate Regression} {Multi-gauge hydrological
  variational data assimilation: Regionalization learning with spatial
  gradients using multilayer perceptron and bayesian-guided multivariate
  regression}.{\BBCQ}
\newblock
\APACjournalVolNumPages{arXiv preprint arXiv:2307.02497}{}{}{}.
\newblock
\begin{APACrefURL} \url{https://arxiv.org/abs/2307.02497} \end{APACrefURL}
\newblock
\begin{APACrefDOI} \doi{10.48550/arXiv.2307.02497} \end{APACrefDOI}
\PrintBackRefs{\CurrentBib}

\bibitem [\protect \citeauthoryear {%
Huynh%
, Garambois%
, Colleoni%
, Renard%
, Roux%
, Demargne%
\BCBL {}\ \BBA {} Javelle%
}{%
Huynh%
, Garambois%
, Colleoni%
, Renard%
, Roux%
, Demargne%
\BCBL {}\ \BBA {} Javelle%
}{%
{\protect \APACyear {2023}}%
}]{%
huynh2023learning}
\APACinsertmetastar {%
huynh2023learning}%
\begin{APACrefauthors}%
Huynh, N\BPBI N\BPBI T.%
, Garambois, P\BHBI A.%
, Colleoni, F.%
, Renard, B.%
, Roux, H.%
, Demargne, J.%
\BCBL {}\ \BBA {} Javelle, P.%
\end{APACrefauthors}%
\unskip\
\newblock
\APACrefYearMonthDay{2023}{}{}.
\newblock
{\BBOQ}\APACrefatitle {Learning Regionalization within a Differentiable
  High-Resolution Hydrological Model using Accurate Spatial Cost Gradients}
  {Learning regionalization within a differentiable high-resolution
  hydrological model using accurate spatial cost gradients}.{\BBCQ}
\newblock
\APACjournalVolNumPages{arXiv preprint arXiv:2308.02040}{}{}{}.
\newblock
\begin{APACrefURL} \url{https://arxiv.org/abs/2308.02040v1} \end{APACrefURL}
\newblock
\begin{APACrefDOI} \doi{10.48550/arXiv.2308.02040} \end{APACrefDOI}
\PrintBackRefs{\CurrentBib}

\bibitem [\protect \citeauthoryear {%
Javelle%
, Saint-Martin%
, Garandeau%
\BCBL {}\ \BBA {} Janet%
}{%
Javelle%
\ \protect \BOthers {.}}{%
{\protect \APACyear {2019}}%
}]{%
javelle2019flash}
\APACinsertmetastar {%
javelle2019flash}%
\begin{APACrefauthors}%
Javelle, P.%
, Saint-Martin, C.%
, Garandeau, L.%
\BCBL {}\ \BBA {} Janet, B.%
\end{APACrefauthors}%
\unskip\
\newblock
\APACrefYearMonthDay{2019}{}{}.
\newblock
{\BBOQ}\APACrefatitle {Flash flood warnings: Recent achievements in France with
  the national Vigicrues Flash system} {Flash flood warnings: Recent
  achievements in france with the national vigicrues flash system}.{\BBCQ}
\newblock
\APACjournalVolNumPages{United Nations Office for Disaster Risk Reduction,
  Contributing Paper to the Global Assessment Report on Disaster Risk Reduction
  (GAR 2019)}{60}{}{}.
\PrintBackRefs{\CurrentBib}

\bibitem [\protect \citeauthoryear {%
Jay-Allemand%
\ \protect \BOthers {.}}{%
Jay-Allemand%
\ \protect \BOthers {.}}{%
{\protect \APACyear {2024}}%
}]{%
jay2024spatially}
\APACinsertmetastar {%
jay2024spatially}%
\begin{APACrefauthors}%
Jay-Allemand, M.%
, Demargne, J.%
, Garambois, P\BHBI A.%
, Javelle, P.%
, Gejadze, I.%
, Colleoni, F.%
\BDBL {}Fouchier, C.%
\end{APACrefauthors}%
\unskip\
\newblock
\APACrefYearMonthDay{2024}{}{}.
\newblock
{\BBOQ}\APACrefatitle {Spatially distributed calibration of a hydrological
  model with variational optimization constrained by physiographic maps for
  flash flood forecasting in France} {Spatially distributed calibration of a
  hydrological model with variational optimization constrained by physiographic
  maps for flash flood forecasting in france}.{\BBCQ}
\newblock
\APACjournalVolNumPages{Proceedings of IAHS}{385}{}{281--290}.
\newblock
\begin{APACrefDOI} \doi{10.5194/piahs-385-281-2024} \end{APACrefDOI}
\PrintBackRefs{\CurrentBib}

\bibitem [\protect \citeauthoryear {%
Jay-Allemand%
\ \protect \BOthers {.}}{%
Jay-Allemand%
\ \protect \BOthers {.}}{%
{\protect \APACyear {2020}}%
}]{%
jay2020potential}
\APACinsertmetastar {%
jay2020potential}%
\begin{APACrefauthors}%
Jay-Allemand, M.%
, Javelle, P.%
, Gejadze, I.%
, Arnaud, P.%
, Malaterre, P\BHBI O.%
, Fine, J\BHBI A.%
\BCBL {}\ \BBA {} Organde, D.%
\end{APACrefauthors}%
\unskip\
\newblock
\APACrefYearMonthDay{2020}{}{}.
\newblock
{\BBOQ}\APACrefatitle {On the potential of variational calibration for a fully
  distributed hydrological model: application on a Mediterranean catchment} {On
  the potential of variational calibration for a fully distributed hydrological
  model: application on a mediterranean catchment}.{\BBCQ}
\newblock
\APACjournalVolNumPages{Hydrology and Earth System
  Sciences}{24}{11}{5519--5538}.
\PrintBackRefs{\CurrentBib}

\bibitem [\protect \citeauthoryear {%
Kavetski%
, Kuczera%
\BCBL {}\ \BBA {} Franks%
}{%
Kavetski%
\ \protect \BOthers {.}}{%
{\protect \APACyear {2006}}%
}]{%
kavetski2006bayesian}
\APACinsertmetastar {%
kavetski2006bayesian}%
\begin{APACrefauthors}%
Kavetski, D.%
, Kuczera, G.%
\BCBL {}\ \BBA {} Franks, S\BPBI W.%
\end{APACrefauthors}%
\unskip\
\newblock
\APACrefYearMonthDay{2006}{}{}.
\newblock
{\BBOQ}\APACrefatitle {Bayesian analysis of input uncertainty in hydrological
  modeling: 1. Theory} {Bayesian analysis of input uncertainty in hydrological
  modeling: 1. theory}.{\BBCQ}
\newblock
\APACjournalVolNumPages{Water resources research}{42}{3}{}.
\PrintBackRefs{\CurrentBib}

\bibitem [\protect \citeauthoryear {%
Kingma%
\ \BBA {} Ba%
}{%
Kingma%
\ \BBA {} Ba%
}{%
{\protect \APACyear {2014}}%
}]{%
kingma2014adam}
\APACinsertmetastar {%
kingma2014adam}%
\begin{APACrefauthors}%
Kingma, D\BPBI P.%
\BCBT {}\ \BBA {} Ba, J.%
\end{APACrefauthors}%
\unskip\
\newblock
\APACrefYearMonthDay{2014}{}{}.
\newblock
{\BBOQ}\APACrefatitle {Adam: A method for stochastic optimization} {Adam: A
  method for stochastic optimization}.{\BBCQ}
\newblock
\APACjournalVolNumPages{arXiv preprint arXiv:1412.6980}{}{}{}.
\newblock
\begin{APACrefURL} \url{https://arxiv.org/abs/1412.6980} \end{APACrefURL}
\PrintBackRefs{\CurrentBib}

\bibitem [\protect \citeauthoryear {%
Kirchner%
}{%
Kirchner%
}{%
{\protect \APACyear {2006}}%
}]{%
Kirchner2006}
\APACinsertmetastar {%
Kirchner2006}%
\begin{APACrefauthors}%
Kirchner, J\BPBI W.%
\end{APACrefauthors}%
\unskip\
\newblock
\APACrefYearMonthDay{2006}{}{}.
\newblock
{\BBOQ}\APACrefatitle {Getting the right answers for the right reasons: Linking
  measurements, analyses, and models to advance the science of hydrology}
  {Getting the right answers for the right reasons: Linking measurements,
  analyses, and models to advance the science of hydrology}.{\BBCQ}
\newblock
\APACjournalVolNumPages{Water Resources Research}{42}{3}{}.
\newblock
\begin{APACrefDOI} \doi{10.1029/2005WR004362} \end{APACrefDOI}
\PrintBackRefs{\CurrentBib}

\bibitem [\protect \citeauthoryear {%
Kuczera%
, Renard%
, Thyer%
\BCBL {}\ \BBA {} Kavetski%
}{%
Kuczera%
\ \protect \BOthers {.}}{%
{\protect \APACyear {2010}}%
}]{%
kuczera2010}
\APACinsertmetastar {%
kuczera2010}%
\begin{APACrefauthors}%
Kuczera, G.%
, Renard, B.%
, Thyer, M.%
\BCBL {}\ \BBA {} Kavetski, D.%
\end{APACrefauthors}%
\unskip\
\newblock
\APACrefYearMonthDay{2010}{}{}.
\newblock
{\BBOQ}\APACrefatitle {There are no hydrological monsters, just models and
  observations with large uncertainties!} {There are no hydrological monsters,
  just models and observations with large uncertainties!}{\BBCQ}
\newblock
\APACjournalVolNumPages{Hydrological Sciences Journal}{55}{6}{980--991}.
\newblock
\begin{APACrefDOI} \doi{10.1080/02626667.2010.504677} \end{APACrefDOI}
\PrintBackRefs{\CurrentBib}

\bibitem [\protect \citeauthoryear {%
Laloy%
\ \BBA {} Vrugt%
}{%
Laloy%
\ \BBA {} Vrugt%
}{%
{\protect \APACyear {2012}}%
}]{%
laloy_high-dimensional_2012}
\APACinsertmetastar {%
laloy_high-dimensional_2012}%
\begin{APACrefauthors}%
Laloy, E.%
\BCBT {}\ \BBA {} Vrugt, J\BPBI A.%
\end{APACrefauthors}%
\unskip\
\newblock
\APACrefYearMonthDay{2012}{}{}.
\newblock
{\BBOQ}\APACrefatitle {High-dimensional posterior exploration of hydrologic
  models using multiple-try {DREAM}({ZS}) and high-performance computing}
  {High-dimensional posterior exploration of hydrologic models using
  multiple-try {DREAM}({ZS}) and high-performance computing}.{\BBCQ}
\newblock
\APACjournalVolNumPages{Water Resources Research}{48}{1}{}.
\newblock
\begin{APACrefDOI} \doi{10.1029/2011wr010608} \end{APACrefDOI}
\PrintBackRefs{\CurrentBib}

\bibitem [\protect \citeauthoryear {%
Lane%
, Freer%
, Coxon%
\BCBL {}\ \BBA {} Wagener%
}{%
Lane%
\ \protect \BOthers {.}}{%
{\protect \APACyear {2021}}%
}]{%
lane2021incorporating}
\APACinsertmetastar {%
lane2021incorporating}%
\begin{APACrefauthors}%
Lane, R\BPBI A.%
, Freer, J\BPBI E.%
, Coxon, G.%
\BCBL {}\ \BBA {} Wagener, T.%
\end{APACrefauthors}%
\unskip\
\newblock
\APACrefYearMonthDay{2021}{}{}.
\newblock
{\BBOQ}\APACrefatitle {Incorporating uncertainty into multiscale parameter
  regionalization to evaluate the performance of nationally consistent
  parameter fields for a hydrological model} {Incorporating uncertainty into
  multiscale parameter regionalization to evaluate the performance of
  nationally consistent parameter fields for a hydrological model}.{\BBCQ}
\newblock
\APACjournalVolNumPages{Water Resources Research}{57}{10}{e2020WR028393}.
\PrintBackRefs{\CurrentBib}

\bibitem [\protect \citeauthoryear {%
Larnier%
\ \BBA {} Monnier%
}{%
Larnier%
\ \BBA {} Monnier%
}{%
{\protect \APACyear {2020}}%
}]{%
larnier2020hybrid}
\APACinsertmetastar {%
larnier2020hybrid}%
\begin{APACrefauthors}%
Larnier, K.%
\BCBT {}\ \BBA {} Monnier, J.%
\end{APACrefauthors}%
\unskip\
\newblock
\APACrefYearMonthDay{2020}{}{}.
\newblock
{\BBOQ}\APACrefatitle {Hybrid Neural Network--Variational Data Assimilation
  algorithm to infer river discharges from SWOT-like data} {Hybrid neural
  network--variational data assimilation algorithm to infer river discharges
  from swot-like data}.{\BBCQ}
\newblock
\APACjournalVolNumPages{Nonlinear Processes in Geophysics
  Discussions}{}{}{1--30}.
\PrintBackRefs{\CurrentBib}

\bibitem [\protect \citeauthoryear {%
LeCun%
, Bengio%
\BCBL {}\ \BBA {} Hinton%
}{%
LeCun%
\ \protect \BOthers {.}}{%
{\protect \APACyear {2015}}%
}]{%
LeCun2015}
\APACinsertmetastar {%
LeCun2015}%
\begin{APACrefauthors}%
LeCun, Y.%
, Bengio, Y.%
\BCBL {}\ \BBA {} Hinton, G.%
\end{APACrefauthors}%
\unskip\
\newblock
\APACrefYearMonthDay{2015}{}{}.
\newblock
{\BBOQ}\APACrefatitle {Deep learning} {Deep learning}.{\BBCQ}
\newblock
\APACjournalVolNumPages{Nature}{521}{}{436–444}.
\newblock
\begin{APACrefDOI} \doi{10.1038/nature14539} \end{APACrefDOI}
\PrintBackRefs{\CurrentBib}

\bibitem [\protect \citeauthoryear {%
Michel%
}{%
Michel%
}{%
{\protect \APACyear {1989}}%
}]{%
Michel1989}
\APACinsertmetastar {%
Michel1989}%
\begin{APACrefauthors}%
Michel, C.%
\end{APACrefauthors}%
\unskip\
\newblock
\APACrefYearMonthDay{1989}{}{}.
\newblock
{\BBOQ}\APACrefatitle {Hydrologie appliquée aux petits bassins ruraux}
  {Hydrologie appliquée aux petits bassins ruraux}.{\BBCQ}
\newblock
\APACjournalVolNumPages{Hydrology handbook (in French), Cemagref, Antony,
  France}{}{}{}.
\PrintBackRefs{\CurrentBib}

\bibitem [\protect \citeauthoryear {%
Mizukami%
\ \protect \BOthers {.}}{%
Mizukami%
\ \protect \BOthers {.}}{%
{\protect \APACyear {2017}}%
}]{%
mizukami2017towards}
\APACinsertmetastar {%
mizukami2017towards}%
\begin{APACrefauthors}%
Mizukami, N.%
, Clark, M\BPBI P.%
, Newman, A\BPBI J.%
, Wood, A\BPBI W.%
, Gutmann, E\BPBI D.%
, Nijssen, B.%
\BDBL {}Samaniego, L.%
\end{APACrefauthors}%
\unskip\
\newblock
\APACrefYearMonthDay{2017}{}{}.
\newblock
{\BBOQ}\APACrefatitle {Towards seamless large-domain parameter estimation for
  hydrologic models} {Towards seamless large-domain parameter estimation for
  hydrologic models}.{\BBCQ}
\newblock
\APACjournalVolNumPages{Water Resources Research}{53}{9}{8020--8040}.
\PrintBackRefs{\CurrentBib}

\bibitem [\protect \citeauthoryear {%
Monnier%
}{%
Monnier%
}{%
{\protect \APACyear {2021}}%
}]{%
monnier:hal-03040047}
\APACinsertmetastar {%
monnier:hal-03040047}%
\begin{APACrefauthors}%
Monnier, J.%
\end{APACrefauthors}%
\unskip\
\newblock
\APACrefYearMonthDay{2021}{{\APACmonth{11}}}{}.
\newblock
\APACrefbtitle {{Variational Data Assimilation and Model Learning}}
  {{Variational Data Assimilation and Model Learning}}\ [Master].
\newblock
France.
\newblock
\begin{APACrefURL} \url{https://hal.science/hal-03040047} \end{APACrefURL}
\newblock
\APACrefnote{Lecture}
\PrintBackRefs{\CurrentBib}

\bibitem [\protect \citeauthoryear {%
Monnier%
\ \protect \BOthers {.}}{%
Monnier%
\ \protect \BOthers {.}}{%
{\protect \APACyear {2016}}%
}]{%
MONNIER2016}
\APACinsertmetastar {%
MONNIER2016}%
\begin{APACrefauthors}%
Monnier, J.%
, Couderc, F.%
, Dartus, D.%
, Larnier, K.%
, Madec, R.%
\BCBL {}\ \BBA {} Vila, J\BHBI P.%
\end{APACrefauthors}%
\unskip\
\newblock
\APACrefYearMonthDay{2016}{}{}.
\newblock
{\BBOQ}\APACrefatitle {Inverse algorithms for 2D shallow water equations in
  presence of wet dry fronts: Application to flood plain dynamics} {Inverse
  algorithms for 2d shallow water equations in presence of wet dry fronts:
  Application to flood plain dynamics}.{\BBCQ}
\newblock
\APACjournalVolNumPages{Advances in Water Resources}{97}{}{11-24}.
\newblock
\begin{APACrefDOI} \doi{10.1016/j.advwatres.2016.07.005} \end{APACrefDOI}
\PrintBackRefs{\CurrentBib}

\bibitem [\protect \citeauthoryear {%
Mustafa%
, Nossent%
, Ghysels%
\BCBL {}\ \BBA {} Huysmans%
}{%
Mustafa%
\ \protect \BOthers {.}}{%
{\protect \APACyear {2018}}%
}]{%
mustafa_estimation_2018}
\APACinsertmetastar {%
mustafa_estimation_2018}%
\begin{APACrefauthors}%
Mustafa, S\BPBI M\BPBI T.%
, Nossent, J.%
, Ghysels, G.%
\BCBL {}\ \BBA {} Huysmans, M.%
\end{APACrefauthors}%
\unskip\
\newblock
\APACrefYearMonthDay{2018}{{\APACmonth{09}}}{}.
\newblock
{\BBOQ}\APACrefatitle {Estimation and {Impact} {Assessment} of {Input} and
  {Parameter} {Uncertainty} in {Predicting} {Groundwater} {Flow} {With} a
  {Fully} {Distributed} {Model}} {Estimation and {Impact} {Assessment} of
  {Input} and {Parameter} {Uncertainty} in {Predicting} {Groundwater} {Flow}
  {With} a {Fully} {Distributed} {Model}}.{\BBCQ}
\newblock
\APACjournalVolNumPages{Water Resources Research}{54}{9}{6585--6608}.
\newblock
\begin{APACrefDOI} \doi{10.1029/2017WR021857} \end{APACrefDOI}
\PrintBackRefs{\CurrentBib}

\bibitem [\protect \citeauthoryear {%
Odry%
}{%
Odry%
}{%
{\protect \APACyear {2017}}%
}]{%
Odry2017}
\APACinsertmetastar {%
Odry2017}%
\begin{APACrefauthors}%
Odry, J.%
\end{APACrefauthors}%
\unskip\
\newblock
\APACrefYear{2017}.
\unskip\
\newblock
\APACrefbtitle {Prédétermination des débits de crues extrêmes en sites non
  jaugés : régionalisation de la méthode par simulation SHYREG}
  {Prédétermination des débits de crues extrêmes en sites non jaugés :
  régionalisation de la méthode par simulation shyreg}\
  \APACtypeAddressSchool {\BPhD}{}{}.
\unskip\
\newblock
\begin{APACrefURL} \url{http://www.theses.fr/2017AIXM0424} \end{APACrefURL}
\unskip\
\newblock
\APACrefnote{Thèse de doctorat dirigée par Arnaud, Patrick Géosciences de
  l'environnement. Hydrologie Aix-Marseille 2017}
\PrintBackRefs{\CurrentBib}

\bibitem [\protect \citeauthoryear {%
Organde%
\ \protect \BOthers {.}}{%
Organde%
\ \protect \BOthers {.}}{%
{\protect \APACyear {2013}}%
}]{%
organde2013regionalisation}
\APACinsertmetastar {%
organde2013regionalisation}%
\begin{APACrefauthors}%
Organde, D.%
, Arnaud, P.%
, Fine, J\BHBI A.%
, Fouchier, C.%
, Folton, N.%
\BCBL {}\ \BBA {} Lavabre, J.%
\end{APACrefauthors}%
\unskip\
\newblock
\APACrefYearMonthDay{2013}{}{}.
\newblock
{\BBOQ}\APACrefatitle {R{\'e}gionalisation d'une m{\'e}thode de
  pr{\'e}d{\'e}termination de crue sur l'ensemble du territoire fran{\c{c}}ais:
  la m{\'e}thode SHYREG} {R{\'e}gionalisation d'une m{\'e}thode de
  pr{\'e}d{\'e}termination de crue sur l'ensemble du territoire fran{\c{c}}ais:
  la m{\'e}thode shyreg}.{\BBCQ}
\newblock
\APACjournalVolNumPages{Revue des Sciences de l’Eau}{26}{1}{65--78}.
\PrintBackRefs{\CurrentBib}

\bibitem [\protect \citeauthoryear {%
Oudin%
, Andr{\'e}assian%
, Perrin%
, Michel%
\BCBL {}\ \BBA {} Le~Moine%
}{%
Oudin%
\ \protect \BOthers {.}}{%
{\protect \APACyear {2008}}%
}]{%
oudin2008spatial}
\APACinsertmetastar {%
oudin2008spatial}%
\begin{APACrefauthors}%
Oudin, L.%
, Andr{\'e}assian, V.%
, Perrin, C.%
, Michel, C.%
\BCBL {}\ \BBA {} Le~Moine, N.%
\end{APACrefauthors}%
\unskip\
\newblock
\APACrefYearMonthDay{2008}{}{}.
\newblock
{\BBOQ}\APACrefatitle {Spatial proximity, physical similarity, regression and
  ungaged catchments: A comparison of regionalization approaches based on 913
  French catchments} {Spatial proximity, physical similarity, regression and
  ungaged catchments: A comparison of regionalization approaches based on 913
  french catchments}.{\BBCQ}
\newblock
\APACjournalVolNumPages{Water Resources Research}{44}{3}{}.
\PrintBackRefs{\CurrentBib}

\bibitem [\protect \citeauthoryear {%
Oudin%
\ \protect \BOthers {.}}{%
Oudin%
\ \protect \BOthers {.}}{%
{\protect \APACyear {2005}}%
}]{%
oudin2005potential}
\APACinsertmetastar {%
oudin2005potential}%
\begin{APACrefauthors}%
Oudin, L.%
, Hervieu, F.%
, Michel, C.%
, Perrin, C.%
, Andr{\'e}assian, V.%
, Anctil, F.%
\BCBL {}\ \BBA {} Loumagne, C.%
\end{APACrefauthors}%
\unskip\
\newblock
\APACrefYearMonthDay{2005}{}{}.
\newblock
{\BBOQ}\APACrefatitle {Which potential evapotranspiration input for a lumped
  rainfall--runoff model?: Part 2 Towards a simple and efficient potential
  evapotranspiration model for rainfall--runoff modelling} {Which potential
  evapotranspiration input for a lumped rainfall--runoff model?: Part 2 towards
  a simple and efficient potential evapotranspiration model for
  rainfall--runoff modelling}.{\BBCQ}
\newblock
\APACjournalVolNumPages{Journal of hydrology}{303}{1-4}{290--306}.
\PrintBackRefs{\CurrentBib}

\bibitem [\protect \citeauthoryear {%
Oudin%
, Kay%
, Andr{\'e}assian%
\BCBL {}\ \BBA {} Perrin%
}{%
Oudin%
\ \protect \BOthers {.}}{%
{\protect \APACyear {2010}}%
}]{%
oudin2010seemingly}
\APACinsertmetastar {%
oudin2010seemingly}%
\begin{APACrefauthors}%
Oudin, L.%
, Kay, A.%
, Andr{\'e}assian, V.%
\BCBL {}\ \BBA {} Perrin, C.%
\end{APACrefauthors}%
\unskip\
\newblock
\APACrefYearMonthDay{2010}{}{}.
\newblock
{\BBOQ}\APACrefatitle {Are seemingly physically similar catchments truly
  hydrologically similar?} {Are seemingly physically similar catchments truly
  hydrologically similar?}{\BBCQ}
\newblock
\APACjournalVolNumPages{Water Resources Research}{46}{11}{}.
\PrintBackRefs{\CurrentBib}

\bibitem [\protect \citeauthoryear {%
Parajka%
, Merz%
\BCBL {}\ \BBA {} Bl{\"o}schl%
}{%
Parajka%
\ \protect \BOthers {.}}{%
{\protect \APACyear {2005}}%
}]{%
parajka2005comparison}
\APACinsertmetastar {%
parajka2005comparison}%
\begin{APACrefauthors}%
Parajka, J.%
, Merz, R.%
\BCBL {}\ \BBA {} Bl{\"o}schl, G.%
\end{APACrefauthors}%
\unskip\
\newblock
\APACrefYearMonthDay{2005}{}{}.
\newblock
{\BBOQ}\APACrefatitle {A comparison of regionalisation methods for catchment
  model parameters} {A comparison of regionalisation methods for catchment
  model parameters}.{\BBCQ}
\newblock
\APACjournalVolNumPages{Hydrology and Earth System Sciences}{9}{3}{157--171}.
\PrintBackRefs{\CurrentBib}

\bibitem [\protect \citeauthoryear {%
Parajka%
\ \protect \BOthers {.}}{%
Parajka%
\ \protect \BOthers {.}}{%
{\protect \APACyear {2013}}%
}]{%
parajka2013comparative}
\APACinsertmetastar {%
parajka2013comparative}%
\begin{APACrefauthors}%
Parajka, J.%
, Viglione, A.%
, Rogger, M.%
, Salinas, J.%
, Sivapalan, M.%
\BCBL {}\ \BBA {} Bl{\"o}schl, G.%
\end{APACrefauthors}%
\unskip\
\newblock
\APACrefYearMonthDay{2013}{}{}.
\newblock
{\BBOQ}\APACrefatitle {Comparative assessment of predictions in ungauged
  basins--Part 1: Runoff-hydrograph studies} {Comparative assessment of
  predictions in ungauged basins--part 1: Runoff-hydrograph studies}.{\BBCQ}
\newblock
\APACjournalVolNumPages{Hydrology and Earth System
  Sciences}{17}{5}{1783--1795}.
\PrintBackRefs{\CurrentBib}

\bibitem [\protect \citeauthoryear {%
Perrin%
, Michel%
\BCBL {}\ \BBA {} Andrèassian%
}{%
Perrin%
\ \protect \BOthers {.}}{%
{\protect \APACyear {2003}}%
}]{%
perrin2003improvement}
\APACinsertmetastar {%
perrin2003improvement}%
\begin{APACrefauthors}%
Perrin, C.%
, Michel, C.%
\BCBL {}\ \BBA {} Andrèassian, V.%
\end{APACrefauthors}%
\unskip\
\newblock
\APACrefYearMonthDay{2003}{}{}.
\newblock
{\BBOQ}\APACrefatitle {Improvement of a parsimonious model for streamflow
  simulation} {Improvement of a parsimonious model for streamflow
  simulation}.{\BBCQ}
\newblock
\APACjournalVolNumPages{Journal of hydrology}{279}{1-4}{275--289}.
\PrintBackRefs{\CurrentBib}

\bibitem [\protect \citeauthoryear {%
Piotte%
\ \protect \BOthers {.}}{%
Piotte%
\ \protect \BOthers {.}}{%
{\protect \APACyear {2020}}%
}]{%
piotte2020}
\APACinsertmetastar {%
piotte2020}%
\begin{APACrefauthors}%
Piotte, O.%
, Montmerle, T.%
, Fouchier, C.%
, Belleudy, A.%
, Garandeau, L.%
, Janet, B.%
\BDBL {}Organde, D.%
\end{APACrefauthors}%
\unskip\
\newblock
\APACrefYearMonthDay{2020}{}{}.
\newblock
{\BBOQ}\APACrefatitle {Les évolutions du service d'avertissement sur les
  pluies intenses et les crues soudaines en France} {Les évolutions du service
  d'avertissement sur les pluies intenses et les crues soudaines en
  france}.{\BBCQ}
\newblock
\APACjournalVolNumPages{La Houille Blanche}{106}{6}{75--84}.
\newblock
\begin{APACrefDOI} \doi{10.1051/lhb/2020055} \end{APACrefDOI}
\PrintBackRefs{\CurrentBib}

\bibitem [\protect \citeauthoryear {%
Poncelet%
}{%
Poncelet%
}{%
{\protect \APACyear {2016}}%
}]{%
ponce2016}
\APACinsertmetastar {%
ponce2016}%
\begin{APACrefauthors}%
Poncelet, C.%
\end{APACrefauthors}%
\unskip\
\newblock
\APACrefYear{2016}.
\unskip\
\newblock
\APACrefbtitle {Du bassin au paramètre : jusqu'où peut-on régionaliser un
  modèle hydrologique conceptuel ?} {Du bassin au paramètre : jusqu'où
  peut-on régionaliser un modèle hydrologique conceptuel ?}\
  \APACtypeAddressSchool {\BPhD}{}{}.
\unskip\
\newblock
\begin{APACrefURL} \url{http://www.theses.fr/2016PA066550} \end{APACrefURL}
\unskip\
\newblock
\APACrefnote{Thèse de doctorat dirigée par Andréassian, Vazken et Oudin,
  Ludovic Hydrologie Paris 6 2016}
\PrintBackRefs{\CurrentBib}

\bibitem [\protect \citeauthoryear {%
Pujol%
, Garambois%
\BCBL {}\ \BBA {} Monnier%
}{%
Pujol%
\ \protect \BOthers {.}}{%
{\protect \APACyear {2022}}%
}]{%
Pujol-gmd-2022}
\APACinsertmetastar {%
Pujol-gmd-2022}%
\begin{APACrefauthors}%
Pujol, L.%
, Garambois, P\BHBI A.%
\BCBL {}\ \BBA {} Monnier, J.%
\end{APACrefauthors}%
\unskip\
\newblock
\APACrefYearMonthDay{2022}{}{}.
\newblock
{\BBOQ}\APACrefatitle {Multi-dimensional hydrological--hydraulic model with
  variational data assimilation for river networks and floodplains}
  {Multi-dimensional hydrological--hydraulic model with variational data
  assimilation for river networks and floodplains}.{\BBCQ}
\newblock
\APACjournalVolNumPages{Geoscientific Model Development}{15}{15}{6085--6113}.
\newblock
\begin{APACrefDOI} \doi{10.5194/gmd-15-6085-2022} \end{APACrefDOI}
\PrintBackRefs{\CurrentBib}

\bibitem [\protect \citeauthoryear {%
{Quintana-Segu{\'\i}}%
\ \protect \BOthers {.}}{%
{Quintana-Segu{\'\i}}%
\ \protect \BOthers {.}}{%
{\protect \APACyear {2008}}%
}]{%
Quintana2008}
\APACinsertmetastar {%
Quintana2008}%
\begin{APACrefauthors}%
{Quintana-Segu{\'\i}}, P.%
, {Le Moigne}, P.%
, {Durand}, Y.%
, {Martin}, E.%
, {Habets}, F.%
, {Baillon}, M.%
\BDBL {}{Morel}, S.%
\end{APACrefauthors}%
\unskip\
\newblock
\APACrefYearMonthDay{2008}{{\APACmonth{01}}}{}.
\newblock
{\BBOQ}\APACrefatitle {{Analysis of Near-Surface Atmospheric Variables:
  Validation of the SAFRAN Analysis over France}} {{Analysis of Near-Surface
  Atmospheric Variables: Validation of the SAFRAN Analysis over
  France}}.{\BBCQ}
\newblock
\APACjournalVolNumPages{Journal of Applied Meteorology and
  Climatology}{47}{1}{92}.
\newblock
\begin{APACrefDOI} \doi{10.1175/2007JAMC1636.1} \end{APACrefDOI}
\PrintBackRefs{\CurrentBib}

\bibitem [\protect \citeauthoryear {%
Rakovec%
\ \protect \BOthers {.}}{%
Rakovec%
\ \protect \BOthers {.}}{%
{\protect \APACyear {2016}}%
}]{%
Rakovec_2016}
\APACinsertmetastar {%
Rakovec_2016}%
\begin{APACrefauthors}%
Rakovec, O.%
, Kumar, R.%
, Mai, J.%
, Cuntz, M.%
, Thober, S.%
, Zink, M.%
\BDBL {}Samaniego, L.%
\end{APACrefauthors}%
\unskip\
\newblock
\APACrefYearMonthDay{2016}{}{}.
\newblock
{\BBOQ}\APACrefatitle {Multiscale and Multivariate Evaluation of Water Fluxes
  and States over European River Basins} {Multiscale and multivariate
  evaluation of water fluxes and states over european river basins}.{\BBCQ}
\newblock
\APACjournalVolNumPages{Journal of Hydrometeorology}{17}{1}{287 - 307}.
\newblock
\begin{APACrefDOI} \doi{10.1175/JHM-D-15-0054.1} \end{APACrefDOI}
\PrintBackRefs{\CurrentBib}

\bibitem [\protect \citeauthoryear {%
Razavi%
\ \BBA {} Coulibaly%
}{%
Razavi%
\ \BBA {} Coulibaly%
}{%
{\protect \APACyear {2013}}%
}]{%
razavi2013streamflow}
\APACinsertmetastar {%
razavi2013streamflow}%
\begin{APACrefauthors}%
Razavi, T.%
\BCBT {}\ \BBA {} Coulibaly, P.%
\end{APACrefauthors}%
\unskip\
\newblock
\APACrefYearMonthDay{2013}{}{}.
\newblock
{\BBOQ}\APACrefatitle {Streamflow prediction in ungauged basins: review of
  regionalization methods} {Streamflow prediction in ungauged basins: review of
  regionalization methods}.{\BBCQ}
\newblock
\APACjournalVolNumPages{Journal of hydrologic engineering}{18}{8}{958--975}.
\PrintBackRefs{\CurrentBib}

\bibitem [\protect \citeauthoryear {%
Reichl%
, Western%
, McIntyre%
\BCBL {}\ \BBA {} Chiew%
}{%
Reichl%
\ \protect \BOthers {.}}{%
{\protect \APACyear {2009}}%
}]{%
Reichl_2009}
\APACinsertmetastar {%
Reichl_2009}%
\begin{APACrefauthors}%
Reichl, J\BPBI P\BPBI C.%
, Western, A\BPBI W.%
, McIntyre, N\BPBI R.%
\BCBL {}\ \BBA {} Chiew, F\BPBI H\BPBI S.%
\end{APACrefauthors}%
\unskip\
\newblock
\APACrefYearMonthDay{2009}{}{}.
\newblock
{\BBOQ}\APACrefatitle {Optimization of a similarity measure for estimating
  ungauged streamflow} {Optimization of a similarity measure for estimating
  ungauged streamflow}.{\BBCQ}
\newblock
\APACjournalVolNumPages{Water Resources Research}{45}{10}{}.
\newblock
\begin{APACrefDOI} \doi{10.1029/2008WR007248} \end{APACrefDOI}
\PrintBackRefs{\CurrentBib}

\bibitem [\protect \citeauthoryear {%
Renard%
, Kavetski%
, Kuczera%
, Thyer%
\BCBL {}\ \BBA {} Franks%
}{%
Renard%
\ \protect \BOthers {.}}{%
{\protect \APACyear {2010}}%
}]{%
renard2010}
\APACinsertmetastar {%
renard2010}%
\begin{APACrefauthors}%
Renard, B.%
, Kavetski, D.%
, Kuczera, G.%
, Thyer, M.%
\BCBL {}\ \BBA {} Franks, S\BPBI W.%
\end{APACrefauthors}%
\unskip\
\newblock
\APACrefYearMonthDay{2010}{}{}.
\newblock
{\BBOQ}\APACrefatitle {Understanding predictive uncertainty in hydrologic
  modeling: The challenge of identifying input and structural errors}
  {Understanding predictive uncertainty in hydrologic modeling: The challenge
  of identifying input and structural errors}.{\BBCQ}
\newblock
\APACjournalVolNumPages{Water Resources Research}{46}{5}{}.
\newblock
\begin{APACrefDOI} \doi{10.1029/2009WR008328} \end{APACrefDOI}
\PrintBackRefs{\CurrentBib}

\bibitem [\protect \citeauthoryear {%
Saadi%
, Oudin%
\BCBL {}\ \BBA {} Ribstein%
}{%
Saadi%
\ \protect \BOthers {.}}{%
{\protect \APACyear {2019}}%
}]{%
saadi2019random}
\APACinsertmetastar {%
saadi2019random}%
\begin{APACrefauthors}%
Saadi, M.%
, Oudin, L.%
\BCBL {}\ \BBA {} Ribstein, P.%
\end{APACrefauthors}%
\unskip\
\newblock
\APACrefYearMonthDay{2019}{}{}.
\newblock
{\BBOQ}\APACrefatitle {Random forest ability in regionalizing hourly
  hydrological model parameters} {Random forest ability in regionalizing hourly
  hydrological model parameters}.{\BBCQ}
\newblock
\APACjournalVolNumPages{Water}{11}{8}{1540}.
\PrintBackRefs{\CurrentBib}

\bibitem [\protect \citeauthoryear {%
Samaniego%
, Kumar%
\BCBL {}\ \BBA {} Attinger%
}{%
Samaniego%
\ \protect \BOthers {.}}{%
{\protect \APACyear {2010}}%
}]{%
samaniego2010multiscale}
\APACinsertmetastar {%
samaniego2010multiscale}%
\begin{APACrefauthors}%
Samaniego, L.%
, Kumar, R.%
\BCBL {}\ \BBA {} Attinger, S.%
\end{APACrefauthors}%
\unskip\
\newblock
\APACrefYearMonthDay{2010}{}{}.
\newblock
{\BBOQ}\APACrefatitle {Multiscale parameter regionalization of a grid-based
  hydrologic model at the mesoscale} {Multiscale parameter regionalization of a
  grid-based hydrologic model at the mesoscale}.{\BBCQ}
\newblock
\APACjournalVolNumPages{Water Resources Research}{46}{5}{}.
\PrintBackRefs{\CurrentBib}

\bibitem [\protect \citeauthoryear {%
Samaniego%
\ \protect \BOthers {.}}{%
Samaniego%
\ \protect \BOthers {.}}{%
{\protect \APACyear {2017}}%
}]{%
Samaniego_hess-2017}
\APACinsertmetastar {%
Samaniego_hess-2017}%
\begin{APACrefauthors}%
Samaniego, L.%
, Kumar, R.%
, Thober, S.%
, Rakovec, O.%
, Zink, M.%
, Wanders, N.%
\BDBL {}Attinger, S.%
\end{APACrefauthors}%
\unskip\
\newblock
\APACrefYearMonthDay{2017}{}{}.
\newblock
{\BBOQ}\APACrefatitle {Toward seamless hydrologic predictions across spatial
  scales} {Toward seamless hydrologic predictions across spatial
  scales}.{\BBCQ}
\newblock
\APACjournalVolNumPages{Hydrology and Earth System
  Sciences}{21}{9}{4323--4346}.
\newblock
\begin{APACrefDOI} \doi{10.5194/hess-21-4323-2017} \end{APACrefDOI}
\PrintBackRefs{\CurrentBib}

\bibitem [\protect \citeauthoryear {%
Seibert%
}{%
Seibert%
}{%
{\protect \APACyear {1999}}%
}]{%
seibert1999regionalisation}
\APACinsertmetastar {%
seibert1999regionalisation}%
\begin{APACrefauthors}%
Seibert, J.%
\end{APACrefauthors}%
\unskip\
\newblock
\APACrefYearMonthDay{1999}{}{}.
\newblock
{\BBOQ}\APACrefatitle {Regionalisation of parameters for a conceptual
  rainfall-runoff model} {Regionalisation of parameters for a conceptual
  rainfall-runoff model}.{\BBCQ}
\newblock
\APACjournalVolNumPages{Agricultural and forest meteorology}{98}{}{279--293}.
\PrintBackRefs{\CurrentBib}

\bibitem [\protect \citeauthoryear {%
Sivapalan%
}{%
Sivapalan%
}{%
{\protect \APACyear {2003}}%
}]{%
Sivapalan_2003_regio}
\APACinsertmetastar {%
Sivapalan_2003_regio}%
\begin{APACrefauthors}%
Sivapalan, M.%
\end{APACrefauthors}%
\unskip\
\newblock
\APACrefYearMonthDay{2003}{}{}.
\newblock
{\BBOQ}\APACrefatitle {Prediction in ungauged basins: a grand challenge for
  theoretical hydrology} {Prediction in ungauged basins: a grand challenge for
  theoretical hydrology}.{\BBCQ}
\newblock
\APACjournalVolNumPages{Hydrological Processes}{17}{15}{3163-3170}.
\newblock
\begin{APACrefDOI} \doi{10.1002/hyp.5155} \end{APACrefDOI}
\PrintBackRefs{\CurrentBib}

\bibitem [\protect \citeauthoryear {%
Syed%
, Bouchard-Côté%
, Deligiannidis%
\BCBL {}\ \BBA {} Doucet%
}{%
Syed%
\ \protect \BOthers {.}}{%
{\protect \APACyear {2022}}%
}]{%
syed_non-reversible_2022}
\APACinsertmetastar {%
syed_non-reversible_2022}%
\begin{APACrefauthors}%
Syed, S.%
, Bouchard-Côté, A.%
, Deligiannidis, G.%
\BCBL {}\ \BBA {} Doucet, A.%
\end{APACrefauthors}%
\unskip\
\newblock
\APACrefYearMonthDay{2022}{{\APACmonth{04}}}{}.
\newblock
{\BBOQ}\APACrefatitle {Non-{Reversible} {Parallel} {Tempering}: {A} {Scalable}
  {Highly} {Parallel} {MCMC} {Scheme}} {Non-{Reversible} {Parallel}
  {Tempering}: {A} {Scalable} {Highly} {Parallel} {MCMC} {Scheme}}.{\BBCQ}
\newblock
\APACjournalVolNumPages{Journal of the Royal Statistical Society Series B:
  Statistical Methodology}{84}{2}{321--350}.
\newblock
\begin{APACrefDOI} \doi{10.1111/rssb.12464} \end{APACrefDOI}
\PrintBackRefs{\CurrentBib}

\bibitem [\protect \citeauthoryear {%
Vrugt%
, Ter~Braak%
, Clark%
, Hyman%
\BCBL {}\ \BBA {} Robinson%
}{%
Vrugt%
\ \protect \BOthers {.}}{%
{\protect \APACyear {2008}}%
}]{%
vrugt2008treatment}
\APACinsertmetastar {%
vrugt2008treatment}%
\begin{APACrefauthors}%
Vrugt, J\BPBI A.%
, Ter~Braak, C\BPBI J.%
, Clark, M\BPBI P.%
, Hyman, J\BPBI M.%
\BCBL {}\ \BBA {} Robinson, B\BPBI A.%
\end{APACrefauthors}%
\unskip\
\newblock
\APACrefYearMonthDay{2008}{}{}.
\newblock
{\BBOQ}\APACrefatitle {Treatment of input uncertainty in hydrologic modeling:
  Doing hydrology backward with Markov chain Monte Carlo simulation} {Treatment
  of input uncertainty in hydrologic modeling: Doing hydrology backward with
  markov chain monte carlo simulation}.{\BBCQ}
\newblock
\APACjournalVolNumPages{Water Resources Research}{44}{12}{}.
\PrintBackRefs{\CurrentBib}

\bibitem [\protect \citeauthoryear {%
Wang%
\ \protect \BOthers {.}}{%
Wang%
\ \protect \BOthers {.}}{%
{\protect \APACyear {2023}}%
}]{%
wang2023research}
\APACinsertmetastar {%
wang2023research}%
\begin{APACrefauthors}%
Wang, W.%
, Zhao, Y.%
, Tu, Y.%
, Dong, R.%
, Ma, Q.%
\BCBL {}\ \BBA {} Liu, C.%
\end{APACrefauthors}%
\unskip\
\newblock
\APACrefYearMonthDay{2023}{}{}.
\newblock
{\BBOQ}\APACrefatitle {Research on parameter regionalization of distributed
  hydrological model based on machine learning} {Research on parameter
  regionalization of distributed hydrological model based on machine
  learning}.{\BBCQ}
\newblock
\APACjournalVolNumPages{Water}{15}{3}{518}.
\PrintBackRefs{\CurrentBib}

\bibitem [\protect \citeauthoryear {%
Wid{\'e}n-Nilsson%
, Halldin%
\BCBL {}\ \BBA {} Xu%
}{%
Wid{\'e}n-Nilsson%
\ \protect \BOthers {.}}{%
{\protect \APACyear {2007}}%
}]{%
widen2007global}
\APACinsertmetastar {%
widen2007global}%
\begin{APACrefauthors}%
Wid{\'e}n-Nilsson, E.%
, Halldin, S.%
\BCBL {}\ \BBA {} Xu, C\BHBI y.%
\end{APACrefauthors}%
\unskip\
\newblock
\APACrefYearMonthDay{2007}{}{}.
\newblock
{\BBOQ}\APACrefatitle {Global water-balance modelling with WASMOD-M: Parameter
  estimation and regionalisation} {Global water-balance modelling with
  wasmod-m: Parameter estimation and regionalisation}.{\BBCQ}
\newblock
\APACjournalVolNumPages{Journal of Hydrology}{340}{1-2}{105--118}.
\PrintBackRefs{\CurrentBib}

\bibitem [\protect \citeauthoryear {%
Zhu%
, Byrd%
, Lu%
\BCBL {}\ \BBA {} Nocedal%
}{%
Zhu%
\ \protect \BOthers {.}}{%
{\protect \APACyear {1997}}%
}]{%
Zhu1997}
\APACinsertmetastar {%
Zhu1997}%
\begin{APACrefauthors}%
Zhu, C.%
, Byrd, R\BPBI H.%
, Lu, P.%
\BCBL {}\ \BBA {} Nocedal, J.%
\end{APACrefauthors}%
\unskip\
\newblock
\APACrefYearMonthDay{1997}{}{}.
\newblock
{\BBOQ}\APACrefatitle {Algorithm 778: L-BFGS-B: Fortran Subroutines for
  Large-Scale Bound-Constrained Optimization.} {Algorithm 778: L-bfgs-b:
  Fortran subroutines for large-scale bound-constrained optimization.}{\BBCQ}
\newblock
\APACjournalVolNumPages{ACM Trans. Math. Softw.}{23}{4}{550-560}.
\newblock
\begin{APACrefURL}
  \url{http://dblp.uni-trier.de/db/journals/toms/toms23.html#ZhuBLN97}
  \end{APACrefURL}
\PrintBackRefs{\CurrentBib}

\end{thebibliography}

%Reference citation instructions and examples:
%
% Please use ONLY \cite and \citeA for reference citations.
% \cite for parenthetical references
% ...as shown in recent studies (Simpson et al., 2019)
% \citeA for in-text citations
% ...Simpson et al. (2019) have shown...
%
%
%...as shown by \citeA{jskilby}.
%...as shown by \citeA{lewin76}, \citeA{carson86}, \citeA{bartoldy02}, and \citeA{rinaldi03}.
%...has been shown \cite{jskilbye}.
%...has been shown \cite{lewin76,carson86,bartoldy02,rinaldi03}.
%... \cite <i.e.>[]{lewin76,carson86,bartoldy02,rinaldi03}.
%...has been shown by \cite <e.g.,>[and others]{lewin76}.
%
% apacite uses < > for prenotes and [ ] for postnotes
% DO NOT use other cite commands (e.g., \citet, \citep, \citeyear, \nocite, \citealp, etc.).
%

\end{document}